\documentclass[journal]{IEEEtran}

\usepackage{xr}
\usepackage{url}            %
\usepackage{booktabs}       %
\usepackage{arydshln}
\usepackage{amsfonts}       %
\usepackage{nicefrac}       %
\usepackage{microtype}      %
\usepackage{graphicx}
\usepackage{subcaption}
\usepackage{soul}
\usepackage{tablefootnote}
\usepackage{multirow}
\usepackage{amsmath}
\usepackage{amsthm}  %
\usepackage{amssymb} %
\theoremstyle{definition}
\newtheorem{definition}{Definition}[]
\usepackage{color}

\usepackage{hyperref}       %

\newcommand{\furl}[1]{\footnote{\url{http://#1}}}

\usepackage{balance}

\ifCLASSOPTIONcompsoc
  \usepackage[nocompress]{cite}
\else
  \usepackage{cite}
\fi

\ifCLASSINFOpdf
\else
\fi

\begin{document}
\title{A Survey on Knowledge Graphs: \\Representation, Acquisition and Applications}

\author{Shaoxiong Ji,
        Shirui Pan, ~\IEEEmembership{Member,~IEEE,}
        ~Erik~Cambria,~\IEEEmembership{Senior~Member,~IEEE,}\\
        ~Pekka~Marttinen, ~Philip S. Yu, ~\IEEEmembership{Life Fellow,~IEEE}
\thanks{Manuscript received August 09, 2020; revised November xx, 2020; accepted March 30, 2021. This work is supported in part by  NSF under grants III-1763325, III-1909323, SaTC-1930941, in part by the Agency for Science, Technology and Research (A*STAR) under its AME Programmatic Funding Scheme (Project \#A18A2b0046), and in part by the Academy of Finland (grants 336033, 315896), BusinessFinland (grant 884/31/2018), and EU H2020 (grant 101016775).
 \textit{(Corresponding author: Shirui Pan.)}
}
\IEEEcompsocitemizethanks{\IEEEcompsocthanksitem S. Ji and P. Marttinen are with Aalto University, Finland.
E-mail: \{shaoxiong.ji;~pekka.marttinen\}@aalto.fi
\IEEEcompsocthanksitem S. Pan is with the Department of Data Science and AI, Faculty of IT, Monash University, Australia. E-mail:~shirui.pan@monash.edu
\IEEEcompsocthanksitem E. Cambria is with Nanyang Technological University, Singapore. E-mail:~cambria@ntu.edu.sg
\IEEEcompsocthanksitem P.S. Yu is with University of Illinois at Chicago, USA.  E-mail:~psyu@uic.edu
}%
}

\markboth{IEEE Transactions on Neural Networks and Learning Systems, 2021}%
{Shell \MakeLowercase{\textit{et al.}}: Bare Demo of IEEEtran.cls for Computer Society Journals}

\maketitle

\begin{abstract}
Human knowledge provides a formal understanding of the world. 
Knowledge graphs that represent structural relations between entities have become an increasingly popular research direction towards cognition and human-level intelligence. 
In this survey, we provide a comprehensive review of knowledge graph covering overall research topics about 1) knowledge graph representation learning, 2) knowledge acquisition and completion, 3) temporal knowledge graph, and 4) knowledge-aware applications, and summarize recent breakthroughs and perspective directions to facilitate future research. 
We propose a full-view categorization and new taxonomies on these topics. 
Knowledge graph embedding is organized from four aspects of representation space, scoring function, encoding models, and auxiliary information. For knowledge acquisition, especially knowledge graph completion, embedding methods, path inference, and logical rule reasoning, are reviewed.
We further explore several emerging topics, including meta relational learning, commonsense reasoning, and temporal knowledge graphs.
To facilitate future research on knowledge graphs, we also provide a curated collection of datasets and open-source libraries on different tasks. 
In the end, we have a thorough outlook on several promising research directions. 
\end{abstract}


\IEEEpeerreviewmaketitle

\section{Introduction}\label{sec:introduction}
\IEEEPARstart{I}{ncorporating} human knowledge is one of the research directions of artificial intelligence (AI). 
Knowledge representation and reasoning, inspired by human problem solving, is to represent knowledge for intelligent systems to gain the ability to solve complex tasks~\cite{newell1959report, shortliffe2012computer}. 
Recently, knowledge graphs as a form of structured human knowledge have drawn great research attention from both the academia and the industry~\cite{dong2014knowledge, nickel2015review, wang2017knowledge, hogan2020knowledge}. 
A knowledge graph is a structured representation of facts, consisting of entities, relationships, and semantic descriptions. Entities can be real-world objects and abstract concepts, relationships represent the relation between entities, and semantic descriptions of entities, and their relationships contain types and properties with a well-defined meaning. Property graphs or attributed graphs are widely used, in which nodes and relations have properties or attributes. 

\begin{figure}[htbp]
\begin{center}
\begin{subfigure}[b]{0.23\textwidth}
\includegraphics[width=\textwidth]{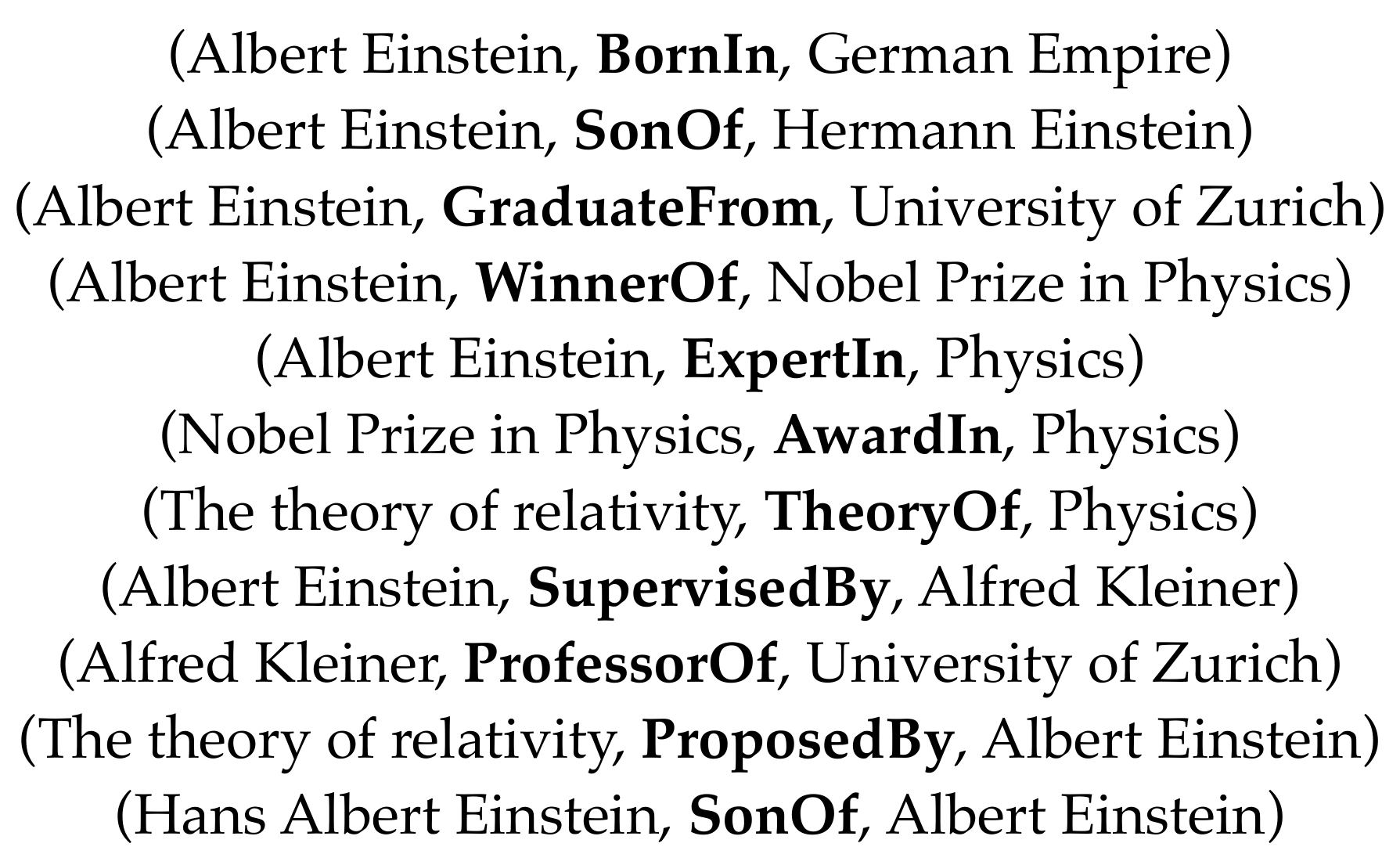}
\caption{Factual triples in knowledge base.}
\label{fig:kb}
\end{subfigure}
\begin{subfigure}[b]{0.23\textwidth}
\includegraphics[width=\textwidth]{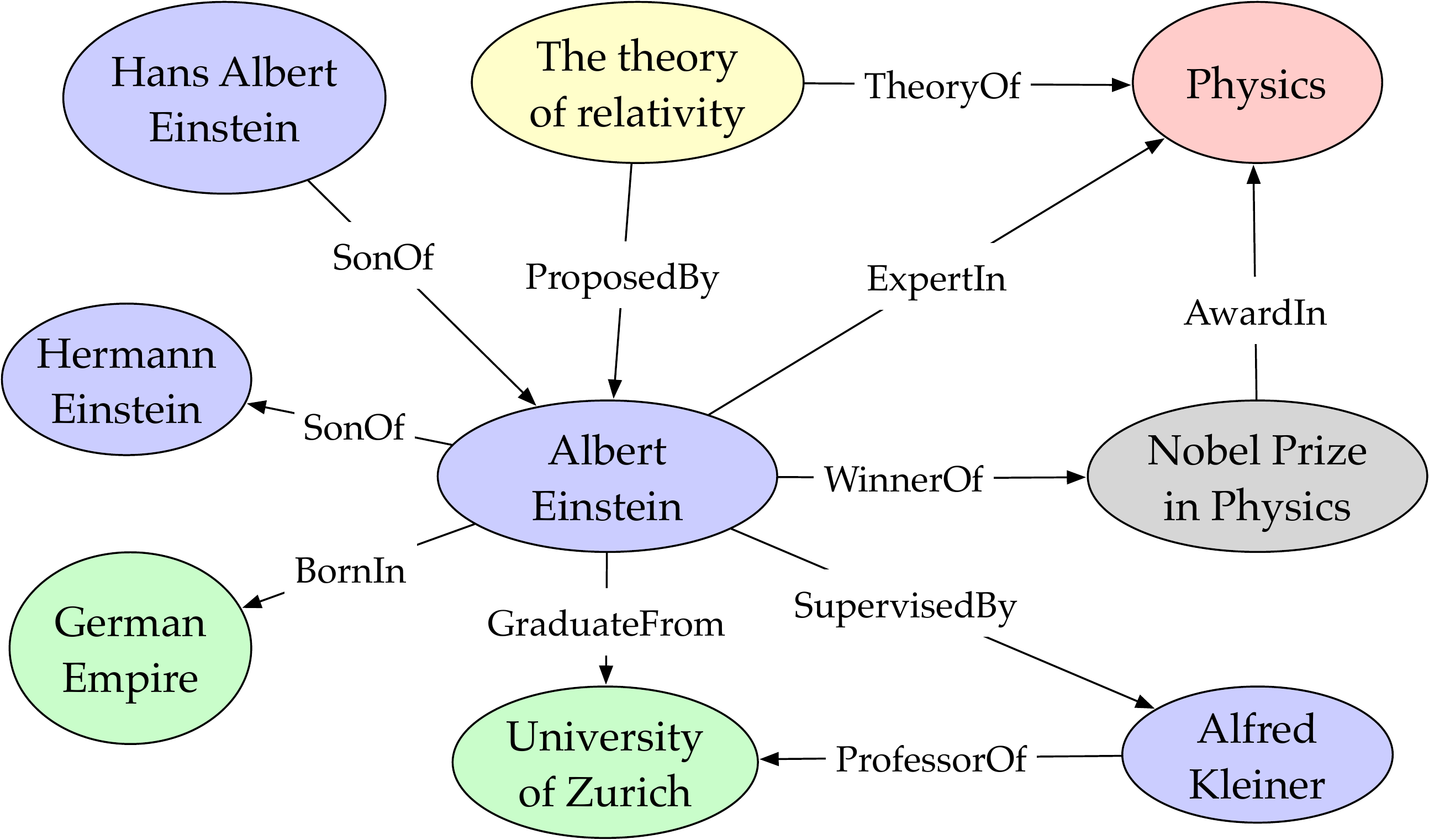}
\caption{Entities and relations in knowledge graph.}
\label{fig:kg}
\end{subfigure}
\caption{An example of knowledge base and knowledge graph.}
\label{fig:example}
\end{center}
\end{figure}

The term of knowledge graph is synonymous with knowledge base with a minor difference. A knowledge graph can be viewed as a graph when considering its graph structure~\cite{stokman1988structuring}. When it involves formal semantics, it can be taken as a knowledge base for interpretation and inference over facts~\cite{bordes2011learning}. Examples of knowledge base and knowledge graph are illustrated in Fig.~\ref{fig:example}. Knowledge can be expressed in a factual triple in the form of $(\textit{head},~\texttt{relation}, \textit{tail})$ or $(\textit{subject},~\texttt{predicate}, \textit{object})$ under the resource description framework (RDF), for example, $(\textit{Albert~Einstein},~\texttt{WinnerOf},~\textit{Nobel~Prize})$. It can also be represented as a directed graph with nodes as entities and edges as relations.
For simplicity and following the trend of the research community, this paper uses the terms knowledge graph and knowledge base interchangeably. 

Recent advances in knowledge-graph-based research focus on knowledge representation learning (KRL) or knowledge graph embedding (KGE) by mapping entities and relations into low-dimensional vectors while capturing their semantic meanings~\cite{wang2017knowledge, lin2018knowledge}. 
Specific knowledge acquisition tasks include knowledge graph completion (KGC), triple classification, entity recognition, and relation extraction. 
Knowledge-aware models benefit from the integration of heterogeneous information, rich ontologies and semantics for knowledge representation, and multi-lingual knowledge. Thus, many real-world applications such as recommendation systems and question answering have been brought about prosperity with the ability of commonsense understanding and reasoning. 
Some real-world products, for example, Microsoft's Satori and Google's Knowledge Graph~\cite{dong2014knowledge}, have shown a strong capacity to provide more efficient services. 

This paper conducts a comprehensive survey of current literature on knowledge graphs, which enriches graphs with more context, intelligence, and semantics for knowledge acquisition and knowledge-aware applications. Our main contributions are summarized as follows.

\begin{itemize}
 \item \textbf{Comprehensive review.} We conduct a comprehensive review of the origin of knowledge graph and modern techniques for relational learning on knowledge graphs. Major neural architectures of knowledge graph representation learning and reasoning are introduced and compared. Moreover, we provide a complete overview of many applications in different domains. 
 \item \textbf{Full-view categorization and new taxonomies.} A full-view categorization of research on knowledge graph, together with fine-grained new taxonomies are presented. Specifically, in the high-level, we review the research on knowledge graphs in four aspects: KRL, knowledge acquisition, temporal knowledge graphs, and knowledge-aware applications. For KRL, we further propose fine-grained taxonomies into four views, including representation space, scoring function, encoding models, and auxiliary information. For knowledge acquisition, KGC is reviewed under embedding-based ranking, relational path reasoning, logical rule reasoning, and meta relational learning; entity acquisition tasks are divided into entity recognition, typing, disambiguation, and alignment; and relation extraction is discussed according to the neural paradigms. 
 \item \textbf{Wide coverage on emerging advances.} We provide wide coverage on emerging topics, including transformer-based knowledge encoding, graph neural network (GNN) based knowledge propagation, reinforcement learning-based path reasoning, and meta relational learning.
 \item \textbf{Summary and outlook on future directions.} This survey provides a summary of each category and highlights promising future research directions. 
\end{itemize}

The remainder of this survey is organized as follows: first, an overview of knowledge graphs including history, notations, definitions, and categorization is given in Section~\ref{sec:overview}; then, we discuss KRL in Section~\ref{sec:krl} from four scopes; next, our review goes to tasks of knowledge acquisition and temporal knowledge graphs in Section~\ref{sec:acquisition} and Section~\ref{sec:temporal};
downstream applications are introduced in Section~\ref{sec:application}; finally, we discuss future research directions, together with a conclusion in the end. Other information, including KRL model training and a collection of knowledge graph datasets and open-source implementations, can be found in the appendices.

\section{Overview}
\label{sec:overview}
\subsection{A Brief History of Knowledge Bases}
Knowledge representation has experienced a long-period history of development in the fields of logic and AI. The idea of graphical knowledge representation firstly dated back to 1956 as the concept of semantic net proposed by Richens~\cite{richens1956preprogramming}, while the symbolic logic knowledge can go back to the General Problem Solver~\cite{newell1959report} in 1959. 
The knowledge base is firstly used with knowledge-based systems for reasoning and problem-solving. MYCIN~\cite{shortliffe2012computer} is one of the most famous rule-based expert systems for medical diagnosis with a knowledge base of about 600 rules. 
Later, the community of human knowledge representation saw the development of frame-based language, rule-based, and hybrid representations. Approximately at the end of this period, the Cyc project\furl{cyc.com} began, aiming at assembling human knowledge. 
Resource description framework (RDF)\footnote{Released as W3C recommendation in 1999 available at \url{http://w3.org/TR/1999/REC-rdf-syntax-19990222}.} and Web Ontology Language (OWL)\furl{w3.org/TR/owl-guide} were released in turn, and became important standards of the Semantic Web\furl{w3.org/standards/semanticweb}. 
Then, many open knowledge bases or ontologies were published, such as WordNet, DBpedia, YAGO, and Freebase. %
Stokman and Vries~\cite{stokman1988structuring} proposed a modern idea of structure knowledge in a graph in 1988. However, it was in 2012 that the concept of knowledge graph gained great popularity since its first launch by Google's search engine\furl{blog.google/products/search/introducing-knowledge-graph-things-not}, where the knowledge fusion framework called Knowledge Vault~\cite{dong2014knowledge} was proposed to build large-scale knowledge graphs.
A brief road map of knowledge base history is illustrated in Fig.~\ref{fig:history} in Appendix \ref{sup:history}.
Many general knowledge graph databases and domain-specific knowledge bases have been released to facilitate research. We introduce more general and domain-specific knowledge bases in Appendices~\ref{app:general-datasets} and~\ref{app:domain-specific}.

\subsection{Definitions and Notations}
Most efforts have been made to give a definition by describing general semantic representation or essential characteristics. However, there is no such wide-accepted formal definition. 
Paulheim~\cite{paulheim2017knowledge} defined four criteria for knowledge graphs.
Ehrlinger and W{\"o}{\ss}~\cite{ehrlinger2016towards} analyzed several existing definitions and proposed Definition~\ref{def:kg16}, which emphasizes the reasoning engine of knowledge graphs. Wang et al.~\cite{wang2017knowledge} proposed a definition as a multi-relational graph in Definition~\ref{def:kg17}. 
Following previous literature, we define a knowledge graph as $\mathcal{G}= \{\mathcal{E}, \mathcal{R}, \mathcal{F}\}$, where $\mathcal{E}$, $\mathcal{R}$ and $\mathcal{F}$ are sets of entities, relations and facts, respectively. 
A fact is denoted as a triple $(h, r, t) \in \mathcal{F}$.

\theoremstyle{definition}
\begin{definition}[Ehrlinger and W{\"o}{\ss}\cite{ehrlinger2016towards}]
\label{def:kg16}
A knowledge graph acquires and integrates information into an ontology and applies a reasoner to derive new knowledge.
\end{definition}

\theoremstyle{definition}
\begin{definition}[Wang et al.\cite{wang2017knowledge}]
\label{def:kg17}
A knowledge graph is a multi-relational graph composed of entities and relations which are regarded as nodes and different types of edges, respectively.
\end{definition}

Specific notations and their descriptions are listed in Table~\ref{tab:notations}. Details of several mathematical operations are explained in Appendix~\ref{sup:math}.

\begin{table}[htp]
\scriptsize
\caption{Notations and descriptions.}
\begin{center}
\begin{tabular}{l l}
\toprule
Notation & Description \\
\midrule
$\mathcal{G}$ & A knowledge graph \\
$ \mathcal{F}$ & A set of facts \\
$(h, r, t)$ & A triple of head, relation and tail \\
$(\mathbf{h}, \mathbf{r}, \mathbf{t})$ & Embedding of head, relation and tail \\
$ r\in \mathcal{R}, e \in \mathcal{E}$ & Relation set and entity set \\
$v \in \mathcal{V}$ & Vertex in vertice set \\
$\xi \in \mathcal{E_G}$ & Edge in edge set \\
$e_s, e_q, e_t$ & Source/query/current entity \\
$r_q$ & Query relation \\
$< w_1, \dots, w_n>$ & Text corpus \\
$d_{\cdot}(\cdot)$ & Distance metric in specific space \\
$f_r(\mathbf{h},\mathbf{t})$ & Scoring function \\
$\sigma(\cdot)$, $g(\cdot)$ & Non-linear activation function \\
$\mathbf{M}_r$ & Mapping matrix \\
$\mathbf{\widehat M}$ & Tensor \\
$\mathcal{L}$ & Loss function \\
\hdashline
$\mathbb{R}^d$ & $d$ dimensional real-valued space \\
$\mathbb{C}^d$ & $d$ dimensional complex space \\
$\mathbb{H}^d$ & $d$ dimensional hypercomplex space \\
$\mathbb{T}^d$ & $d$ dimensional torus space \\
$\mathbb{B}_{c}^{d}$ & $d$ dimensional hyperbolic space with curvature $c$ \\
$\mathcal{N}(\mathbf{u}, \sigma^2\mathbf{I}) $ & Gaussian distribution \\
\hdashline
$\langle \mathbf{h}, \mathbf{t}\rangle$ & Hermitian dot product \\
$\mathbf{t} \otimes \mathbf{r}$ & Hamilton product \\
$\mathbf{h} \circ \mathbf{t}$, $\mathbf{h} \odot \mathbf{t}$ & Hadmard (element-wise) product \\
$\mathbf{h} \star \mathbf{t}$ & Circular correlation \\
$\operatorname{concat}()$, $[\mathbf{h}, \mathbf{r}]$ & Vectors/matrices concatenation \\
$\boldsymbol{\omega}$ & Convolutional filters \\
$*$ & Convolution operator \\
\bottomrule
\end{tabular}
\end{center}
\label{tab:notations}
\end{table}%

\subsection{Categorization of Research on Knowledge Graph}
\begin{figure}[htbp]
\begin{center}
\includegraphics[width=0.49\textwidth]{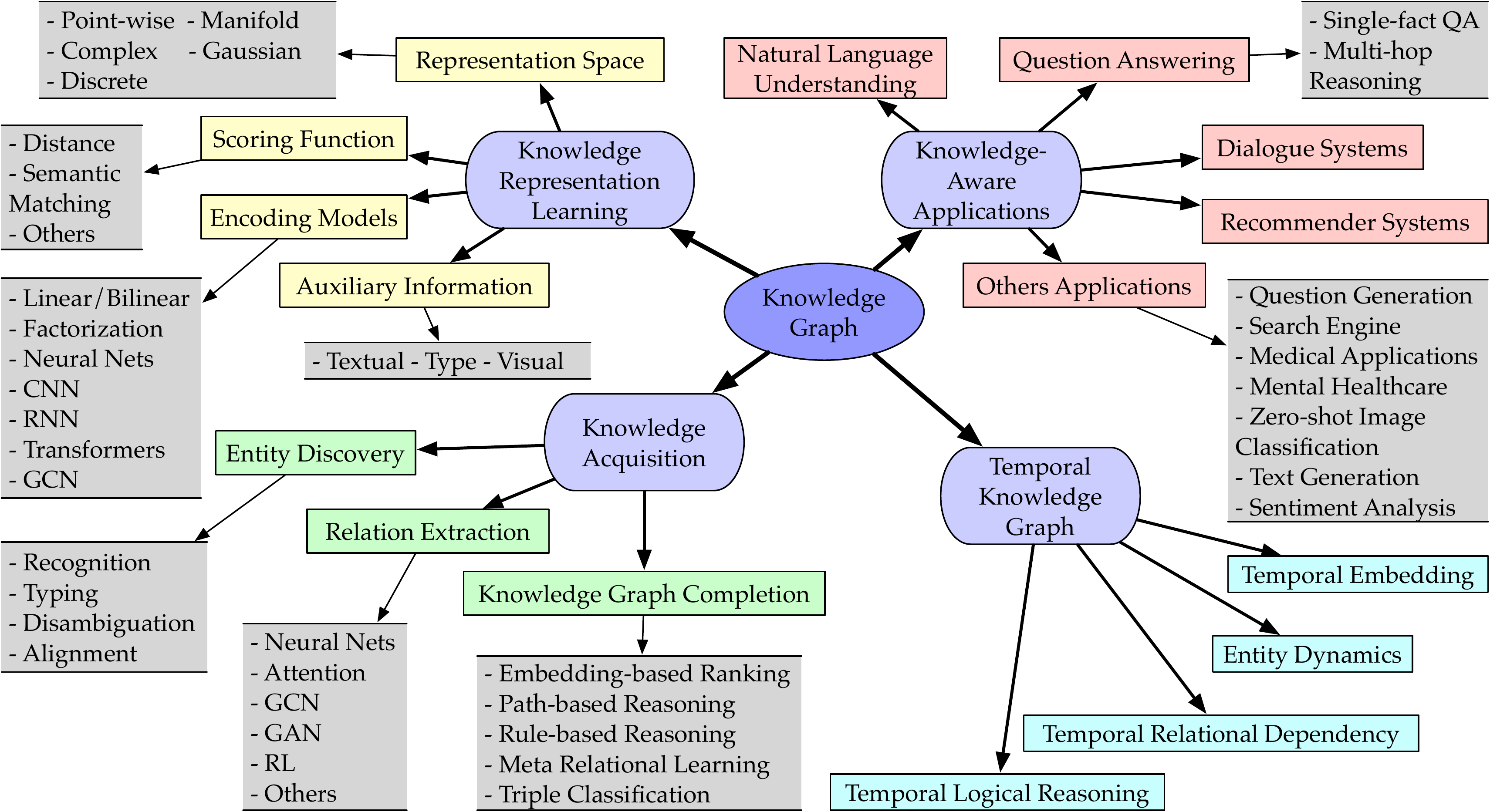}
\caption{Categorization of research on knowledge graphs.}
\label{fig:categorization}
\end{center}
\end{figure}

This survey provides a comprehensive literature review on the research of knowledge graphs, namely KRL, knowledge acquisition, and a wide range of downstream knowledge-aware applications, where many recent advanced deep learning techniques are integrated. The overall categorization of the research is illustrated in Fig.~\ref{fig:categorization}.

\vspace{2mm}
\textbf{Knowledge Representation Learning} is a critical research issue of knowledge graph which paves the way for many knowledge acquisition tasks and downstream applications. 
We categorize KRL into four aspects of \textit{representation space}, \textit{scoring function}, \textit{encoding models} and \textit{auxiliary information}, providing a clear workflow for developing a KRL model. Specific ingredients include: 
\begin{enumerate}
 \item \textit{representation space} in which the relations and entities are represented;
 \item \textit{scoring function} for measuring the plausibility of factual triples; 
 \item \textit{encoding models} for representing and learning relational interactions; 
 \item \textit{auxiliary information} to be incorporated into the embedding methods. %
\end{enumerate}

Representation learning includes point-wise space, manifold, complex vector space, Gaussian distribution, and discrete space. 
Scoring metrics are generally divided into distance-based and similarity matching based scoring functions. 
Current research focuses on encoding models, including linear/bilinear models, factorization, and neural networks. Auxiliary information considers textual, visual, and type information.

\vspace{2mm}
\textbf{Knowledge Acquisition} tasks are divided into three categories, i.e., KGC, relation extraction, and entity discovery. The first one is for expanding existing knowledge graphs, while the other two discover new knowledge (aka relations and entities) from the text. KGC falls into the following categories: embedding-based ranking, relation path reasoning, rule-based reasoning, and meta relational learning. Entity discovery includes recognition, disambiguation, typing, and alignment. Relation extraction models utilize attention mechanism, graph convolutional networks (GCNs), adversarial training, reinforcement learning, deep residual learning, and transfer learning.

\vspace{2mm}
\textbf{Temporal Knowledge Graphs} incorporate temporal information for representation learning. This survey categorizes four research fields, including temporal embedding, entity dynamics, temporal relational dependency, and temporal logical reasoning.

\vspace{2mm}
\textbf{Knowledge-aware Applications} include natural language understanding (NLU), question answering, recommendation systems, and miscellaneous real-world tasks, which inject knowledge to improve representation learning.

\subsection{Related~Surveys}
Previous survey papers on knowledge graphs mainly focus on statistical relational learning~\cite{nickel2015review}, knowledge graph refinement~\cite{paulheim2017knowledge}, Chinese knowledge graph construction~\cite{wu2018survey}, knowledge reasoning~\cite{chen2020review}, KGE~\cite{wang2017knowledge} or KRL~\cite{lin2018knowledge}. The latter two surveys are more related to our work. Lin et al.~\cite{lin2018knowledge} presented KRL in a linear manner, with a concentration on quantitative analysis. 
Wang et al.~\cite{wang2017knowledge} categorized KRL according to scoring functions and specifically focused on the type of information utilized in KRL. It provides a general view of current research only from the perspective of scoring metrics. Our survey goes deeper to the flow of KRL and provides a full-scaled view from four-folds, including representation space, scoring function, encoding models, and auxiliary information. Besides, our paper provides a comprehensive review of knowledge acquisition and knowledge-aware applications with several emerging topics such as knowledge-graph-based reasoning and few-shot learning discussed. 

\section{Knowledge~Representation~Learning}
\label{sec:krl}

KRL is also known as KGE, multi-relation learning, and statistical relational learning in the literature. This section reviews recent advances on distributed representation learning with rich semantic information of entities and relations form four scopes including representation space (representing entities and relations, \textbf{Sec.~\ref{sec:space}}), scoring function (measuring the plausibility of facts, \textbf{Sec.~\ref{sec:function}}), encoding models (modeling the semantic interaction of facts, \textbf{Sec.~\ref{sec:encoding}}), and auxiliary information (utilizing external information, \textbf{Sec.~\ref{sec:multi}}). We further provide a summary in \textbf{Sec.~\ref{sec:summarykrl}}. The training strategies for KRL models are reviewed in Appendix~\ref{sup:KRL-training}.

\subsection{Representation~Space}\label{sec:space}
\begin{figure*}[htbp]
\begin{center}
\begin{subfigure}[b]{0.20\textwidth}
 \centering
 \includegraphics[height=3.6cm]{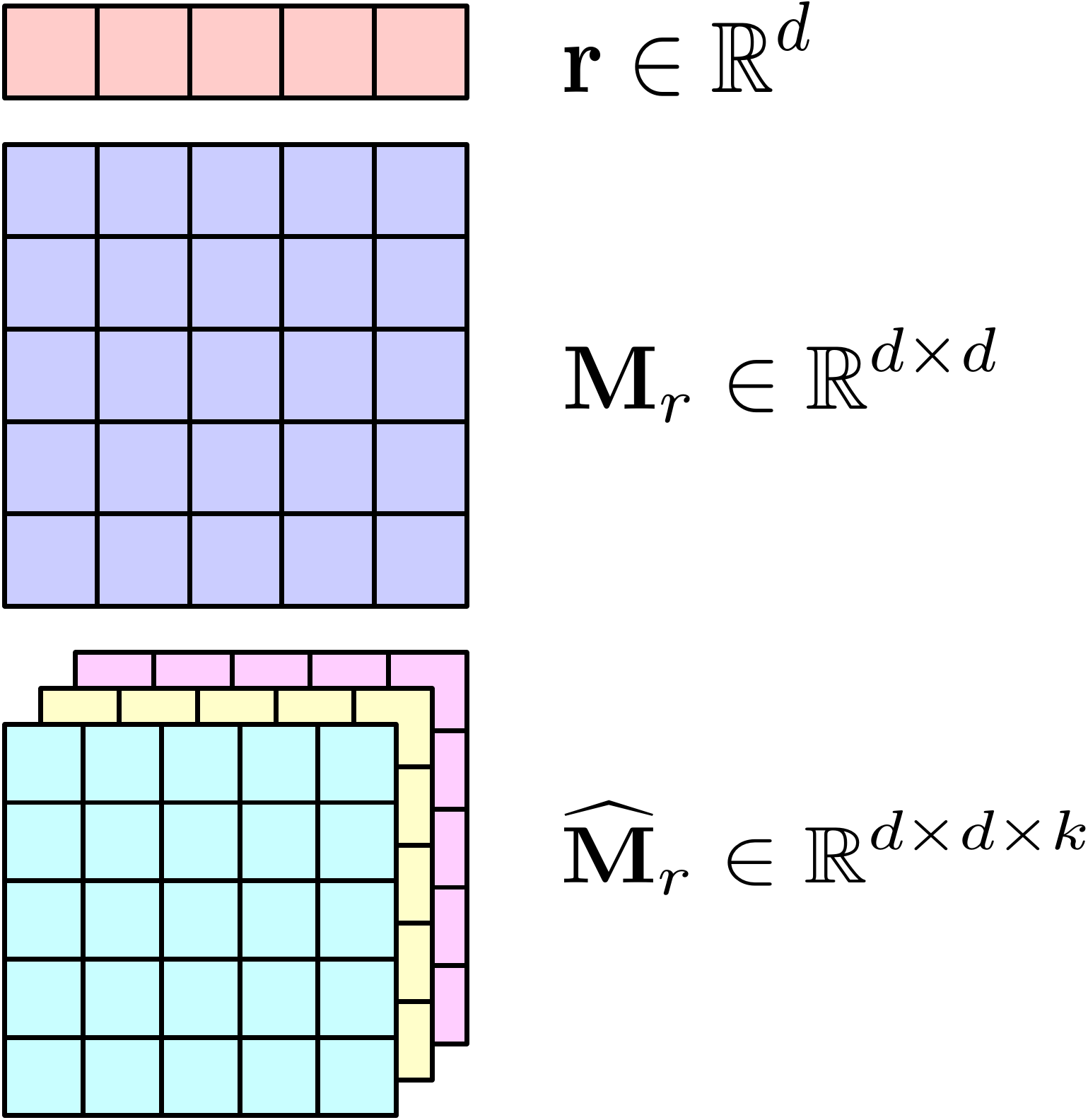}
 \caption{Point-wise space.}
 \label{fig:space-point}
\end{subfigure}
\quad
\begin{subfigure}[b]{0.20\textwidth}
 \centering
 \includegraphics[height=3.6cm]{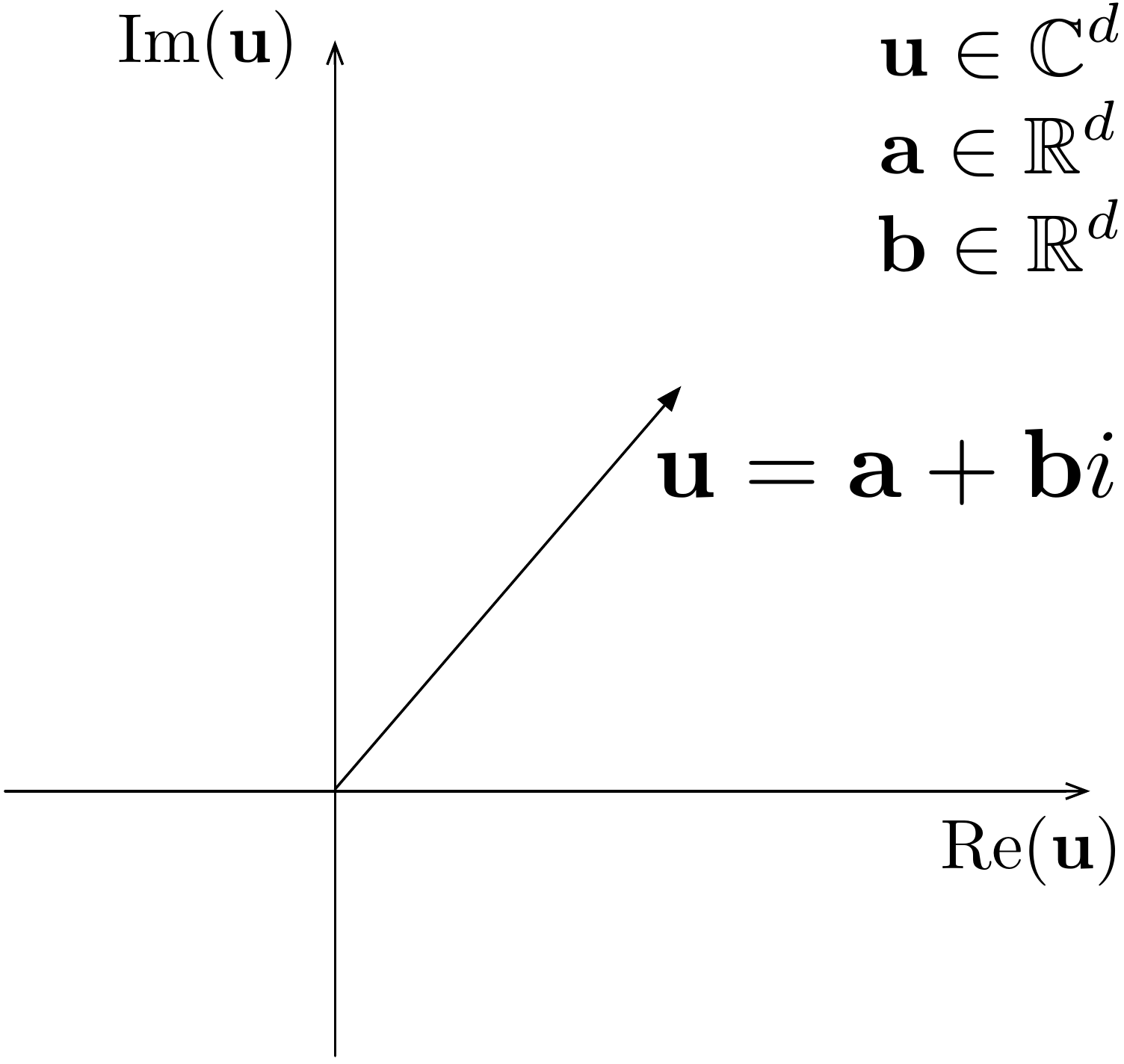}
 \caption{Complex vector space.}
 \label{fig:space-complex}
\end{subfigure}
\quad
 \begin{subfigure}[b]{0.3\textwidth}
 \centering
 \includegraphics[height=3.6cm]{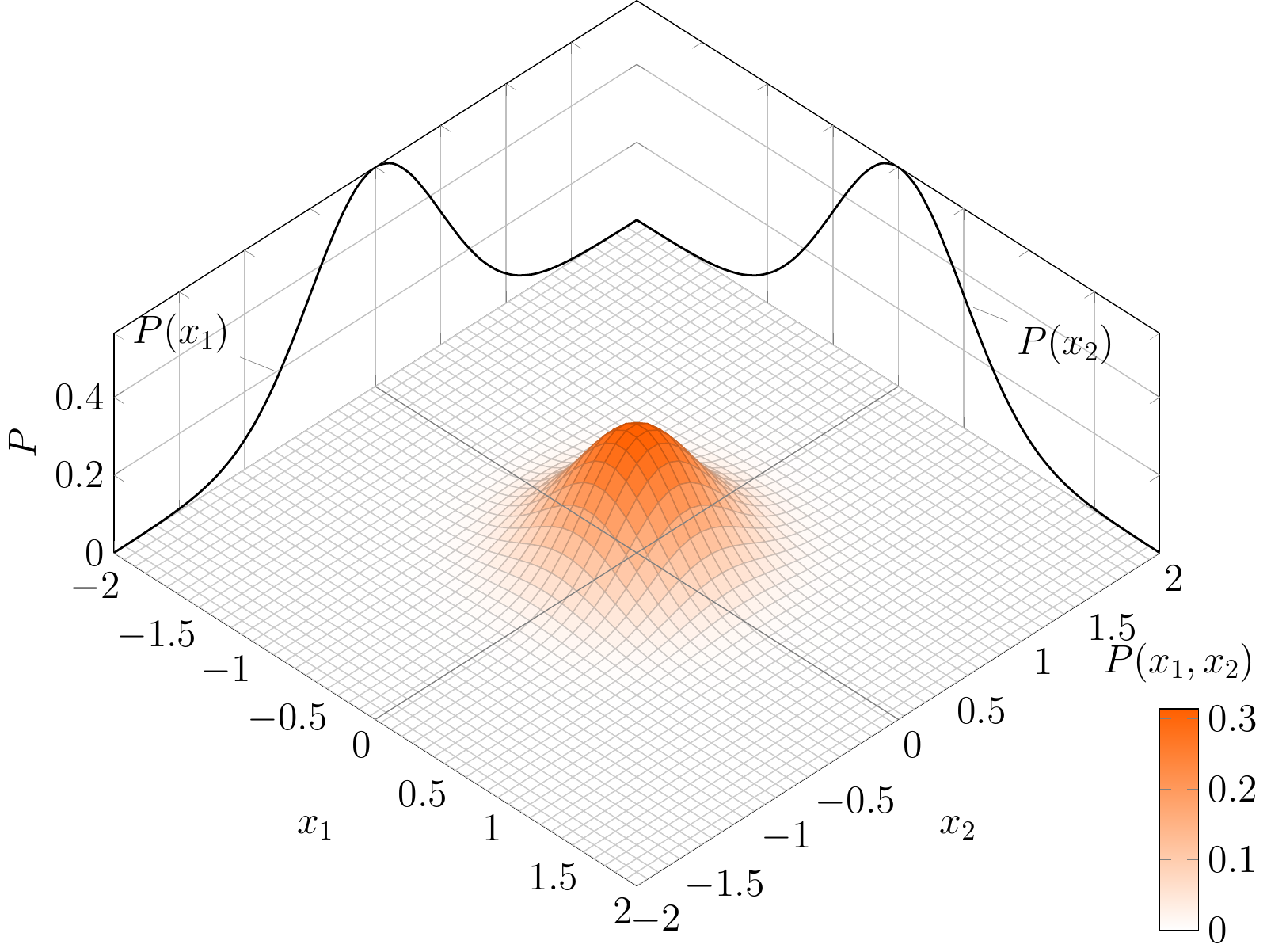}
 \caption{Gaussian distribution.}
 \label{fig:space-gaussian}
\end{subfigure}
\quad
\begin{subfigure}[b]{0.22\textwidth}
 \centering
 \includegraphics[height=3.6cm]{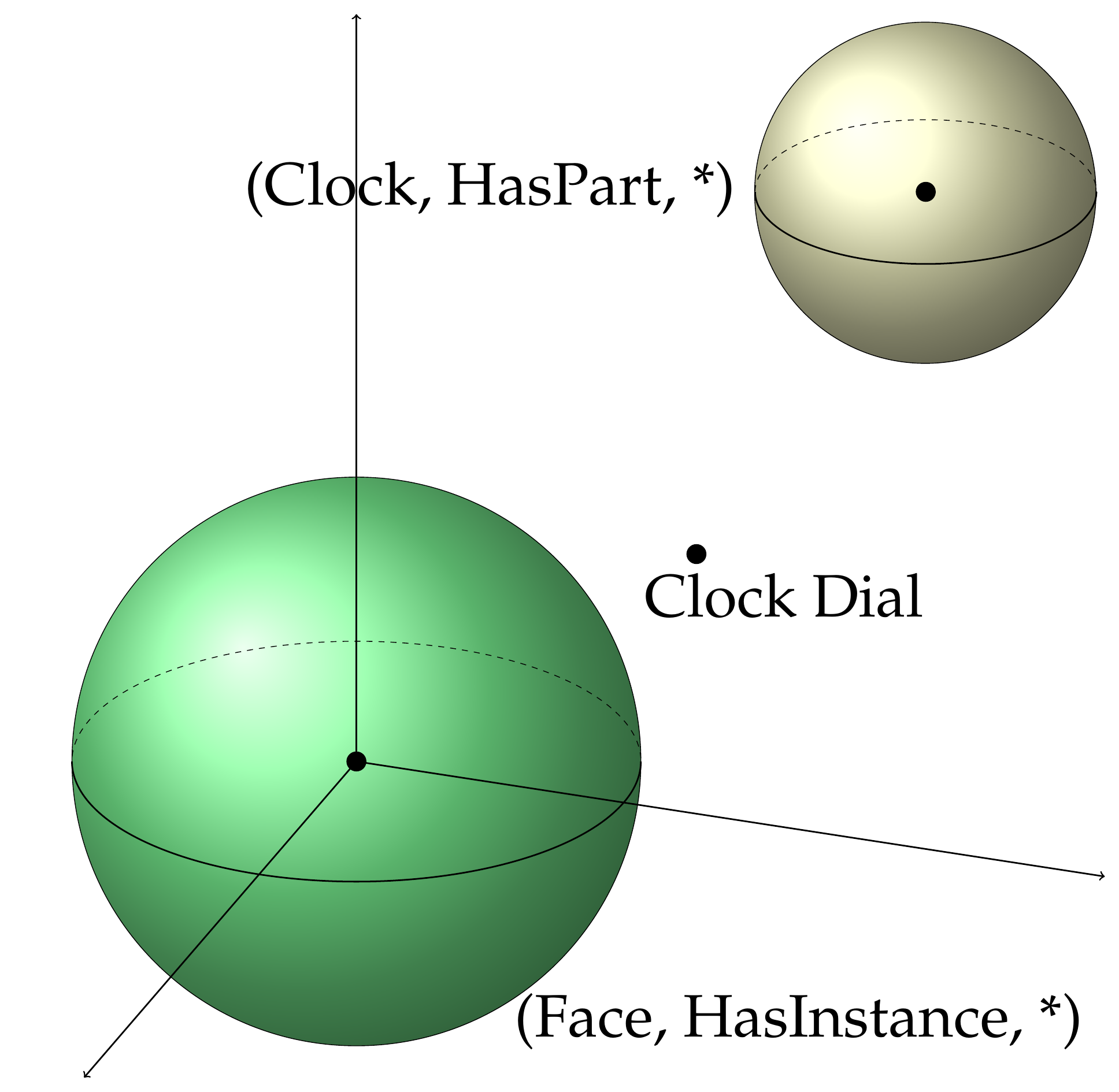}
 \caption{Manifold space.}
 \label{fig:space-manifold}
\end{subfigure}
\caption{An illustration of knowledge representation in different spaces.}
\label{fig:space}
\end{center}
\end{figure*}

The key issue of representation learning is to learn low-dimensional distributed embedding of entities and relations. 
Current literature mainly uses real-valued point-wise space (Fig.~\ref{fig:space-point}) including vector, matrix and tensor space, while other kinds of space such as complex vector space (Fig.~\ref{fig:space-complex}), Gaussian space (Fig.~\ref{fig:space-gaussian}), and manifold (Fig.~\ref{fig:space-manifold}) are utilized as well. 
The embedding space should follow three conditions, i.e., differentiability, calculation possibility, and definability of a scoring function~\cite{ebisu2018toruse}.

\subsubsection{Point-Wise~Space}
Point-wise Euclidean space is widely applied for representing entities and relations, projecting relation embedding in vector or matrix space, or capturing relational interactions.
TransE~\cite{bordes2013translating} represents entities and relations in $d$-dimension vector space, i.e., $\mathbf{h},\mathbf{t}, \mathbf{r} \in \mathbb{R}^d$, and makes embeddings follow the translational principle $\mathbf{h}+ \mathbf{r} \approx \mathbf{t}$.
To tackle this problem of insufficiency of a single space for both entities and relations, TransR~\cite{lin2015learning} then further introduces separated spaces for entities and relations. The authors projected entities ($\mathbf{h}, \mathbf{t} \in \mathbb{R}^{k}$) into relation ($\mathbf{r} \in \mathbb{R}^{d}$) space by a projection matrix $\mathbf{M_r} \in \mathbb{R}^{k\times d} $.
NTN~\cite{socher2013reasoning} models entities across multiple dimensions by a bilinear tensor neural layer.
The relational interaction between head and tail $\mathbf{h}^T\mathbf{\widehat{M}}\mathbf{t}$ is captured as a tensor denoted as $\mathbf{\widehat{M}}\in \mathbb{R}^{d\times d \times k}$. 
Instead of using the Cartesian coordinate system, HAKE~\cite{zhang2020learning} captures semantic hierarchies by mapping entities into the polar coordinate system, i.e., entity embeddings $\mathbf{e}_m \in \mathbb{R}^d$ and $\mathbf{e}_{p} \in[0,2 \pi)^{d}$ in the modulus and phase part, respectively. 

Many other translational models such as TransH~\cite{wang2014knowledge} also use similar representation space, while semantic matching models use plain vector space (e.g., HolE~\cite{nickel2016holographic}) and relational projection matrix (e.g., ANALOGY~\cite{liu2017analogical}).
Principles of these translational and semantic matching models are introduced in Section~\ref{sec:distance} and~\ref{sec:semantic}, respectively. 

\subsubsection{Complex~Vector~Space}
\label{sec:space-complex}
Instead of using a real-valued space, entities and relations are represented in a complex space, where $\mathbf{h}, \mathbf{t}, \mathbf{r} \in \mathbb{C}^d$. 
Take head entity as an example, $\mathbf{h}$ has a real part $\operatorname{Re}(\mathbf{h})$ and an imaginary part $\operatorname{Im}(\mathbf{h})$, i.e., $\mathbf{h} = \operatorname{Re}(\mathbf{h})+i \operatorname{Im}(\mathbf{h})$. 
ComplEx~\cite{trouillon2016complex} firstly introduces complex vector space shown in Fig.~\ref{fig:space-complex} which can capture both symmetric and antisymmetric relations. 
Hermitian dot product is used to do composition for relation, head and the conjugate of tail.
Inspired by Euler's identity $e^{i \theta}=\cos \theta+i \sin \theta$, RotatE~\cite{sun2018rotate} proposes a rotational model taking relation as a rotation from head entity to tail entity in complex space as $\mathbf{t}=\mathbf{h} \circ \mathbf{r}$ where $\circ$ denotes the element-wise Hadmard product. 
QuatE~\cite{zhang2019quaternion} extends the complex-valued space into hypercomplex $\mathbf{h}, \mathbf{t}, \mathbf{r} \in \mathbb{H}^d$ by a quaternion $Q=a+b \mathbf{i}+c \mathbf{j}+d \mathbf{k}$ with three imaginary components, where the quaternion inner product, i.e., the Hamilton product $\mathbf{h} \otimes \mathbf{r}$, is used as compositional operator for head entity and relation.
With the introduction of the rotational Hadmard product in complex space, RotatE~\cite{sun2018rotate} can also capture inversion and composition patterns as well as symmetry and antisymmetry. 
QuatE~\cite{zhang2019quaternion} uses Hamilton product to capture latent inter-dependencies within the four-dimensional space of entities and relations and gains a more expressive rotational capability than RotatE.

\subsubsection{Gaussian~Distribution}
Inspired by Gaussian word embedding, the density-based embedding model KG2E~\cite{he2015learning} introduces Gaussian distribution to deal with the (un)certainties of entities and relations.
The authors embedded entities and relations into multi-dimensional Gaussian distribution $\mathcal{H} \sim \mathcal{N}\left(\boldsymbol{\mu}_h, \mathbf{\Sigma}_h\right)$ and $\mathcal{T} \sim \mathcal{N}\left(\boldsymbol{\mu}_t, \mathbf{\Sigma}_t\right)$. The mean vector $\mathbf{u}$ indicates entities and relations' position, and the covariance matrix $\mathbf{\Sigma}$ models their (un)certainties. Following the translational principle, the probability distribution of entity transformation $\mathcal{H}-\mathcal{T}$ is denoted as $\mathcal{P}_{e} \sim \mathcal{N}\left(\boldsymbol{\mu}_{h}-\boldsymbol{\mu}_{t}, \mathbf{\Sigma}_{h}+\mathbf{\Sigma}_{t}\right)$. 
Similarly, TransG~\cite{xiao2016transg} represents entities with Gaussian distributions, while it draws a mixture of Gaussian distribution for relation embedding, where the $m$-th component translation vector of relation $r$ is denoted as $\mathbf{u}_{r, m}=\mathbf{t}-\mathbf{h} \sim \mathcal{N}\left(\mathbf{u}_{\mathbf{t}}-\mathbf{u}_{\mathbf{h}},\left(\sigma_{h}^{2}+\sigma_{t}^{2}\right) \mathbf{E}\right)$.

\subsubsection{Manifold and Group}
\label{sec:manifold}
This section reviews knowledge representation in manifold space, Lie group, and dihedral group. A manifold is a topological space, which could be defined as a set of points with neighborhoods by the set theory. The group is algebraic structures defined in abstract algebra.
Previous point-wise modeling is an ill-posed algebraic system where the number of scoring equations is far more than the number of entities and relations. Moreover, embeddings are restricted in an overstrict geometric form even in some methods with subspace projection. 
To tackle these issues, ManifoldE~\cite{xiao2016one} extends point-wise embedding into manifold-based embedding. 
The authors introduced two settings of manifold-based embedding, i.e., Sphere and Hyperplane. An example of a sphere is shown in Fig.~\ref{fig:space-manifold}.
For the sphere setting, Reproducing Kernel Hilbert Space is used to represent the manifold function. 
Another ``hyperplane'' setting is introduced to enhance the model with intersected embeddings.
ManifoldE~\cite{xiao2016one} relaxes the real-valued point-wise space into manifold space with a more expressive representation from the geometric perspective. When the manifold function and relation-specific manifold parameter are set to zero, the manifold collapses into a point.

Hyperbolic space, a multidimensional Riemannian manifold with a constant negative curvature $-c~(c>0): \mathbb{B}^{d, c}=\left\{\mathbf{x} \in \mathbb{R}^{d}:\|\mathbf{x}\|^{2}<\frac{1}{c}\right\}$, is drawing attention for its capacity of capturing hierarchical information. 
MuRP~\cite{balazevic2019multi} represents the multi-relational knowledge graph in Poincaré ball of hyperbolic space $\mathbb{B}_{c}^{d}=\left\{\mathbf{x} \in \mathbb{R}^{d}: c\|\mathbf{x}\|^{2}<1\right\}$.
While it fails to capture logical patterns and suffers from constant curvature. Chami et al.~\cite{chami2020low} leverages expressive hyperbolic isometries and learns a relation-specific absolute curvature $c_r$ in the hyperbolic space.

TorusE~\cite{ebisu2018toruse} solves the regularization problem of TransE via embedding in an n-dimensional torus space which is a compact Lie group. With the projection from vector space into torus space defined as $\pi : \mathbb{R}^{n} \rightarrow T^{n}, x \mapsto[x]$, entities and relations are denoted as $[\mathbf{h}],[\mathbf{r}],[\mathbf{t}] \in \mathbb{T}^{n}$. Similar to TransE, it also learns embeddings following the relational translation in torus space, i.e., $[\mathbf{h}]+[\mathbf{r}] \approx [\mathbf{t}]$.
Recently, DihEdral~\cite{xu2019relation} proposes a dihedral symmetry group preserving a 2-dimensional polygon. 
It utilizes a finite non-Abelian group to preserve the relational properties of symmetry/skew-symmetry, inversion, and composition effectively with the rotation and reflection properties in the dihedral group.

\subsection{Scoring~Function}\label{sec:function}
\begin{figure}[htbp]
\begin{center}
\begin{subfigure}[b]{0.21\textwidth}
\includegraphics[width=\textwidth]{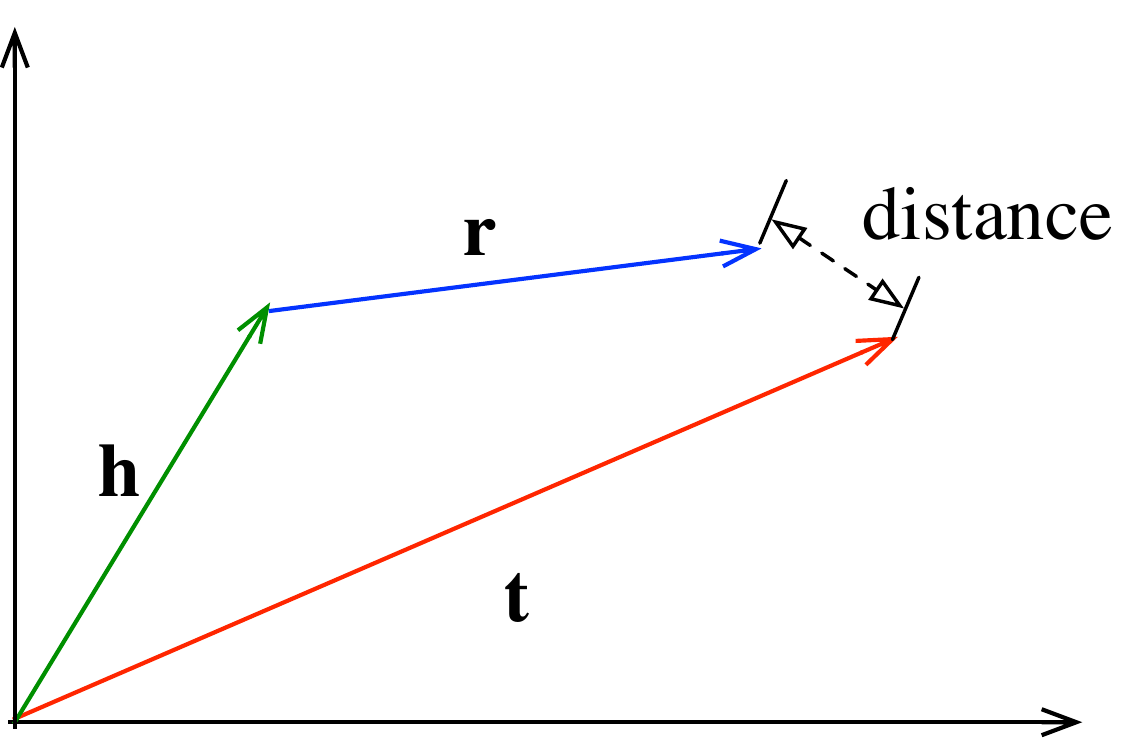}
\caption{Translational distance-based scoring of TransE.}
\label{fig:distance}
\end{subfigure}
\quad
\begin{subfigure}[b]{0.21\textwidth}
\includegraphics[width=\textwidth]{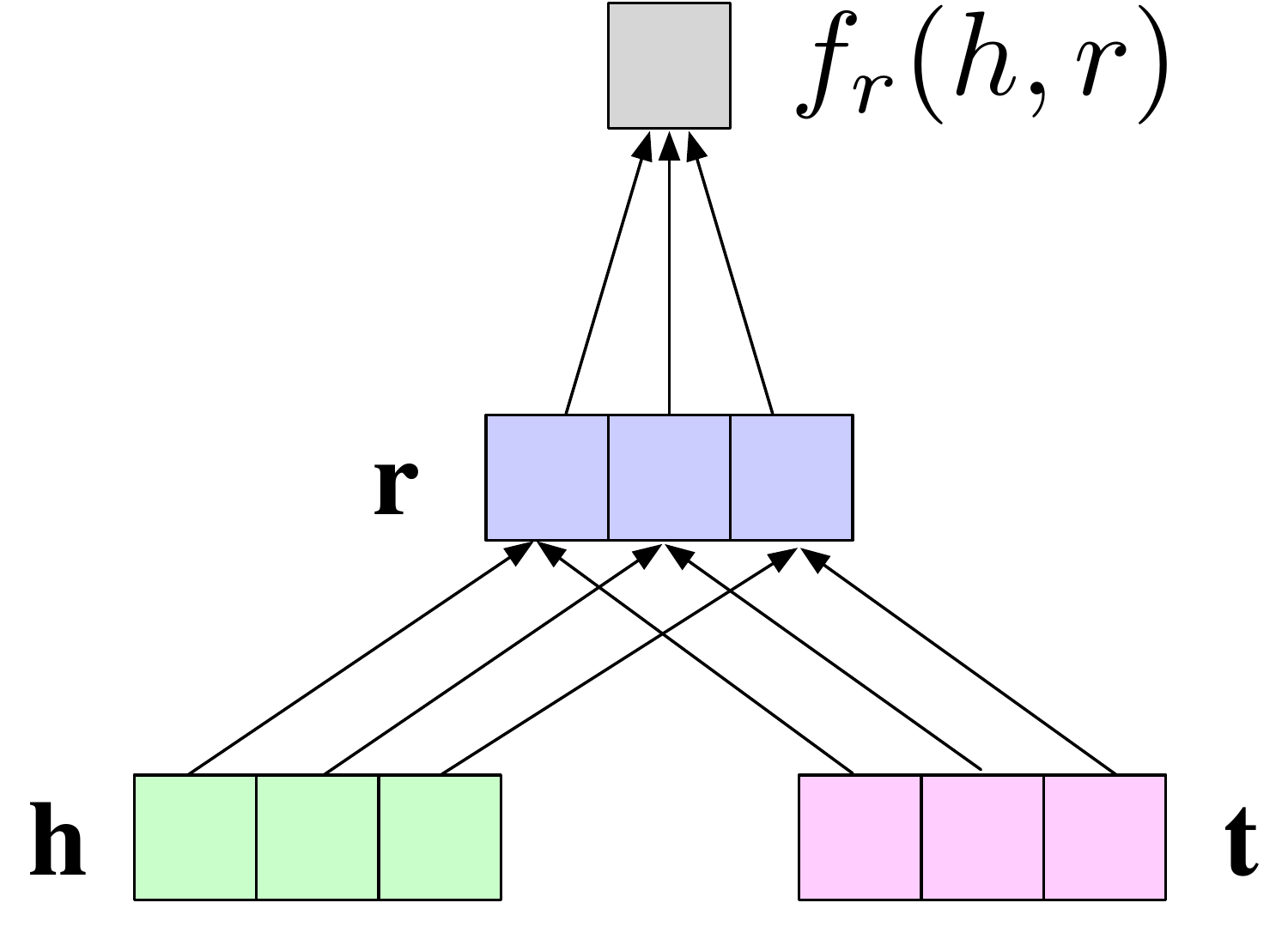}
\caption{Semantic similarity-based scoring of DistMult.}
\label{fig:semantic}
\end{subfigure}
\caption{Illustrations of distance-based and similarity matching based scoring functions taking TransE\cite{bordes2013translating} and DistMult\cite{yang2014embedding} as examples.}
\label{fig:scoring}
\end{center}
\end{figure}

The scoring function is used to measure the plausibility of facts, also referred to as the energy function in the energy-based learning framework. Energy-based learning aims to learn the energy function $\mathcal{E}_\theta(x)$ (parameterized by $\theta$ taking $x$ as input) and to make sure positive samples have higher scores than negative samples. 
In this paper, the term of the scoring function is adopted for unification. 
There are two typical types of scoring functions, i.e., distance-based (Fig.~\ref{fig:distance}) and similarity-based (Fig.~\ref{fig:semantic}) functions, to measure the plausibility of a fact. 
Distance-based scoring function measures the plausibility of facts by calculating the distance between entities, where addictive translation with relations as $\mathbf{h} + \mathbf{r} \approx \mathbf{t}$ is widely used.
Semantic similarity based scoring measures the plausibility of facts by semantic matching. It usually adopts a multiplicative formulation, i.e., $\mathbf{h}^{\top} \mathbf{M}_{r} \approx \mathbf{t}^{\top}$, to transform head entity near the tail in the representation space.

\subsubsection{Distance-based Scoring Function}
\label{sec:distance}
An intuitive distance-based approach is to calculate the Euclidean distance between the relational projection of entities. 
Structural Embedding (SE)~\cite{bordes2011learning} uses two projection matrices and $L_1$ distance to learn structural embedding as
\begin{equation}\small \label{eq:SE}
f_r(h,t) = \| \mathbf{M}_{r,1}\mathbf{h} - \mathbf{M}_{r,2}\mathbf{t} \|_{L_1}.
\end{equation}
A more intensively used principle is the translation-based scoring function that aims to learn embeddings by representing relations as translations from head to tail entities. 
Bordes et al.~\cite{bordes2013translating} proposed TransE by assuming that the added embedding of $\textbf{h}+ \textbf{r}$ should be close to the embedding of $\textbf{t}$ with the scoring function is defined under $L_1$ or $L_2$ constraints as
\begin{equation}\small
\label{eq:TransE}
f_r(h,t) = \|\mathbf{h} + \mathbf{r}-\mathbf{t}\|_{L_1 \slash L_2}.
\end{equation}
Since that, many variants and extensions of TransE have been proposed. For example, 
TransH~\cite{wang2014knowledge} projects entities and relations into a hyperplane,
TransR~\cite{lin2015learning} introduces separate projection spaces for entities and relations, 
and TransD~\cite{ji2015knowledge} constructs dynamic mapping matrices $\mathbf{M}_{rh}=\mathbf{r}_{p} \mathbf{h}_{p}^{\top}+\mathbf{I}$ and $\mathbf{M}_{rt}=\mathbf{r}_{p} \mathbf{t}_{p}^{\top}+\mathbf{I}$ by the projection vectors $\mathbf{h}_p,\mathbf{t}_p,\mathbf{r}_p\in \mathbb{R}^n$.
By replacing Euclidean distance, TransA~\cite{xiao2015transa} uses Mahalanobis distance to enable more adaptive metric learning.
Previous methods used additive score functions, TransF~\cite{feng2016knowledge} relaxes the strict translation and uses dot product as $f_{r}(h, t)=(\mathbf{h}+\mathbf{r})^{\top} \mathbf{t}$.
To balance the constraints on head and tail, a flexible translation scoring function is further proposed.

Recently, ITransF~\cite{xie2017interpretable} enables hidden concepts discovery and statistical strength transferring by learning associations between relations and concepts via sparse attention vectors, with scoring function defined as 
 \begin{equation}\small
 \label{eq:ITransF}
 f_{r}(h, t)=\left\|\boldsymbol{\alpha}_{r}^{H} \cdot \mathbf{D} \cdot \mathbf{h}+\mathbf{r}-\boldsymbol{\alpha}_{r}^{T} \cdot \mathbf{D} \cdot \mathbf{t}\right\|_{\ell},
 \end{equation}
 where $\mathbf{D}\in \mathbb{R}^{n\times d \times d}$ is stacked concept projection matrices of entities and relations and $\boldsymbol{\alpha}_{r}^{H}, \boldsymbol{\alpha}_{r}^{T} \in[0,1]^{n}$ are attention vectors calculated by sparse softmax,
TransAt~\cite{qian2018translating} integrates relation attention mechanism with translational embedding, and TransMS~\cite{yang2019transms} transmits multi-directional semantics with nonlinear functions and linear bias vectors, with the scoring function as
\begin{equation}\small
\label{eq:TransMS}
 f_{r}(\mathbf{h}, \mathbf{t})=\|-\tanh (\mathbf{t} \circ \mathbf{r}) \circ \mathbf{h}+\mathbf{r} -\tanh (\mathbf{h} \circ \mathbf{r}) \circ \mathbf{t} +\alpha \cdot(\mathbf{h} \circ \mathbf{t})\|_{\ell_{1 / 2}}.
\end{equation}

KG2E~\cite{he2015learning} in Gaussian space and ManifoldE~\cite{xiao2016one} with manifold also use the translational distance-based scoring function.
KG2E uses two scoring methods, i.e, asymmetric KL-divergence 
and symmetric expected likelihood.
While the scoring function of ManifoldE is defined as 
\begin{equation}\small
\label{eq:ManifoldE}
f_{r}(h, t)=\left\|\mathcal{M}(h, r, t)-D_{r}^{2}\right\|^{2},
\end{equation}
where $\mathcal{M}$ is the manifold function, and $D_r$ is a relation-specific manifold parameter.

\subsubsection{Semantic~Matching}
\label{sec:semantic}
Another direction is to calculate the semantic similarity. 
SME~\cite{bordes2014semantic} proposes to semantically match separate combinations of entity-relation pairs of $(h, r)$ and $(r, t)$. Its scoring function is defined with two versions of matching blocks - linear and bilinear block, i.e.,
\begin{equation}\small
\label{eq:SME}
f_r(h,t) = g_\text{left}(\mathbf{h}, \mathbf{r})^\top g_\text{right}(\mathbf{r}, \mathbf{t}).
\end{equation}
The linear matching block is defined as $g_\text{left}(h,t) = \mathbf{M}_{l,1}\mathbf{h}^\top + \mathbf{M}_{l,2} \mathbf{r}^\top + \mathbf{b}_l^\top$, and the bilinear form is $g_\text{left}(\mathbf{h}, \mathbf{r})=\left(\mathbf{M}_{l,1} \mathbf{h}\right) \circ\left(\mathbf{M}_{l,2} \mathbf{r}\right)+\mathbf{b}_{l}^\top$. 
By restricting relation matrix $M_r$ to be diagonal for multi-relational representation learning, DistMult~\cite{yang2014embedding} proposes a simplified bilinear formulation defined as
\begin{equation}\small
\label{eq:DistMult}
f_r(h, t)= \mathbf{h}^{\top} \operatorname{diag}(\mathbf{M}_{r}) \mathbf{t}.
\end{equation}

To capture productive interactions in relational data and compute efficiently, 
HolE~\cite{nickel2016holographic} introduces a circular correlation of embedding, which can be interpreted as a compressed tensor product, to learn compositional representations.
By defining a perturbed holographic compositional operator as 
$p(\boldsymbol{a}, \boldsymbol{b} ; \boldsymbol{c})=(\boldsymbol{c} \circ \boldsymbol{a}) \star \boldsymbol{b}$,
where $\mathbf{c}$ is a fixed vector, the expanded holographic embedding model HolEx~\cite{xue2018expanding} interpolates the HolE and full tensor product method.
It can be viewed as linear concatenation of perturbed HolE.
Focusing on multi-relational inference, ANALOGY~\cite{liu2017analogical} models analogical structures of relational data. 
It's scoring function is defined as
\begin{equation}\small\label{eq:ANALOGY}
f_r(h, t)= \mathbf{h}^{\top} \mathbf{M}_{r} \mathbf{t},
\end{equation}
with relation matrix constrained to be normal matrices in linear mapping, i.e., $\mathbf{M}_r^\top\mathbf{M}_r = \mathbf{M}_r \mathbf{M}_r^\top$ for analogical inference. 
HolE with Fourier transformed in the frequency domain can be viewed as a special case of ComplEx~\cite{hayashi2017equivalence}, which connects holographic and complex embeddings. The analogical embedding framework~\cite{liu2017analogical} can recover or equivalently obtain several models such as DistMult, ComplEx and HolE by restricting the embedding dimension and scoring function. 
Crossover interactions are introduced by CrossE~\cite{zhang2019interaction} with an interaction matrix $\mathbf{C}\in\mathbb{R}^{n_r\times d}$ to simulate the bi-directional interaction between entity and relation. 
The relation specific interaction is obtained by looking up interaction matrix as $\mathbf{c}_{r}=\mathbf{x}_{r}^{\top} \mathbf{C}$.
By combining the interactive representations and matching with tail embedding, the scoring function is defined as 
\begin{equation}\small
f(h, r, t)=\sigma\left(\tanh \left(\mathbf{c}_{r} \circ \mathbf{h}+\mathbf{c}_{r} \circ \mathbf{h} \circ \mathbf{r}+\mathbf{b}\right) \mathbf{t}^{\top}\right).
\end{equation}
The semantic matching principle can be encoded by neural networks further discussed in Sec.~\ref{sec:encoding}.

The two methods mentioned above in Sec.~\ref{sec:manifold} with group representation also follow the semantic matching principle.
The scoring function of TorusE~\cite{ebisu2018toruse} is defined as:
\begin{equation}\small
\min _{(x, y) \in([h]+[r]) \times[t]}\|x-y\|_{i}.
\end{equation}
By modeling $2L$ relations as group elements, the scoring function of DihEdral~\cite{xu2019relation} is defined as the summation of components:
\begin{equation}\small
f_r(h,t)=\mathbf{h}^{\top} \mathbf{R} \mathbf{t}=\sum_{l=1}^{L} \mathbf{h}^{(l) \top} \mathbf{R}^{(l)} \mathbf{t}^{(l)},
\end{equation}
where the relation matrix $\mathbf{R}$ is defined in block diagonal form for $\mathbf{R}^{(l)} \in \mathbb{D}_{K}$, and entities are embedded in real-valued space for $\mathbf{h}^{(l)}$ and $\mathbf{t}^{(l)} \in \mathbb{R}^2$.

\subsection{Encoding~Models}
\label{sec:encoding}
\begin{figure*}[htbp]
\begin{center}
\begin{subfigure}[b]{0.15\textwidth}
\centering
\includegraphics[height=2.8cm]{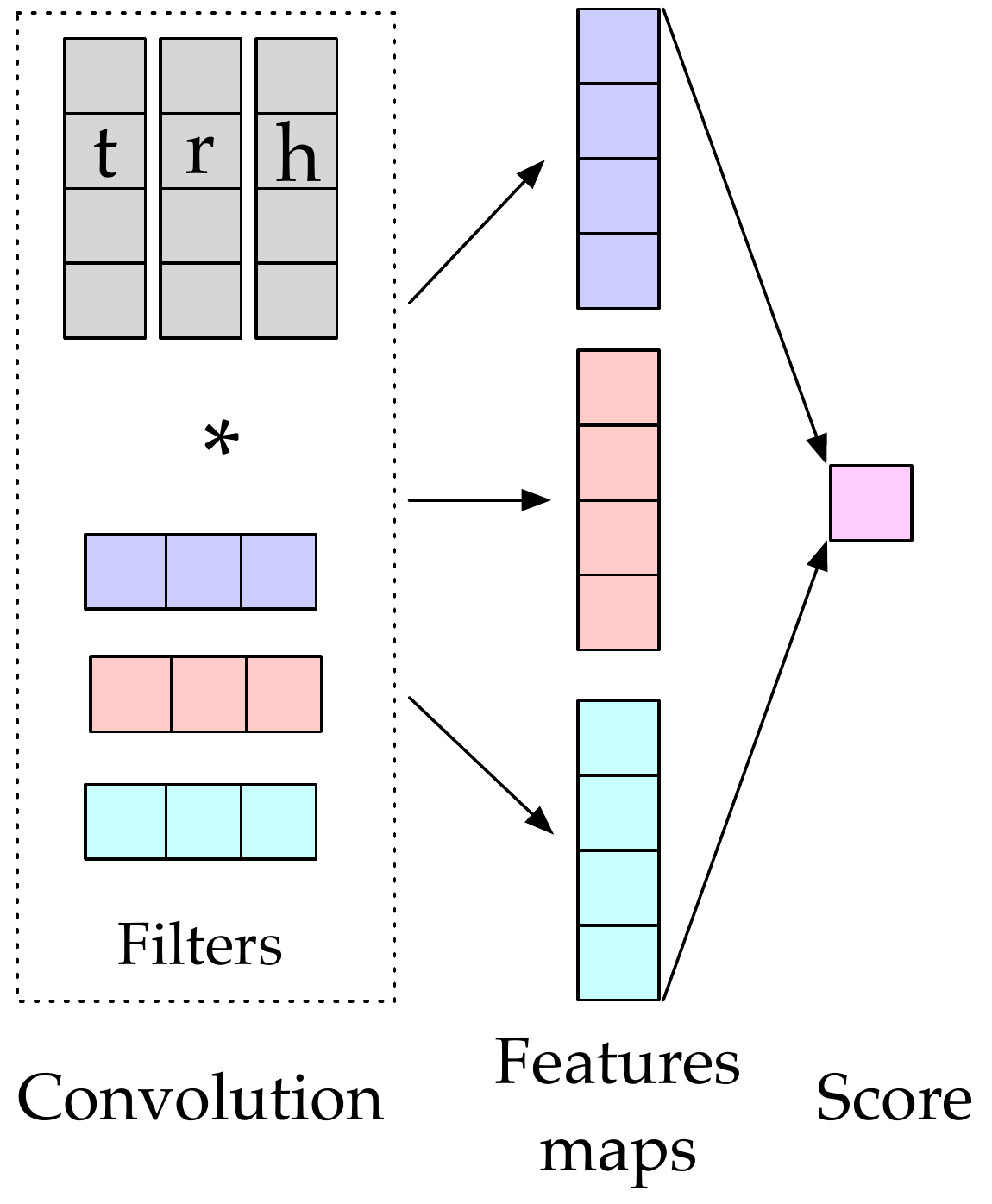}
\caption{CNN.}
\label{fig:CNN}
\end{subfigure}
\quad
\begin{subfigure}[b]{0.2\textwidth}
\centering
\includegraphics[height=2.8cm]{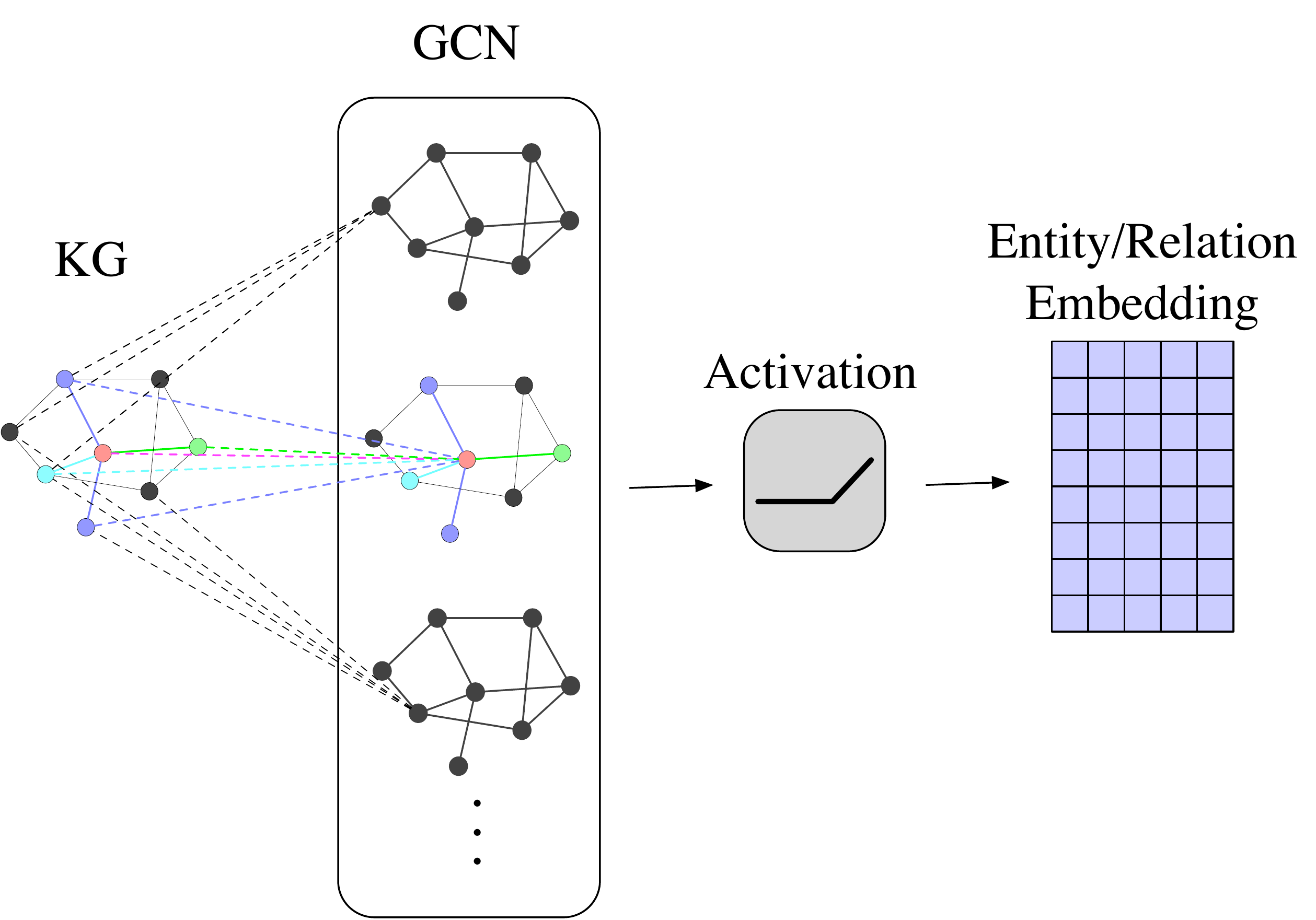}
\caption{GCN.}
\label{fig:SACN}
\end{subfigure}
\quad
\begin{subfigure}[b]{0.25\textwidth}
\centering
\includegraphics[height=2.6cm]{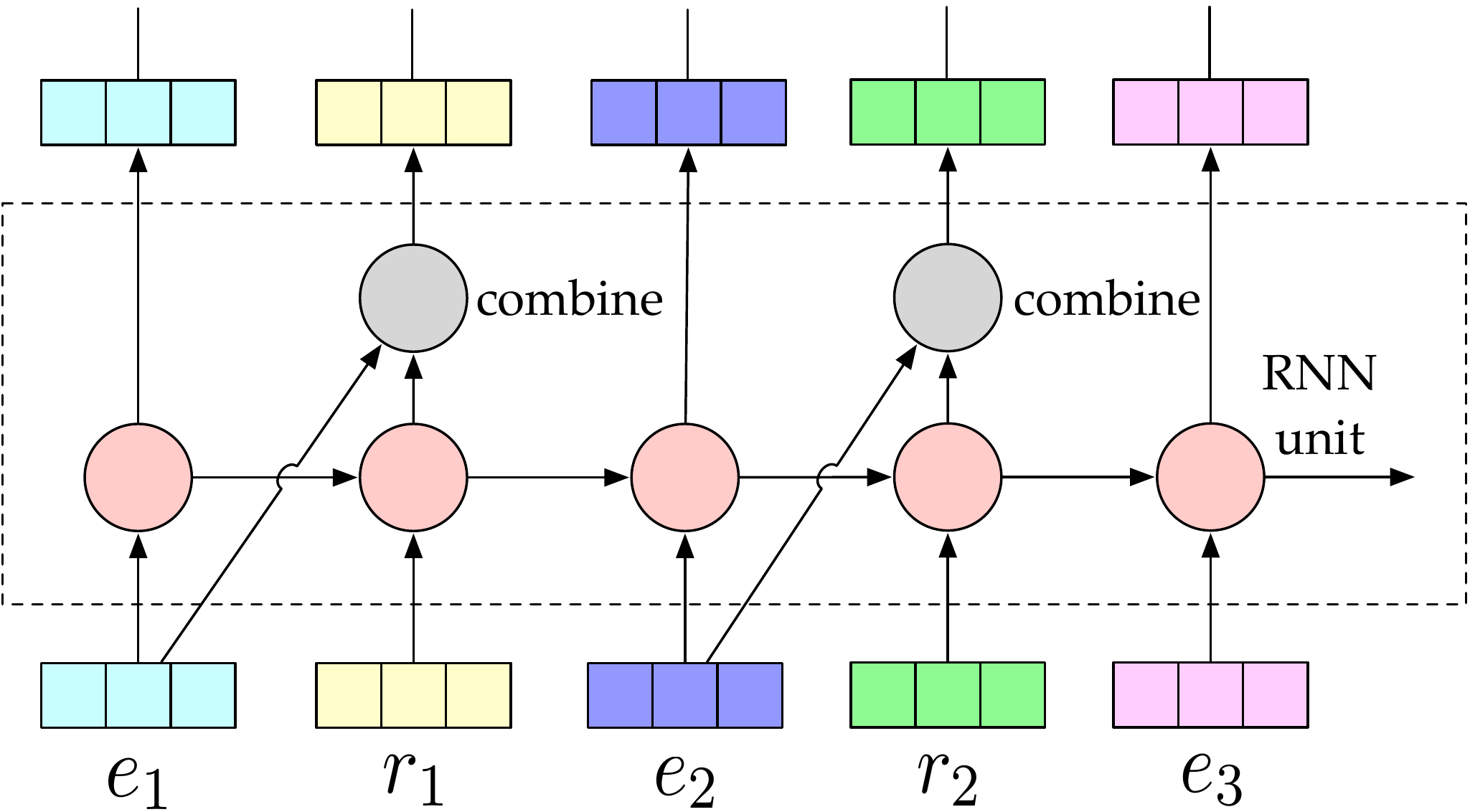}
\caption{RSN.}
\label{fig:RSN}
\end{subfigure}
\quad
\begin{subfigure}[b]{0.3\textwidth}
\centering
\includegraphics[height=2.6cm]{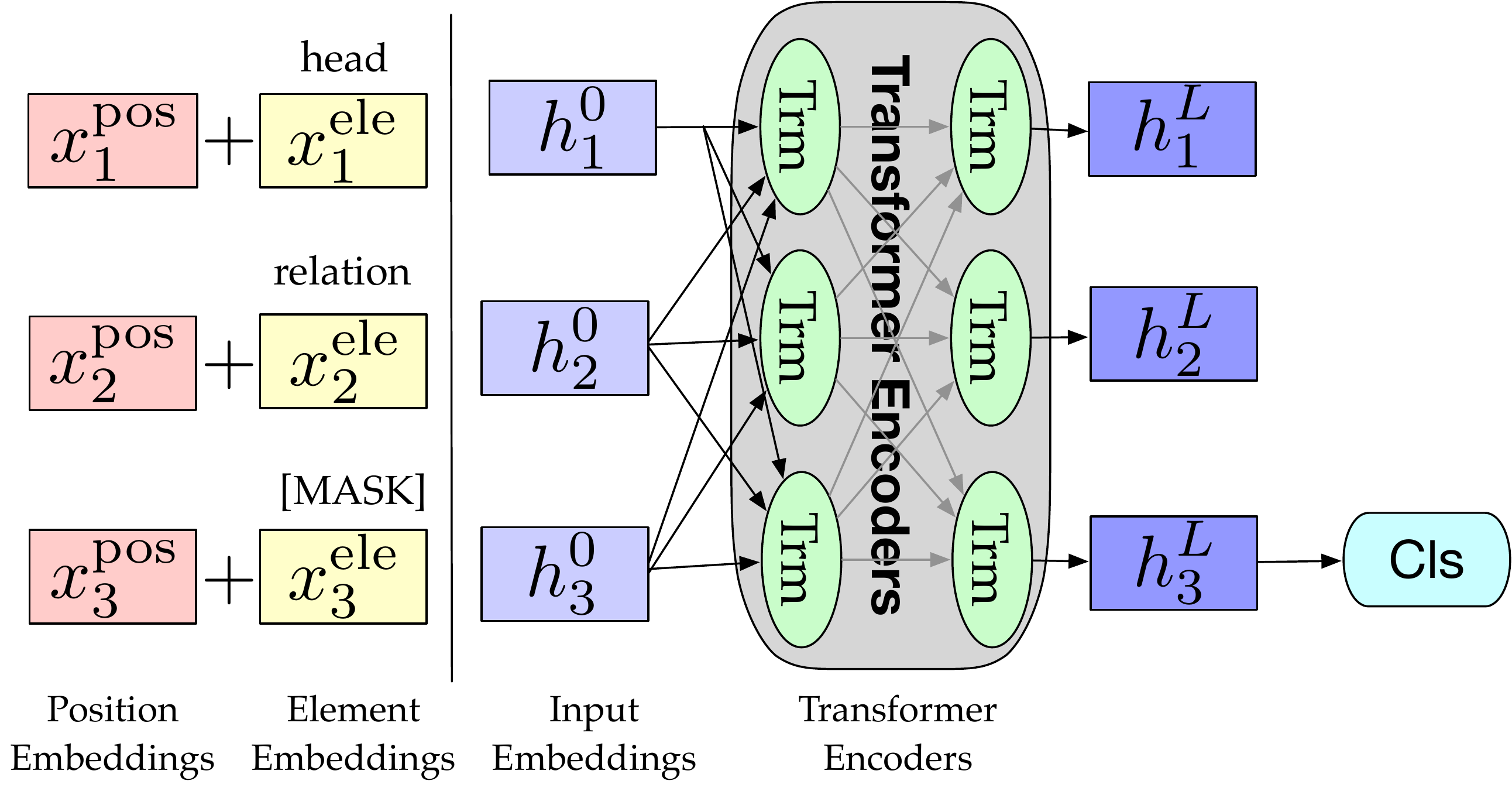}
\caption{Transformer.}
\label{fig:transformer}
\end{subfigure}
\caption{Illustrations of neural encoding models. (a) CNN~\cite{nguyen2017novel} input triples into dense layer and convolution operation to learn semantic representation, (b) GCN~\cite{shang2018end} acts as encoder of knowledge graphs to produce entity and relation embeddings. (c) RSN~\cite{guo2019learning} encodes entity-relation sequences and skips relations discriminatively. (d) Transformer-based CoKE~\cite{wang2019coke} encodes triples as sequences with an entity replaced by \texttt{[MASK]}.}
\label{fig:models}
\end{center}
\end{figure*}

This section introduces models that encode the interactions of entities and relations through specific model architectures, including linear/bilinear models, factorization models, and neural networks. Linear models formulate relations as a linear/bilinear mapping by projecting head entities into a representation space close to tail entities. Factorization aims to decompose relational data into low-rank matrices for representation learning. Neural networks encode relational data with non-linear neural activation and more complex network structures by matching semantic similarity of entities and relations.
Several neural models are illustrated in Fig.~\ref{fig:models}. 

\subsubsection{Linear/Bilinear~Models}
Linear/bilinear models encode interactions of entities and relations by applying linear operation as:
\begin{equation}\small
\label{eq:linear}
g_{r}\left(\mathbf{h}, \mathbf{t}\right)=\mathbf{M}_{r}^{T}\left(\begin{array}{c}{\mathbf{h}} \\ {\mathbf{t}}\end{array}\right),
\end{equation}
or bilinear transformation operations as Eq.~\ref{eq:ANALOGY}.
Canonical methods with linear/bilinear encoding include SE~\cite{bordes2011learning}, SME~\cite{bordes2014semantic}, DistMult~\cite{yang2014embedding}, ComplEx~\cite{trouillon2016complex}, and ANALOGY~\cite{liu2017analogical}.
For TransE~\cite{bordes2013translating} with L2 regularization, the scoring function can be expanded to the form with only linear transformation with one-dimensional vectors, i.e.,
\begin{equation}\small
\label{eq:expansion}
\left\|\mathbf{h}+\mathbf{r}-\mathbf{t}\right\|_{2}^{2}=2 \mathbf{r}^{T}\left(\mathbf{h}-\mathbf{t}\right)-2 \mathbf{h}^{T} \mathbf{t}+\left\| \mathbf{r}\right\|_{2}^{2}+\left\|\mathbf{h}\right\|_{2}^{2}+\left\|\mathbf{t}\right\|_{2}^{2}.
\end{equation}
Wang et al.~\cite{wang2018multi} studied various bilinear models and evaluated their expressiveness and connections by introducing the concepts of universality and consistency. The authors further showed that the ensembles of multiple linear models can improve the prediction performance through experiments. Recently, to solve the independence embedding issue of entity vectors in canonical Polyadia decomposition,
SimplE~\cite{kazemi2018simple} introduces the inverse of relations and calculates the average canonical Polyadia score of $(h, r, t)$ and $(t, r^{-1}, h)$ as
\begin{equation}\small
\label{eq:SimplE}
f_r(h,t)=\frac{1}{2}\left( \mathbf{h} \circ \mathbf{r} \mathbf{t}+ \mathbf{t}\circ \mathbf{r}^{\prime} \mathbf{t}\right),
\end{equation}
where $\mathbf{r}^\prime$ is the embedding of inversion relation.
Embedding models in the bilinear family such as RESCAL, DistMult, HolE and ComplEx can be transformed from one into another with certain constraints ~\cite{wang2018multi}. 
More bilinear models are proposed from a factorization perspective discussed in the next section.

\subsubsection{Factorization~Models}
\label{sec:factor}
Factorization methods formulated KRL models as three-way tensor $\mathcal{X}$ decomposition. A general principle of tensor factorization can be denoted as $\mathcal{X}_{hrt}\approx \mathbf{h}^\top \mathbf{M}_r \mathbf{t}$, with the composition function following the semantic matching pattern. 
Nickel et al.~\cite{nickel2011three} proposed the three-way rank-$r$ factorization RESCAL over each relational slice of knowledge graph tensor. 
For $k$-th relation of $m$ relations, the $k$-th slice of $\mathcal{X}$ is factorized as
\begin{equation}\small
\mathcal{X}_{k} \approx \mathbf{A} \mathbf{R}_{k} \mathbf{A}^{T}.
\end{equation}
The authors further extended it to handle attributes of entities efficiently~\cite{nickel2012factorizing}.
Jenatton et al.~\cite{jenatton2012latent} then proposed a bilinear structured latent factor model (LFM), which extends RESCAL by decomposing $\mathbf{R}_k=\sum_{i=1}^d \boldsymbol{\alpha}_i^k \mathbf{u}_i \mathbf{v}_i^\top$. 
By introducing three-way Tucker tensor decomposition, TuckER~\cite{balavzevic2019tucker} learns to embed by outputting a core tensor and embedding vectors of entities and relations. 
LowFER~\cite{amin2020lowfer} proposes a multi-modal factorized bilinear pooling mechanism to better fuse entities and relations.
It generalizes the TuckER model and is computationally efficient with low-rank approximation. 

\subsubsection{Neural~Networks}
Neural networks for encoding semantic matching have yielded remarkable predictive performance in recent studies. 
Encoding models with linear/bilinear blocks can also be modeled using neural networks, for example, SME~\cite{bordes2014semantic}. 
Representative neural models include multi-layer perceptron (MLP)~\cite{dong2014knowledge}, neural tensor network (NTN)~\cite{socher2013reasoning}, and neural association model (NAM)~\cite{liu2016probabilistic}. 
They generally feed entities or relations or both into deep neural networks and compute a semantic matching score.
MLP~\cite{dong2014knowledge} encodes entities and relations together into a fully-connected layer, and uses a second layer with sigmoid activation for scoring a triple as
\begin{equation}\small
\label{eq:MLP}
f_r(h, t)=\operatorname{\sigma}(\mathbf{w}^\top \operatorname{\sigma}(\mathbf{W}[\mathbf{h}, \mathbf{r}, \mathbf{t}])),
\end{equation}
where $\mathbf{W} \in \mathbb{R}^{n\times 3d}$ is the weight matrix and $[\mathbf{h}, \mathbf{r}, \mathbf{t}]$ is a concatenation of three vectors.
NTN~\cite{socher2013reasoning} takes entity embeddings as input associated with a relational tensor and outputs predictive score in as
\begin{equation}\small
\label{eq:NTN}
f_r(h,t) = \mathbf{r}^\top \sigma (\mathbf{h}^T\mathbf{\widehat{M}}\mathbf{t} + \mathbf{M}_{r,1}\mathbf{h} + \mathbf{M}_{r,2}\mathbf{t} + \mathbf{b}_r),
\end{equation}
where $\mathbf{b}_r\in \mathbb{R}^k$ is bias for relation $r$, $\mathbf{M}_{r,1}$ and $\mathbf{M}_{r,2}$ are relation-specific weight matrices. 
It can be regarded as a combination of MLPs and bilinear models. 
NAM~\cite{liu2016probabilistic} associates the hidden encoding with the embedding of tail entity, and proposes the relational-modulated neural network (RMNN).

\subsubsection{Convolutional~Neural~Networks}
CNNs are utilized for learning deep expressive features.
ConvE~\cite{dettmers2018convolutional} uses 2D convolution over embeddings and multiple layers of nonlinear features to model the interactions between entities and relations by reshaping head entity and relation into 2D matrix, i.e., $\mathbf{M}_h \in \mathbb{R}^{d_w\times d_h}$ and $\mathbf{M}_r \in \mathbb{R}^{d_w\times d_h}$ for $d=d_w\times d_h$. Its scoring function is defined as
\begin{equation}\small
f_{r}\left(h, t\right)=\sigma\left(\operatorname{vec}\left(\sigma\left(\left[\mathbf{M}_h ; \mathbf{M}_r\right] * \boldsymbol{\omega}\right)\right) \mathbf{W}\right) \mathbf{t},
\end{equation}
where $\boldsymbol{\omega}$ is the convolutional filters and $\operatorname{vec}$ is the vectorization operation reshaping a tensor into a vector. ConvE can express semantic information by non-linear feature learning through multiple layers. 
ConvKB~\cite{nguyen2017novel} adopts CNNs for encoding the concatenation of entities and relations without reshaping (Fig.~\ref{fig:CNN}). Its scoring function is defined as
\begin{equation}\small
f_r(h, t)=\operatorname{concat}\left(\sigma\left(\left[\boldsymbol{h}, \boldsymbol{r}, \boldsymbol{t}\right] * \boldsymbol{\omega}\right)\right) \cdot \mathbf{w}.
\end{equation}
The concatenation of a set for feature maps generated by convolution increases the learning ability of latent features. 
Compared with ConvE, which captures the local relationships, ConvKB keeps the transitional characteristic and shows better experimental performance. 
HypER~\cite{balavzevic2019hypernetwork} utilizes hypernetwork $\mathbf{H}$ for 1D relation-specific convolutional filter generation to achieve multi-task knowledge sharing, and meanwhile simplifies 2D ConvE. 
It can also be interpreted as a tensor factorization model when taking hypernetwork and weight matrix as tensors. 

\subsubsection{Recurrent~Neural~Networks}
The MLP- and CNN-based models, as mentioned above, learn triplet-level representations. In comparison, the recurrent networks can capture long-term relational dependencies in knowledge graphs. 
Gardner et al.~\cite{gardner2014incorporating} and Neelakantan et al.~\cite{neelakantan2015compositional} propose RNN-based model over the relation path to learn vector representation without and with entity information, respectively. 
RSN~\cite{guo2019learning} (Fig.~\ref{fig:RSN}) designs a recurrent skip mechanism to enhance semantic representation learning by distinguishing relations and entities. The relational path as $\left(x_{1}, x_{2}, \dots, x_{T}\right)$ with entities and relations in an alternating order is generated by random walk, and it is further used to calculate recurrent hidden state $\mathbf{h}_{t}=\tanh \left(\mathbf{W}_{h} \mathbf{h}_{t-1}+\mathbf{W}_{x} \mathbf{x}_{t}+\mathbf{b}\right)$. The skipping operation is conducted as
\begin{equation}\small\label{eq:RSN}
\mathbf{h}_{t}^{\prime}=\left\{
 \begin{array}{ll}{
 \mathbf{h}_{t}} & {x_{t} \in \mathcal{E}} \\ 
 {\mathbf{S}_{1} \mathbf{h}_{t}+\mathbf{S}_{2} \mathbf{x}_{t-1}} & {x_{t} \in \mathcal{R}}
 \end{array}\right.,
\end{equation}
where $\mathbf{S}_{1}$ and $\mathbf{S}_{2}$ are weight matrices.

\subsubsection{Transformers}
Transformer-based models have boosted contextualized text representation learning.
To utilize contextual information in knowledge graphs, CoKE~\cite{wang2019coke} employs transformers to encode edges and path sequences. Similarly, KG-BERT~\cite{yao2019kgbert} borrows the idea form language model pre-training and takes Bidirectional Encoder Representations from Transformer (BERT) model as an encoder for entities and relations. 

\subsubsection{Graph~Neural~Networks (GNNs)}
GNNs are introduced for learning connectivity structure under an encoder-decoder framework.
R-GCN~\cite{schlichtkrull2018modeling} proposes relation-specific transformation to model the directed nature of knowledge graphs. Its forward propagation is defined as
\begin{equation}\small
x_{i}^{(l+1)}=\sigma\left(\sum_{r \in \mathcal{R}} \sum_{j \in {N}_{i}^{r}} \frac{1}{c_{i, r}} W_{r}^{(l)} x_{j}^{(l)}+W_{0}^{(l)}x_{i}^{(l)}\right),
\end{equation}
where $x_{i}^{(l)} \in \mathbb{R}^{d^{(l)}}$ is the hidden state of the $i$-th entity in $l$-th layer, ${N}_i^r$ is a neighbor set of $i$-th entity within relation $r \in R$, $W_{r}^{(l)}$ and $W_{0}^{(l)}$ are the learnable parameter matrices, and $c_{i, r}$ is normalization such as $c_{i, r}=|{N}_{i}^{r}|$. Here, the GCN~\cite{kipf2016semi} acts as a graph encoder. To enable specific tasks, an encoder model still needs to be developed and integrated into the R-GCN framework. 
R-GCN takes the neighborhood of each entity equally.
SACN~\cite{shang2018end} introduces weighted GCN (Fig.~\ref{fig:SACN}), which defines the strength of two adjacent nodes with the same relation type, to capture the structural information in knowledge graphs by utilizing node structure, node attributes, and relation types. 
The decoder module called Conv-TransE adopts ConvE model as semantic matching metric and preserves the translational property. By aligning the convolutional outputs of entity and relation embeddings with $C$ kernels to be $\mathbf{M}\left(\mathbf{h}, \mathbf{r}\right)\in \mathbb{R}^{C \times d}$, its scoring function is defined as 
\begin{equation}\small
f_r(h, t) = g\left(\operatorname{vec}\left(\mathbf{M}\left(\mathbf{h}, \mathbf{r}\right)\right) W\right) \mathbf{t}.
\end{equation}
Nathani et al.~\cite{nathani2019learning} introduced graph attention networks with multi-head attention as encoder to capture multi-hop neighborhood features by inputing the concatenation of entity and relation embeddings.
CompGCN~\cite{vashishth2020composition} proposes entity-relation composition operations over each edge in the neighborhood of a central node and generalizes previous GCN-based models. 

\subsection{Embedding~with~Auxiliary~Information}
\label{sec:multi}
Multi-modal embedding incorporates external information such as text descriptions, type constraints, relational paths, and visual information, with a knowledge graph itself to facilitate more effective knowledge representation.

\subsubsection{Textual~Description}
Entities in knowledge graphs have textual descriptions denoted as $\mathcal{D}=<w_1, w_2, \dots, w_n>$, providing supplementary semantic information. 
The challenge of KRL with textual description is to embed both structured knowledge and unstructured textual information in the same space. 
Wang et al.~\cite{wang2014knowledge_text} proposed two alignment models for aligning entity space and word space by introducing entity names and Wikipedia anchors. 
DKRL~\cite{xie2016entity_desc} extends TransE~\cite{bordes2013translating} to learn representation directly from entity descriptions by a convolutional encoder.
SSP~\cite{xiao2017ssp} captures the strong correlations between triples and textual descriptions by projecting them in a semantic subspace.
The joint loss function is widely applied when incorporating KGE with textual description. 
Wang et al.~\cite{wang2014knowledge_text} used a three-component loss $\mathcal{L}=\mathcal{L}_K + \mathcal{L}_T + \mathcal{L}_A$ of knowledge model $\mathcal{L}_K$, text model $\mathcal{L}_T$ and alignment model $\mathcal{L}_A$. 
SSP~\cite{xiao2017ssp} uses a two-component objective function $\mathcal{L} = \mathcal{L}_{embed} + \mu \mathcal{L}_{topic}$ of embedding-specific loss $\mathcal{L}_{embed}$ and topic-specific loss $\mathcal{L}_{topic}$ within textual description, traded off by a parameter $\mu$.

\subsubsection{Type~Information}
Entities are represented with hierarchical classes or types, and consequently, relations with semantic types.
SSE~\cite{guo2015semantically} incorporates semantic categories of entities to embed entities belonging to the same category smoothly in semantic space.
TKRL~\cite{xie2016representation_type} proposes type encoder model for projection matrix of entities to capture type hierarchy. 
Noticing that some relations indicate attributes of entities, KR-EAR~\cite{lin2016knowledge} categorizes relation types into attributes and relations and modeled the correlations between entity descriptions.
Zhang et al.~\cite{zhang2018knowledge} extended existing embedding methods with hierarchical relation structure of relation clusters, relations, and sub-relations.

\subsubsection{Visual~Information}
Visual information (e.g., entity images) can be utilized to enrich KRL. 
Image-embodied IKRL~\cite{xie2017image}, containing cross-modal structure-based and image-based representation, encodes images to entity space and follows the translation principle. 
The cross-modal representations make sure that structure-based and image-based representations are in the same representation space. 
 
There remain many kinds of auxiliary information for KRL, such as attributes, relation paths, and logical rules. Wang et al.~\cite{wang2017knowledge} gave a detailed review of using additional information. 
This paper discusses relation path and logical rules under the umbrella of KGC in Sec.~\ref{sec:path} and~\ref{sec:rule}, respectively. 

\subsubsection{Uncertain Information}
Knowledge graphs such as ProBase~\cite{wu2012probase}, NELL~\cite{carlson2010toward}, and ConceptNet~\cite{speer2017conceptnet} contain uncertain information with a confidence score assigned to every relational fact. 
In contrast to classic deterministic knowledge graph embedding, uncertain embedding models aim to capture uncertainty representing the likelihood of relational facts. 
Chen et al.~\cite{chen2019embedding} proposed an uncertain knowledge graph embedding model to simultaneously preserve structural and uncertainty information, where probabilistic soft logic is applied to infer the confidence score. 
Probability calibration takes a post-processing process to adjust probability scores, making predictions probabilistic sense. 
Tabacof and Costabello~\cite{tabacof2019probability} firstly studied probability calibration for knowledge graph embedding under the closed-world assumption, revealing that well-calibrated models can lead to improved accuracy. 
Safavi et al.~\cite{safavi2020evaluating} further explored probability calibration under the more challenging open-world assumption.

\subsection{Summary}\label{sec:summarykrl}
Knowledge representation learning is vital in the research community of knowledge graph. 
This section reviews four folds of KRL with several modern methods summarized in Table~\ref{tab:recent-KRL} and more in Appendix~\ref{sup:KRL-summary}.
Overall, developing a novel KRL model is to answer the following four questions: 1) which representation space to choose; 2) how to measure the plausibility of triplets in a specific space; 3) which encoding model to use for modeling relational interactions; 4) whether to utilize auxiliary information. 
The most popularly used representation space is Euclidean point-based space by embedding entities in vector space and modeling interactions via vector, matrix, or tensor. Other representation spaces, including complex vector space, Gaussian distribution, and manifold space and group, are also studied. Manifold space has an advantage over point-wise Euclidean space by relaxing the point-wise embedding. 
Gaussian embeddings can express the uncertainties of entities and relations, and multiple relation semantics. 
Embedding in complex vector space can effectively model different relational connectivity patterns, especially the symmetry/antisymmetry pattern. 
The representation space plays an essential role in encoding the semantic information of entities and capturing the relational properties.
When developing a representation learning model, appropriate representation space should be selected and designed carefully to match the nature of encoding methods and balance the expressiveness and computational complexity. 
The scoring function with a distance-based metric utilizes the translation principle, while the semantic matching scoring function employs compositional operators.
Encoding models, especially neural networks, play a critical role in modeling interactions of entities and relations. The bilinear models also have drawn much attention, and some tensor factorization can also be regarded as this family. 
Other methods incorporate auxiliary information of textual description, relation/entity types, entity images, and confidence scores.

\begin{table}[htp]
\setlength{\tabcolsep}{2pt}
\scriptsize
\caption{A summary of recent KRL models. See more in Appendix~\ref{sup:KRL-summary}.
}
\begin{center}
\begin{tabular}{ l l l }
\toprule
Model & Ent. \& Rel. embed. & Scoring Function $f_r(h,t)$ \\
\midrule
RotatE~\cite{sun2018rotate} & $\mathbf{h}, \mathbf{t} \in \mathbb{C}^{d}$ , $\mathbf{r} \in \mathbb{C}^{d}$ &$\|\mathbf{h} \circ \mathbf{r}-\mathbf{t}\|$ \\
\hdashline
TorusE~\cite{ebisu2018toruse} & $[\mathbf{h}], [\mathbf{t}] \in \mathbb{T}^{n}$ , $[\mathbf{r}] \in \mathbb{T}^{n}$ & $\min _{(x, y) \in([h]+[r]) \times[t]}\|x-y\|_{i}$ \\
\hdashline
SimplE~\cite{kazemi2018simple} & $\mathbf{h}, \mathbf{t} \in \mathbb{R}^d$ , $\mathbf{r}, \mathbf{r}^\prime\in\mathbb{R}^d$ &$\frac{1}{2}\left( \mathbf{h} \circ \mathbf{r} \mathbf{t}+ \mathbf{t}\circ \mathbf{r}^{\prime} \mathbf{t}\right)$ \\
\hdashline
TuckER~\cite{balavzevic2019tucker} & $\mathbf{h}, \mathbf{t} \in \mathbb{R}^d_e$ , $\mathbf{r}\in\mathbb{R}^d_r$ & $\mathcal{W} \times_{1} \mathbf{h} \times_{2} \mathbf{r} \times_{3} \mathbf{t}$ \\
\hdashline
ITransF~\cite{xie2017interpretable} & $\mathbf{h}, \mathbf{t} \in \mathbb{R}^d$ , $\mathbf{r}\in\mathbb{R}^d$ & $\left\|\boldsymbol{\alpha}_{r}^{H} \cdot \mathbf{D} \cdot \mathbf{h}+\mathbf{r}-\boldsymbol{\alpha}_{r}^{T} \cdot \mathbf{D} \cdot \mathbf{t}\right\|_{\ell}$ \\
\hdashline
HolEx~\cite{xue2018expanding} & $\mathbf{h}, \mathbf{t} \in \mathbb{R}^d$ , $\mathbf{r}\in\mathbb{R}^d$ & $\sum_{j=0}^{l} p\left(\mathbf{h}, \boldsymbol{r} ; \boldsymbol{c}_{j}\right) \cdot \boldsymbol{t}$ \\
\hdashline
CrossE~\cite{zhang2019interaction}& $\mathbf{h}, \mathbf{t} \in \mathbb{R}^d$ , $\mathbf{r}\in\mathbb{R}^d$ & $\sigma\left(\sigma \left(\mathbf{c}_{r} \circ \mathbf{h}+\mathbf{c}_{r} \circ \mathbf{h} \circ \mathbf{r}+\mathbf{b}\right) \mathbf{t}^{\top}\right)$ \\
\hdashline
QuatE~\cite{zhang2019quaternion} &$\mathbf{h}, \mathbf{t} \in \mathbb{H}^d$ , $\mathbf{r} \in \mathbb{H}^d$ & $\mathbf{h} \otimes \frac{\mathbf{r}}{|\mathbf{r}|} \cdot \mathbf{t}$ \\
\hdashline
SACN~\cite{shang2018end} & $\mathbf{h}, \mathbf{t} \in \mathbb{R}^d$ , $\mathbf{r}\in\mathbb{R}^d$ & $g\left(\operatorname{vec}\left(\mathbf{M}\left(\mathbf{h}, \mathbf{r}\right)\right) W\right) \mathbf{t}$ \\
\hdashline
ConvKB~\cite{nguyen2017novel} & $\mathbf{h}, \mathbf{t} \in \mathbb{R}^d$ , $\mathbf{r}\in\mathbb{R}^d$ &$\operatorname{concat}\left(g\left(\left[\boldsymbol{h}, \boldsymbol{r}, \boldsymbol{t}\right] * \mathbf{\omega}\right)\right) \mathbf{w}$ \\
\hdashline
\multirow{2}{4em}{ConvE~\cite{dettmers2018convolutional}} & $\mathbf{M}_h \in \mathbb{R}^{d_w\times d_h}, \mathbf{t} \in \mathbb{R}^d$ & \multirow{2}{20pt}{$\sigma\left(\operatorname{vec}\left(\sigma\left(\left[\mathbf{M}_h ; \mathbf{M}_r\right] * \boldsymbol{\omega}\right)\right) \mathbf{W}\right) \mathbf{t}$}\\
& $\mathbf{M}_r \in\mathbb{R}^{d_w\times d_h}$ &\\
\hdashline
\multirow{2}{4em}{DihEdral~\cite{xu2019relation}} & $\mathbf{h}^{(l)}, \mathbf{t}^{(l)} \in \mathbb{R}^{2}$ & \multirow{2}{10pt}{$\sum_{l=1}^{L} \mathbf{h}^{(l) \top} \mathbf{R}^{(l)} \mathbf{t}^{(l)}$} \\
  & $\mathbf{R}^{(l)} \in \mathbb{D}_{K}$ & \\  
\hdashline
\multirow{2}{4em}{HAKE~\cite{zhang2020learning}} & $\mathbf{h}_{m}, \mathbf{t}_{m} \in \mathbb{R}^{d}, \mathbf{r}_{m} \in \mathbb{R}_{+}^{d} $  & \multirow{2}{10pt}{$-\left\|\mathbf{h}_{m} \circ \mathbf{r}_{m}-\mathbf{t}_{m}\right\|_{2}-\lambda\left\|\sin \left(\left(\mathbf{h}_{p}+\mathbf{r}_{p}-\mathbf{t}_{p}\right) / 2\right)\right\|_{1}$} \\
& $\mathbf{h}_{p}, \mathbf{r}_{p}, \mathbf{t}_{p} \in[0,2 \pi)^{d}$ & \\
\hdashline

MuRP~\cite{balazevic2019multi}	&	$\mathbf{h}, \mathbf{t}, \mathbf{r} \in \mathbb{B}_{c}^{d}, b_{h}, b_{t} \in \mathbb{R}$	&	$-d_{\mathbb{B}}\left(\mathbf{h}^{(r)}, \mathbf{t}^{(r)}\right)^{2}+b_{s}+b_{o}$	\\
\hdashline
AttH~\cite{chami2020low}	&	$\mathbf{h}, \mathbf{t}, \mathbf{r} \in \mathbb{B}_{c}^{d}, b_{h}, b_{t} \in \mathbb{R}$	&	$-d_{\mathbb{B}}^{c_{r}}\left(Q(h, r), \mathbf{e}_{t}^{H}\right)^{2}+b_{h}+b_{t}$\\	
\hdashline
LowFER~\cite{amin2020lowfer}	&	$\mathbf{h}, \mathbf{t} \in \mathbb{R}^{d}, \mathbf{r} \in \mathbb{R}^{d}$	&	$\left(\mathbf{S}^{k} \operatorname{diag}\left(\mathbf{U}^{T} \mathbf{h}\right) \mathbf{V}^{T} \mathbf{r}\right)^{T} \mathbf{t}$	\\
\bottomrule 
\end{tabular}
\end{center}
\label{tab:recent-KRL}
\end{table}

\section{Knowledge Acquisition}
\label{sec:acquisition}
Knowledge acquisition aims to construct knowledge graphs from unstructured text and other structured or semi-structured sources, complete an existing knowledge graph, and discover and recognize entities and relations. 
Well-constructed and large-scale knowledge graphs can be useful for many downstream applications and empower knowledge-aware models with commonsense reasoning, thereby paving the way for AI. 
The main tasks of knowledge acquisition include relation extraction, KGC, and other entity-oriented acquisition tasks such as entity recognition and entity alignment. 
Most methods formulate KGC and relation extraction separately. These two tasks, however, can also be integrated into a unified framework. 
Han et al.~\cite{han2018neural} proposed a joint learning framework with mutual attention for data fusion between knowledge graphs and text, which solves KGC and relation extraction from text. 
There are also other tasks related to knowledge acquisition, such as triple classification~\cite{dong2019triple}, relation classification~\cite{zhou2016attention}, and open knowledge enrichment~\cite{cao2020open}. 
In this section, three categories of knowledge acquisition techniques, namely  KGC, entity discovery, and relation extraction are reviewed thoroughly. 

\subsection{Knowledge~Graph~Completion}
\label{sec:KGC}
Because of the nature of incompleteness of knowledge graphs, KGC is developed to add new triples to a knowledge graph. 
Typical subtasks include link prediction, entity prediction, and relation prediction.

Preliminary research on KGC focused on learning low-dimensional embedding for triple prediction. In this survey, we term those methods as \textit{embedding-based methods}. Most of them, however, failed to capture multi-step relationships. Thus, recent work turns to explore multi-step relation paths and incorporate logical rules, termed as \textit{relation path inference} and \textit{rule-based reasoning}, respectively.
Triple classification as an associated task of KGC, which evaluates the correctness of a factual triple, is additionally reviewed in this section. 

\subsubsection{Embedding-based~Models}
\begin{figure}[htbp]
\begin{center}
\begin{subfigure}[b]{0.24\textwidth}
	\begin{center}
	\includegraphics[height=2.2cm]{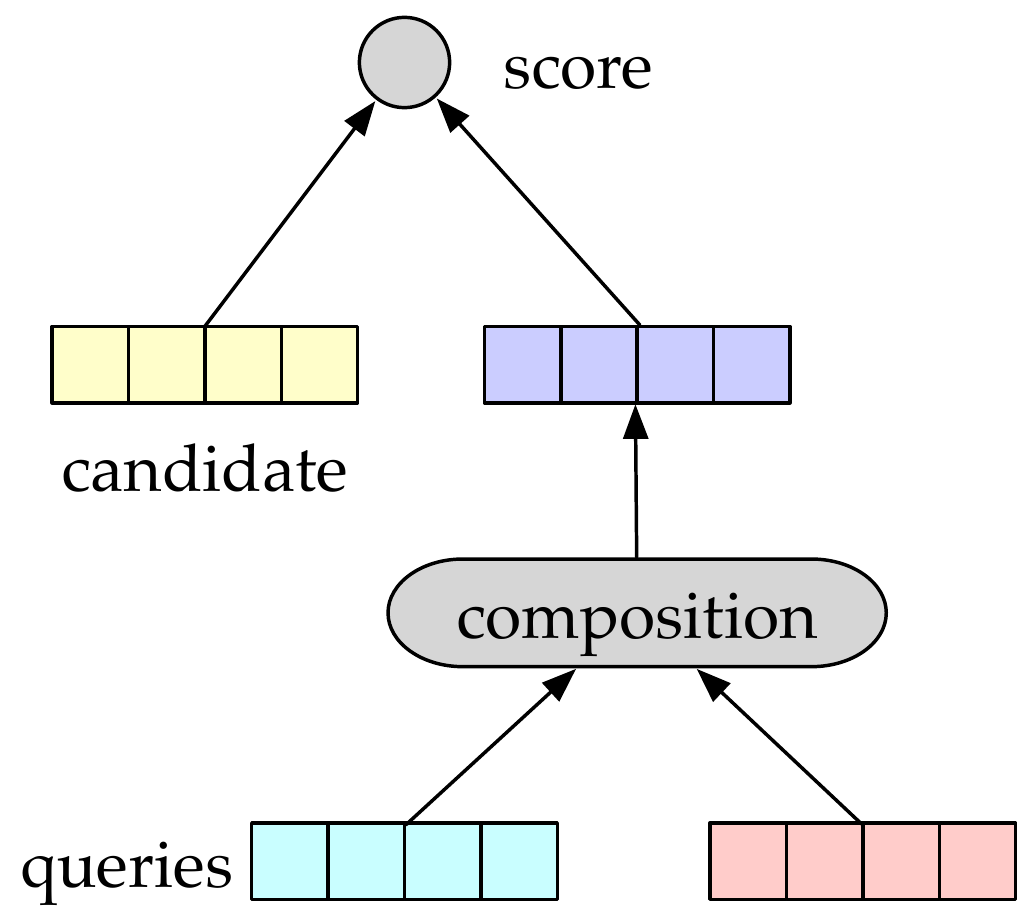}
	\caption{Embedding-based Ranking.}
	\label{fig:kgc-embedding}
	\end{center}
\end{subfigure}
\quad
\begin{subfigure}[b]{0.22\textwidth}
	\begin{center}
	\includegraphics[height=2.2cm]{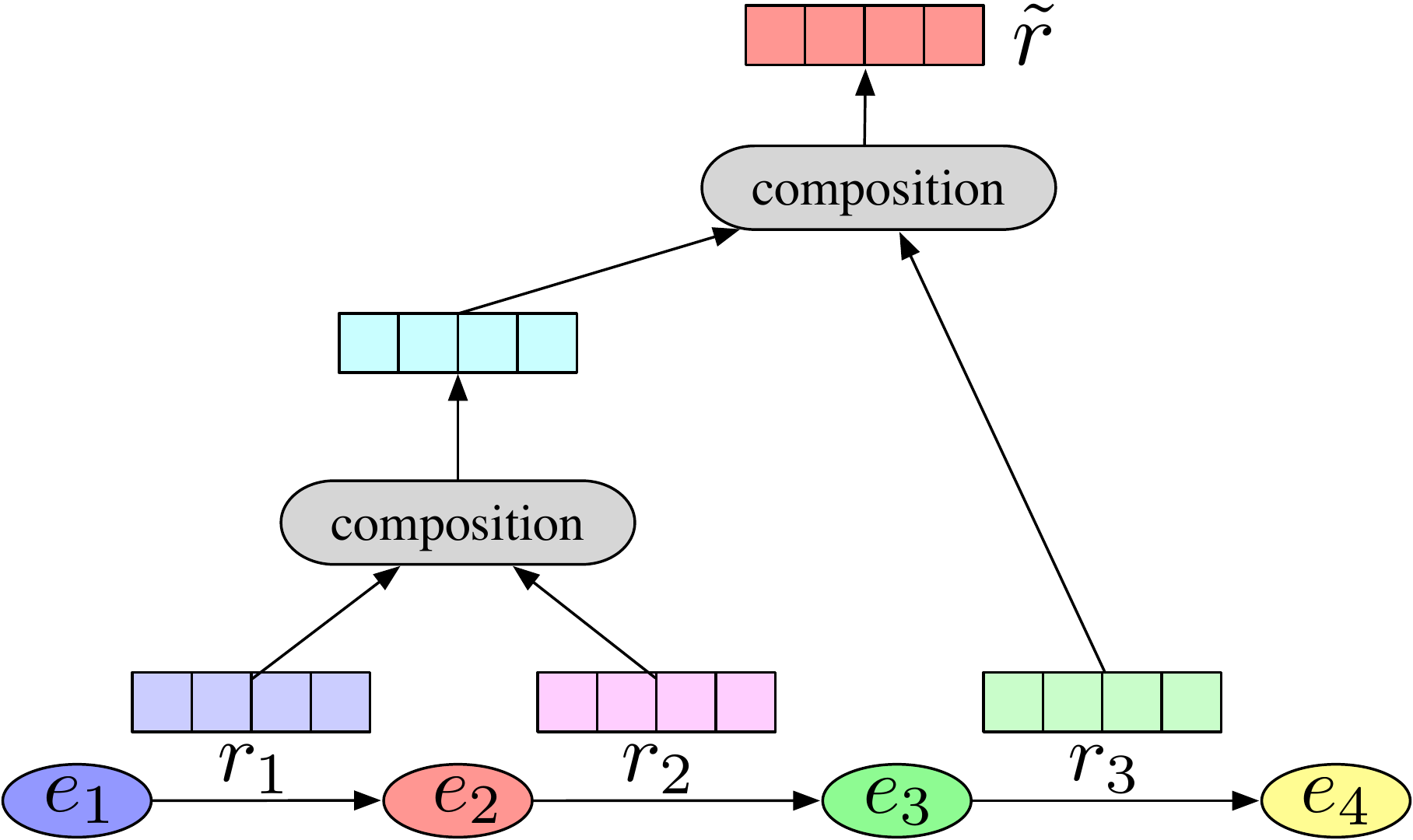}
	\caption{Relation paths~\cite{neelakantan2015compositional}.}
	\label{fig:kgc-path}
	\end{center}
\end{subfigure}
\caption{Illustrations of embedding-based ranking and relation path reasoning.}
\label{fig:reasoning}
\end{center}
\end{figure}

Taking entity prediction as an example, embedding-based ranking methods, as shown in Fig.~\ref{fig:kgc-embedding}, firstly learn embedding vectors based on existing triples. By replacing the tail entity or head entity with each entity $e\in \mathcal{E}$, those methods calculate scores of all the candidate entities and rank the top $k$ entities.
Aforementioned KRL methods (e.g., TransE~\cite{bordes2013translating}, TransH~\cite{wang2014knowledge}, TransR~\cite{lin2015learning}, HolE~\cite{nickel2016holographic}, and R-GCN~\cite{schlichtkrull2018modeling}) and joint learning methods like DKRL~\cite{xie2016entity_desc} with textual information can been used for KGC. 

Unlike representing inputs and candidates in the unified embedding space, ProjE~\cite{shi2017proje} proposes a combined embedding by space projection of the known parts of input triples, i.e., $(h, r, ?)$ or $(?, r, t)$, and the candidate entities with the candidate-entity matrix $\mathbf{W}^{c} \in \mathbb{R}^{s \times d}$, where $s$ is the number of candidate entities. The embedding projection function including a neural combination layer and a output projection layer is defined as $h(\mathbf{e}, \mathbf{r})=g\left(\mathbf{W}^{c} \sigma(\mathbf{e} \oplus \mathbf{r})+b_{p}\right)$, 
where $\mathbf{e} \oplus \mathbf{r}=\mathbf{D}_{e} \mathbf{e}+\mathbf{D}_{r} \mathbf{r}+\mathbf{b}_{c}$ is the combination operator of input entity-relation pair. Previous embedding methods do not differentiate entities and relation prediction, and ProjE does not support relation prediction.
Based on these observations, SENN~\cite{guan2018shared} distinguishes three KGC subtasks explicitly by introducing a unified neural shared embedding with adaptively weighted general loss function to learn different latent features.
Existing methods rely heavily on existing connections in knowledge graphs and fail to capture the evolution of factual knowledge or entities with a few connections. 
ConMask~\cite{shi2018open} proposes relationship-dependent content masking over the entity description to select relevant snippets of given relations, and CNN-based target fusion to complete the knowledge graph with unseen entities. It can only make a prediction when query relations and entities are explicitly expressed in the text description. Previous methods are discriminative models that rely on preprepared entity pairs or text corpus. Focusing on the medical domain, REMEDY~\cite{zhang2018generative} proposes a generative model, called conditional relationship variational autoencoder, for entity pair discovery from latent space.

\subsubsection{Relation~Path~Reasoning}
\label{sec:path}
Embedding learning of entities and relations has gained remarkable performance in some benchmarks, but it fails to model complex relation paths. 
Relation path reasoning turns to leverage path information over the graph structure. 
Random walk inference has been widely investigated; for example, the Path-Ranking Algorithm (PRA)~\cite{lao2010relational} chooses a relational path under a combination of path constraints and conducts maximum-likelihood classification. 
To improve path search, Gardner et al.~\cite{gardner2014incorporating} introduced vector space similarity heuristics in random walk by incorporating textual content, which also relieves the feature sparsity issue in PRA.  
Neural multi-hop relational path modeling is also studied. Neelakantan et al.~\cite{neelakantan2015compositional} developed an RNN model to compose the implications of relational paths by applying compositionality recursively (in Fig.~\ref{fig:kgc-path}). 
Chain-of-Reasoning~\cite{das2017chains}, a neural attention mechanism to enable multiple reasons, represents logical composition across all relations, entities, and text.
Recently, DIVA~\cite{chen2018variational} proposes a unified variational inference framework that takes multi-hop reasoning as two sub-steps of path-finding (a prior distribution for underlying path inference) and path-reasoning (a likelihood for link classification). 

\subsubsection{RL-based~Path~Finding} 
Deep reinforcement learning (RL) is introduced for multi-hop reasoning by formulating path-finding between entity pairs as sequential decision making, specifically a Markov decision process (MDP). 
The policy-based RL agent learns to find a step of relation to extending the reasoning paths via the interaction between the knowledge graph environment, where the policy gradient is utilized for training RL agents.

DeepPath~\cite{xiong2017deeppath} firstly applies RL into relational path learning and develops a novel reward function to improve accuracy, path diversity, and path efficiency. It encodes states in the continuous space via a translational embedding method and takes the relation space as its action space. 
Similarly, MINERVA~\cite{das2017go} takes path walking to the correct answer entity as a sequential optimization problem by maximizing the expected reward. It excludes the target answer entity and provides more capable inference. 
Instead of using a binary reward function, Multi-Hop~\cite{lin2018multi} proposes a soft reward mechanism. Action dropout is also adopted to mask some outgoing edges during training to enable more effective path exploration. 
M-Walk~\cite{shen2018m} applies an RNN controller to capture the historical trajectory and uses the Monte Carlo Tree Search (MCTS) for effective path generation.
By leveraging text corpus with the sentence bag of current entity denoted as $b_{e_{t}}$, CPL~\cite{fu2019collaborative} proposes collaborative policy learning for pathfinding and fact extraction from text. 

With source, query and current entity denoted as $e_s$, $e_q$ and $e_t$, and query relation denoted as $r_q$, the MDP environment and policy networks of these methods are summarized in Table~\ref{tab:RL}, where MINERVA, M-Walk and CPL use binary reward. For the policy networks, DeepPath uses fully-connected network, the extractor of CPL employs CNN, while the rest uses recurrent networks. 
\begin{table*}[htp]
\setlength{\tabcolsep}{2pt}
\scriptsize
\caption{Comparison of RL-based path finding for knowledge graph reasoning.}
\begin{center}
\begin{tabular}{lll l l}
\toprule
Method & State $s_t$ & Action $a_t$ & Reward $\gamma$ & Policy Network \\
\midrule
\multirow{3}{4em}{DeepPath~\cite{xiong2017deeppath}} & & & Global~$1~\mathrm{e}_{\mathrm{t}}=\mathrm{e}_{\mathrm{q}}$ or $-1~\mathrm{e}_{\mathrm{t}}\not=\mathrm{e}_{\mathrm{q}}$ & \\
 & $(\mathbf{e}_t, \mathbf{e}_{q} - \mathbf{e}_t)$ & $\{r\}$ & Efficiency~$\frac{1}{\operatorname{length}(p)}$ & Fully-connected network (FCN) \\
 & & & Diversity~$-\frac{1}{|F|} \sum_{i=1}^{|F|} \cos \left(\mathbf{p}, \mathbf{p}_{i}\right)$ & \\
\hdashline
MINERVA~\cite{das2017go} & $(e_t, e_s, r_q, e_q)$& $\left\{\left(e_{t}, r, v\right)\right\}$ & $\mathbb{I}\left\{\mathrm{e}_{\mathrm{t}}=\mathrm{e}_{\mathrm{q}}\right\}$ & $\mathbf{h}_{{t}}={LSTM}\left(\mathbf{h}_{\mathbf{t}-\mathbf{1}},\left[\mathbf{a}_{\mathbf{t}-\mathbf{1}} ; \mathbf{o}_{\mathbf{t}}\right]\right)$\\
\hdashline
Multi-Hop~\cite{lin2018multi} & $\left(e_{t},\left(e_{s}, r_{q}\right)\right)$ & $\left\{\left(r^{\prime}, e^{\prime}\right) |\left(e_{t}, r^{\prime}, e^{\prime}\right) \in \mathcal{G}\right\}$ & $\gamma +\left(1-\gamma \right) f_{r_q}\left(e_{s}, e_{T}\right)$& $\mathbf{h}_{t}= {LSTM}\left(\mathbf{h}_{t-1}, \mathbf{a}_{t-1}\right)$ \\
\hdashline
 M-Walk~\cite{shen2018m}& $s_{t-1} \cup\left\{a_{t-1}, v_{t}, \mathcal{E_G}_{v_{t}}, \mathcal{V}_{v_{t}}\right\}$& $\bigcup_{t}\mathcal{E_G}_{v_{t}} \cup\{\mathrm{STOP}\}$ & $\mathbb{I}\left\{\mathrm{e}_{\mathrm{t}}=\mathrm{e}_{\mathrm{q}}\right\}$& GRU-RNN + FCN\\
 \hdashline
 CPL~\cite{fu2019collaborative} Reasoner & $\left(e_{s}, r_{q}, h_{t}\right)$ & $\{ \xi \in \mathcal{E_G}\}$ & $\mathbb{I}\left\{\mathrm{e}_{\mathrm{t}}=\mathrm{e}_{\mathrm{q}}\right\}$ & $\mathbf{h}_{t}={LSTM}\left(\mathbf{h}_{t-1},\left[\mathbf{r}_{t}, \mathbf{e}_{t}\right]\right)$ \\
 CPL~\cite{fu2019collaborative} Extractor & $\left(b_{e_{t}}, e_{t}\right)$ & $\left\{\left(r^{\prime}, e^{\prime}\right)\right\}_{\left(e_{t}, r^{\prime}, e^{\prime}\right)} \in b_{e_t}$ & step-wise delayed from reasoner & PCNN-ATT \\
\bottomrule
\end{tabular}
\end{center}
\label{tab:RL}
\end{table*}

\subsubsection{Rule-based~Reasoning} 
\label{sec:rule}
To better make use of the symbolic nature of knowledge, another research direction of KGC is logical rule learning. 
A rule is defined by the head and body in the form of $head \leftarrow body$. The $head$ is an atom, i.e., a fact with variable subjects and/or objects, while the body can be a set of atoms. For example, given relations $\texttt{sonOf}$, $\texttt{hasChild}$ and $\texttt{gender}$, and entities $X$ and $Y$, there is a rule in the reverse form of logic programming as:
\begin{equation*}\small
(\textit{Y}, \texttt{sonOf}, \textit{X}) \leftarrow (\textit{X}, \texttt{hasChild}, \textit{Y}) \wedge (\textit{Y}, \texttt{gender}, \textit{Male})
\end{equation*}
Logical rules can been extracted by rule mining tools like AMIE~\cite{galarraga2013amie}.
The recent RLvLR~\cite{omran2019embedding} proposes a scalable rule mining approach with efficient rule searching and pruning, and uses the extracted rules for link prediction. 

\begin{figure}[htbp]
\begin{center}
\begin{subfigure}[b]{0.22\textwidth}
	\begin{center}
	\includegraphics[height=3.2cm]{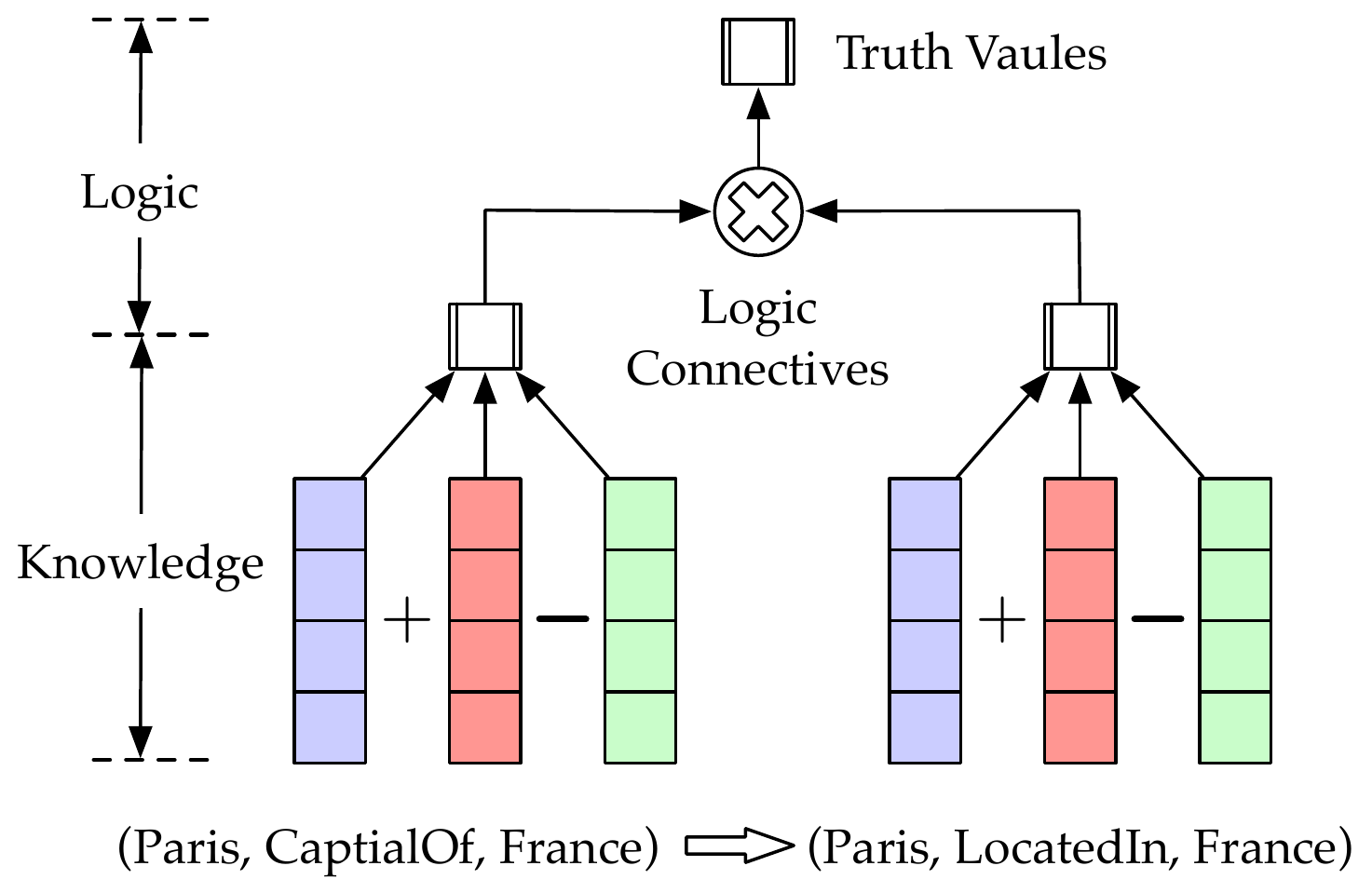}
	\caption{KALE~\cite{guo2016jointly}.}
	\label{fig:KALE}
	\end{center}
\end{subfigure}
\qquad
\begin{subfigure}[b]{0.22\textwidth}
	\begin{center}
	\includegraphics[height=3.2cm]{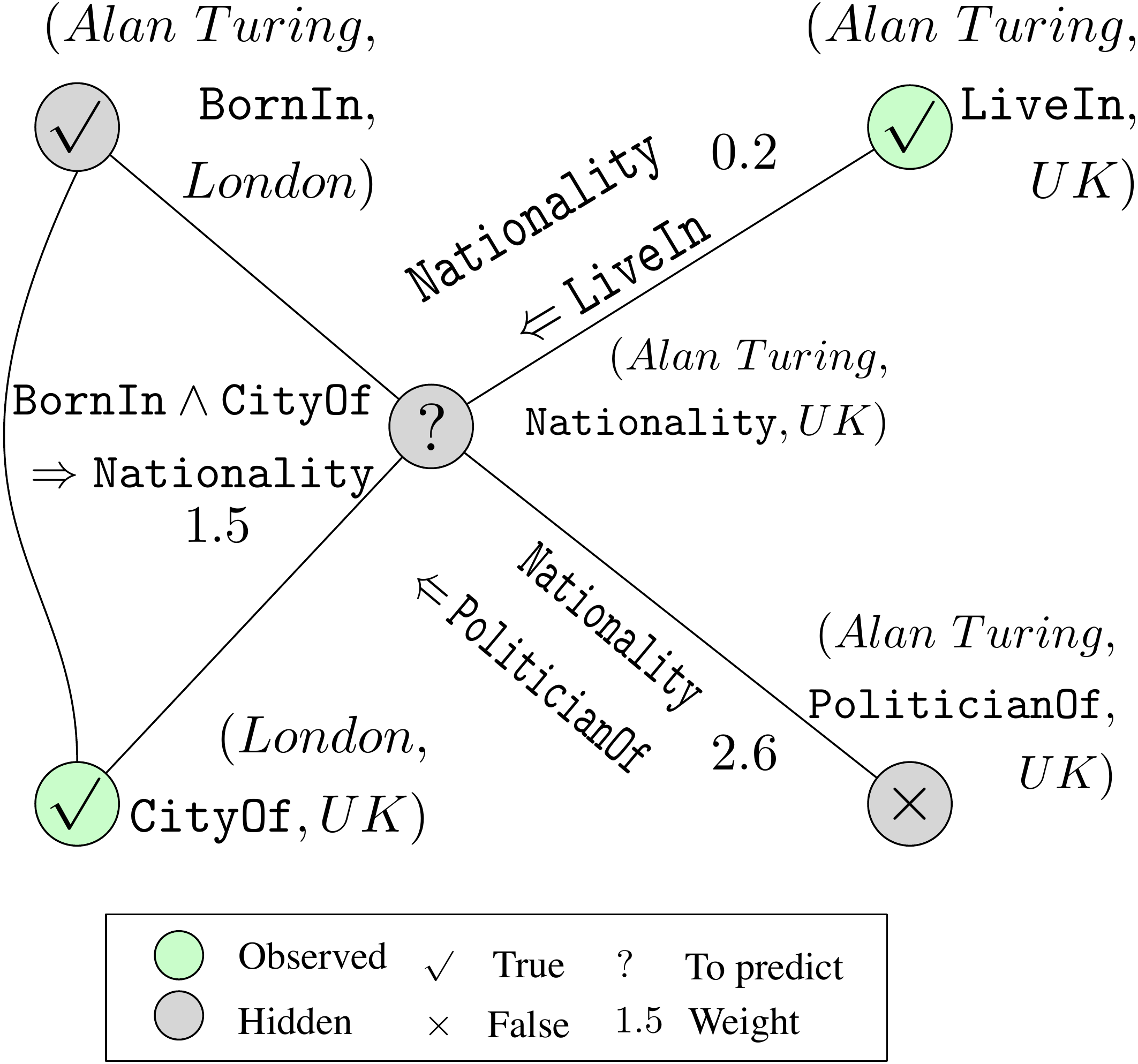}
	\caption{pLogicNet~\cite{qu2019probabilistic}.}
	\label{fig:pLogicNet}
	\end{center}
\end{subfigure}
\caption{Illustrations of logical rule learning.}
\label{fig:reasoning-rule}
\end{center}
\end{figure}

More research attention focuses on injecting logical rules into embeddings to improve reasoning, with joint learning or iterative training applied to incorporate first-order logic rules.
For example, KALE~\cite{guo2016jointly} proposes a unified joint model with t-norm fuzzy logical connectives defined for compatible triples and logical rules embedding. Specifically, three compositions of logical conjunction, disjunction, and negation are defined to compose the truth value of a complex formula. Fig.~\ref{fig:KALE} illustrates a simple first-order Horn clause inference.
RUGE~\cite{guo2018knowledge} proposes an iterative model, where soft rules are utilized for soft label prediction from unlabeled triples and labeled triples for embedding rectification. 
IterE~\cite{zhang2019iteratively} proposes an iterative training strategy with three components of embedding learning, axiom induction, and axiom injection. 

The logical rule is one kind of auxiliary information; meanwhile, it can incorporate prior knowledge, enabling the ability of interpretable multi-hop reasoning and paving the way for generalization even in few-shot labeled relational triples. 
However, logic rules alone can only cover a limited number of relational facts in knowledge graphs and suffer colossal  search space. 
The combination of neural and symbolic computation has complementary advantages that utilize efficient data-driven learning and differentiable optimization and exploit prior logical knowledge for precise and interpretable inference. 
Incorporating rule-based learning for knowledge representation is principally to add regularizations or constraints to representations. 
Neural Theorem Provers (NTP)~\cite{rocktaschel2017end} learns logical rules for multi-hop reasoning, which utilizes a radial basis function kernel for differentiable computation on vector space. 
NeuralLP~\cite{yang2017differentiable} enables gradient-based optimization to be applicable in the inductive logic programming, where a neural controller system is proposed by integrating attention mechanism and auxiliary memory. 
Neural-Num-LP~\cite{wang2020differentiable} extends NeuralLP to learn numerical rules with dynamic programming and cumulative sum operations. 
pLogicNet~\cite{qu2019probabilistic} proposes probabilistic logic neural networks (Fig.~\ref{fig:pLogicNet}) to leverage first-order logic and learn effective embedding by combining the advantages of Markov logic networks and KRL methods while handling the uncertainty of logic rules. ExpressGNN~\cite{zhang2020efficient} generalizes pLogicNet by tuning graph networks and embedding and achieves more efficient logical reasoning. 

\subsubsection{Meta~Relational~Learning}
The long-tail phenomena exist in the relations of knowledge graphs. Meanwhile, the real-world scenario of knowledge is dynamic, where unseen triples are usually acquired. The new scenario, called as \textit{meta relational learning} or \textit{few-shot relational learning}, requires models to predict new relational facts with only a very few samples.

Targeting at the previous two observations, GMatching~\cite{xiong2018one} develops a metric based few-shot learning method with entity embeddings and local graph structures. It encodes one-hop neighbors to capture the structural information with R-GCN and then takes the structural entity embedding for multi-step matching guided by long short-term memory (LSTM) networks to calculate the similarity scores.
Meta-KGR~\cite{lv2019adapting}, an optimization-based meta-learning approach, adopts model agnostic meta-learning for fast adaption and reinforcement learning for entity searching and path reasoning. 
Inspired by model-based and optimization-based meta-learning, MetaR~\cite{chen2019meta} transfers relation-specific meta information from support set to query set, and archives fast adaption via loss gradient of high-order relational representation. 
Zhang et al.~\cite{zhang2020few} proposed joint modules of heterogeneous graph encoder, recurrent autoencoder, and matching network to complete new relational facts with few-shot references.
Qin et al.~\cite{qin2020generative} utilized GAN to generate reasonable embeddings for unseen relations under the zero-shot learning setting.
Baek et al.~\cite{baek2020learning} proposed a transductive meta-learning framework, called Graph Extrapolation Networks (GEN), for few-shot out-of-graph link prediction in knowledge graphs. 

\subsubsection{Triple~Classification}
Triple classification is to determine whether facts are correct in testing data, which is typically regarded as a binary classification problem. The decision rule is based on the scoring function with a specific threshold. 
Aforementioned embedding methods could be applied for triple classification, including translational distance-based methods like TransH~\cite{wang2014knowledge} and TransR~\cite{lin2015learning} and semantic matching-based methods such as NTN~\cite{socher2013reasoning}, HolE~\cite{nickel2016holographic} and ANALOGY~\cite{liu2017analogical}.

Vanilla vector-based embedding methods failed to deal with $1$-to-$n$ relations. Recently, Dong et al.~\cite{dong2019triple} extended the embedding space into region-based $n$-dimensional balls where the tail region is in the head region for $1$-to-$n$ relation using fine-grained type chains, i.e., tree-structure conceptual clusterings. 
This relaxation of embedding to $n$-balls turns triple classification into a geometric containment problem and improves the performance for entities with long type chains. However, it relies on the type chains of entities and suffers from the scalability problem.

\subsection{Entity~Discovery}
This section distinguishes entity-based knowledge acquisition into several fractionized tasks, i.e., entity recognition, entity disambiguation, entity typing, and entity alignment. We term them as \textit{entity~discovery} as they all explore entity-related knowledge under different settings.

\subsubsection{Entity~Recognition}
Entity recognition or named entity recognition (NER), when it focuses on specifically named entities, is a task that tags entities in text. Hand-crafted features such as capitalization patterns and language-specific resources like gazetteers are applied in many pieces of literature. Recent work applies sequence-to-sequence neural architectures, for example, LSTM-CNN~\cite{chiu2016named} for learning character-level and word-level features and encoding partial lexicon matches. 
Lample et al.~\cite{lample2016neural} proposed stacked neural architectures by stacking LSTM layers and CRF layers, i.e., LSTM-CRF (in Fig.~\ref{fig:entity-recoginition}) and Stack-LSTM. 
MGNER~\cite{xia2019multi} proposes an integrated framework with entity position detection in various granularities and attention-based entity classification for both nested and non-overlapping named entities. 
Hu et al.~\cite{hu2020leveraging} distinguished multi-token and single-token entities with multi-task training.
Recently, Li et al.~\cite{li2020unified} formulated flat and nested NER as a unified machine reading comprehension framework by referring annotation guidelines to construct query questions. 
Pretrained language models with knowledge graphs such as ERNIE~\cite{sun2020ernie} and K-BERT~\cite{weijie2019kbert} have been applied into NER and achieved improved performance.

\subsubsection{Entity~Typing}
Entity typing includes coarse and fine-grained types, while the latter uses a tree-structured type category and is typically regarded as multi-class and multi-label classification. To reduce label noise, PLE~\cite{ren2016label} focuses on correct type identification and proposes a partial-label embedding model with a heterogeneous graph for the representation of entity mentions, text features, and entity types and their relationships. 
To tackle the increasing growth of typeset and noisy labels, Ma et al.~\cite{ma2016label} proposed prototype-driven label embedding with hierarchical information for zero-shot fine-grained named entity typing. 
Recent studies utilize embedding-based approaches. For example, JOIE~\cite{hao2019universal} learns joint embeddings of instance- and ontology-view graphs and formulates entity typing as top-$k$ ranking to predict associated concepts. 
ConnectE~\cite{zhao2020connecting} explores local typing and global triple knowledge to enhance joint embedding learning.

\subsubsection{Entity~Disambiguation}
Entity disambiguation or entity linking is a unified task which links entity mentions to the corresponding entities in a knowledge graph. For example, Einstein won the Noble Prize in Physics in 1921. The entity mention of ``Einstein'' should be linked to the entity of \texttt{Albert~Einstein}. 
The contemporary end-to-end learning approaches have made efforts through representation learning of entities and mentions, for example, DSRM~\cite{huang2015leveraging} for modeling entity semantic relatedness and EDKate~\cite{fang2016entity} for the joint embedding of entity and text. 
Ganea and Hofmann~\cite{ganea2017deep} proposed an attentive neural model over local context windows for entity embedding learning and differentiable message passing for inferring ambiguous entities. 
By regarding relations between entities as latent variables, Le and Titov~\cite{le2018improving} developed an end-to-end neural architecture with relation-wise and mention-wise normalization. 

\subsubsection{Entity~Alignment}
\begin{figure}[htbp]
\begin{center}
\begin{subfigure}[b]{0.23\textwidth}
	\begin{center}
	\includegraphics[height=3cm]{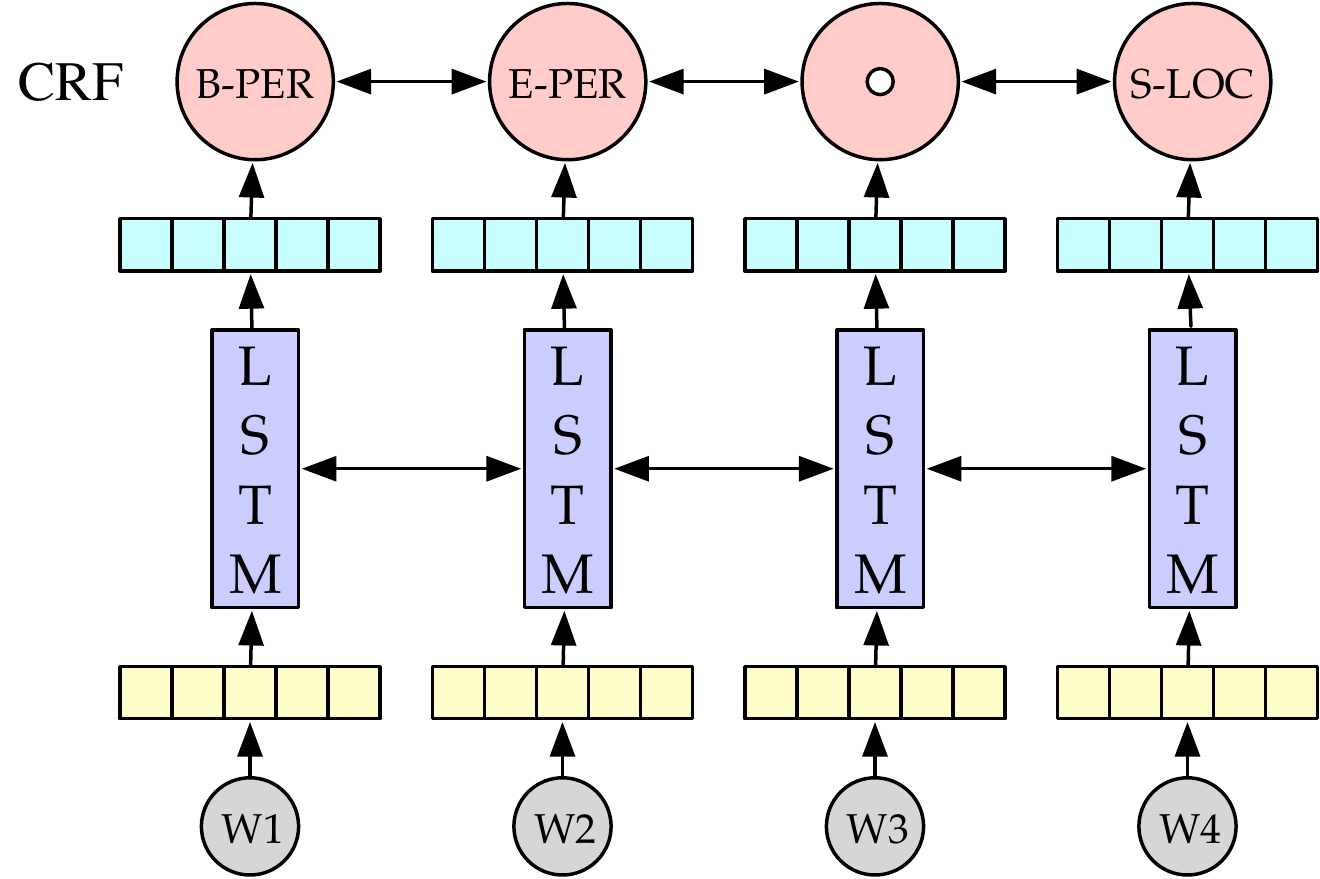}
	\caption{Entity recognition with LSTM-CRF~\cite{lample2016neural}.}
	\label{fig:entity-recoginition}
	\end{center}
\end{subfigure}
\quad
\begin{subfigure}[b]{0.23\textwidth}
	\begin{center}
	\includegraphics[height=3cm]{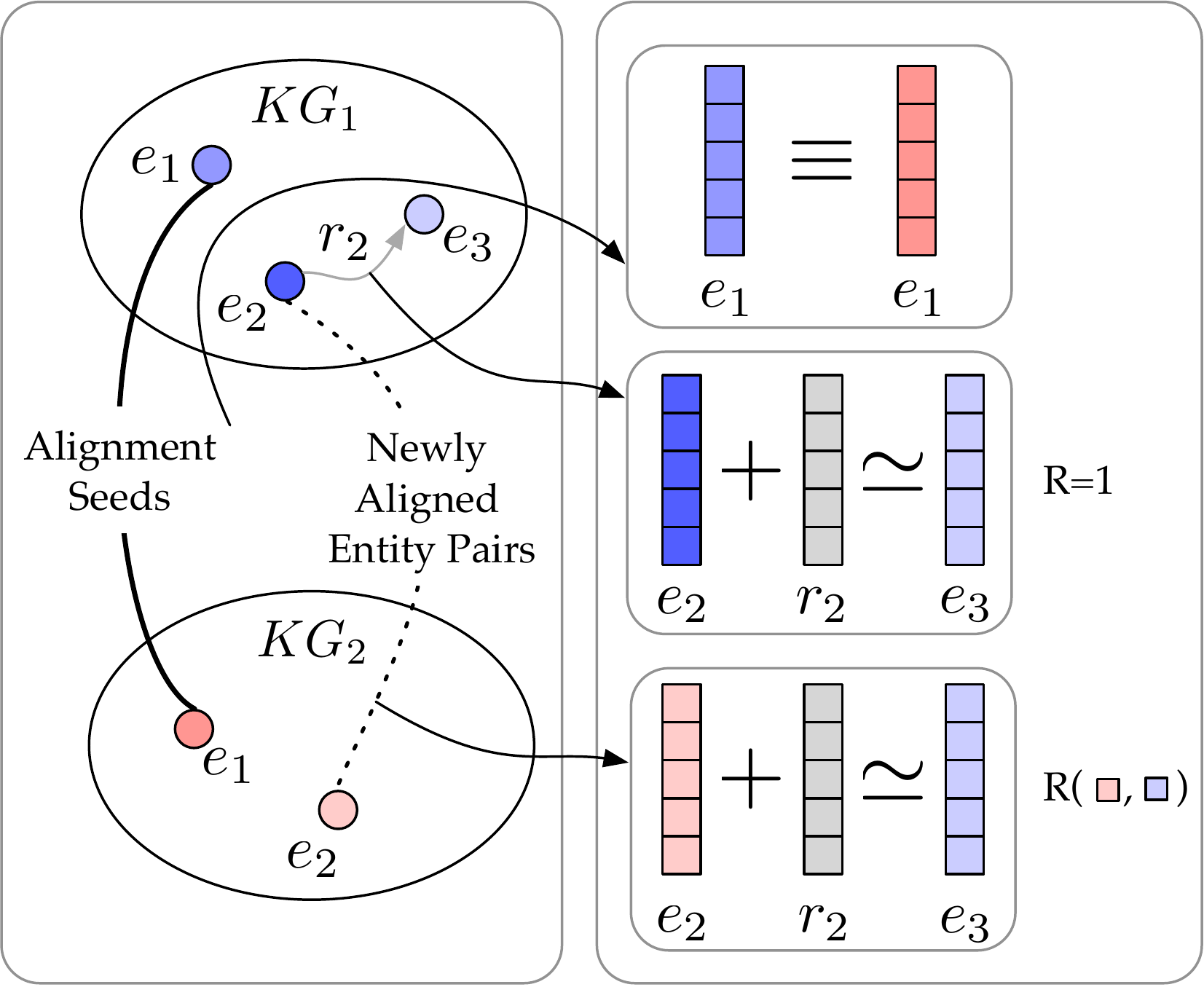}
	\caption{Entity alignment with IPTransE~\cite{zhu2017iterative}.}
	\label{fig:entity-alignment}
	\end{center}
\end{subfigure}
\caption{Illustrations of several entity discovery tasks.}
\label{fig:entity}
\end{center}
\end{figure}

The tasks, as mentioned earlier, involve entity discovery from text or a single knowledge graph, while entity alignment (EA) aims to fuse knowledge among various knowledge graphs. 
Given $\mathcal{E}_1$ and $\mathcal{E}_2$ as two different entity sets of two different knowledge graphs, EA is to find an alignment set $A=\{ (e_1, e_2) \in \mathcal{E}_1\times \mathcal{E}_2 | e_1 \equiv e_2 \}$, where entity $e_1$ and entity $e_2$ hold an equivalence relation $\equiv$. 
In practice, a small set of alignment seeds (i.e., synonymous entities appear in different knowledge graphs) is given to start the alignment process, as shown in the left box of Fig.~\ref{fig:entity-alignment}. 

Embedding-based alignment calculates the similarity between the embeddings of a pair of entities.
MTransE~\cite{chen2017multilingual} firstly studies entity alignment in the multilingual scenario. It considers distance-based axis calibration, translation vectors, and linear transformations for cross-lingual entity matching and triple alignment verification. 
Following the translation-based and linear transformation models, IPTransE~\cite{zhu2017iterative} proposes an iterative alignment model by mapping entities into a unified representation space under a joint embedding framework (Fig.~\ref{fig:entity-alignment}) through aligned translation as $\left\|\mathbf{e}_{1}+\mathbf{r}^{\left(\mathcal{E}_{1} \rightarrow \mathcal{E}_{2}\right)}-\mathbf{e}_{2}\right\|$, linear transformation as $\left\|\mathbf{M}^{\left(\mathcal{E}_{1} \rightarrow \mathcal{E}_{2}\right)} \mathbf{e}_{1}-\mathbf{e}_{2}\right\|$, 
and parameter sharing as $\mathbf{e}_{1} \equiv \mathbf{e}_{2}$.
To solve error accumulation in iterative alignment, BootEA~\cite{sun2018bootstrapping} proposes a bootstrapping approach in an incremental training manner, together with an editing technique for checking newly-labeled alignment. 

Additional information of entities is also incorporated for refinement, for example, JAPE~\cite{sun2017cross} capturing the correlation between cross-lingual attributes, KDCoE~\cite{chen2018co} embedding multi-lingual entity descriptions via co-training, MultiKE~\cite{zhang2019multi} learning multiple views of the entity name, relation, and attributes, and alignment with character attribute embedding~\cite{trsedya2019entity}.
Entity alignment has been intensively studied in recent years. We recommend Sun et al.’s quantitative survey~\cite{sun2020benchmarking} for detailed reading.

\subsection{Relation~Extraction}
\begin{figure}[htbp]
\begin{center}
	\includegraphics[width=0.3\textwidth]{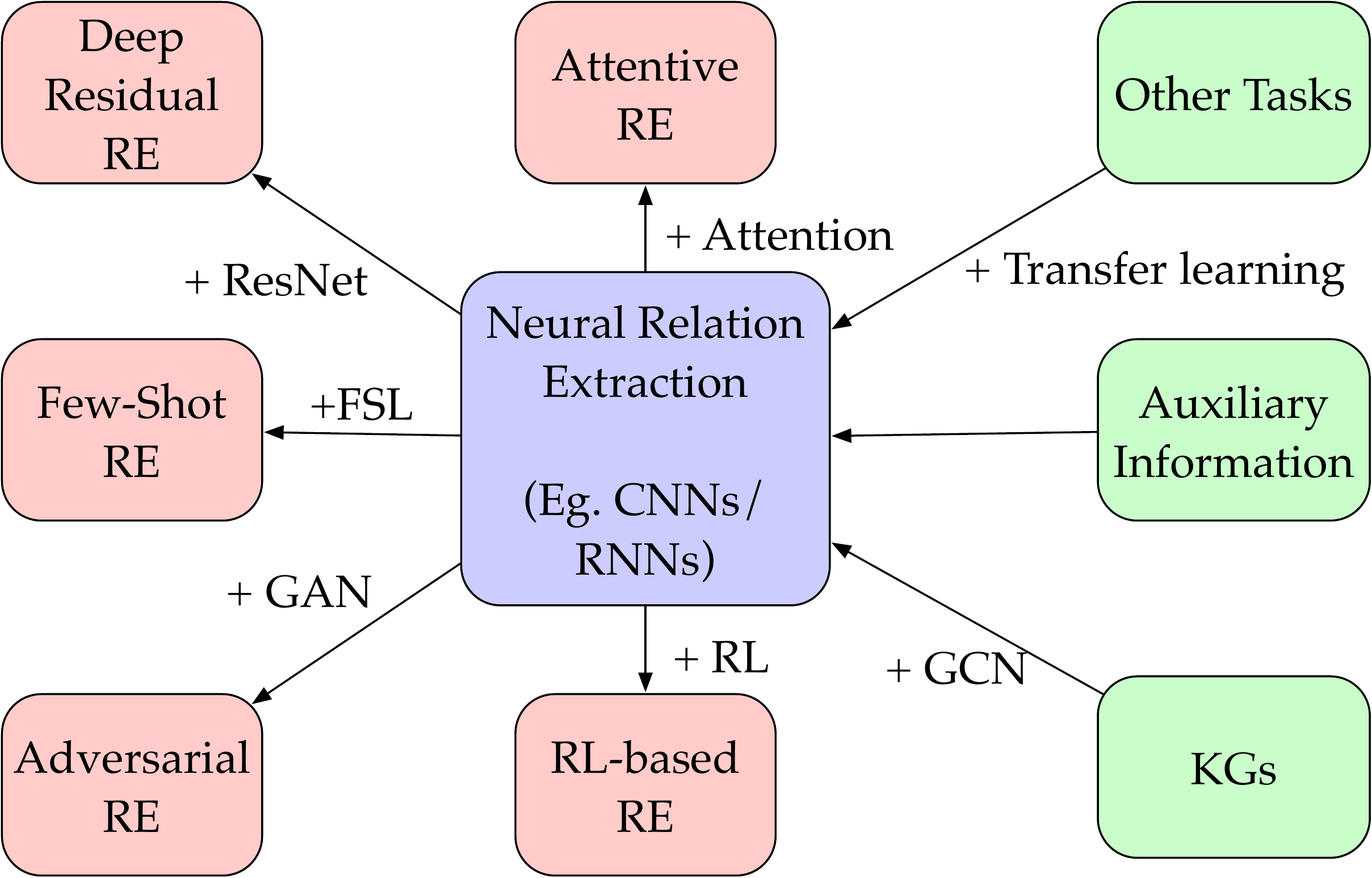}
\caption{An overview of neural relation extraction.}
\label{fig:nre}
\end{center}
\end{figure}

Relation extraction is a key task to build large-scale knowledge graphs automatically by extracting unknown relational facts from plain text and adding them into knowledge graphs.
Due to the lack of labeled relational data, distant supervision~\cite{craven1999constructing}, also referred to as weak supervision or self-supervision, uses heuristic matching to create training data by assuming that sentences containing the same entity mentions may express the same relation under the supervision of a relational database. 
Mintz et al.~\cite{mintz2009distant} adopted the distant supervision for relation classification with textual features, including lexical and syntactic features, named entity tags, and conjunctive features. 
Traditional methods rely highly on feature engineering~\cite{mintz2009distant}, with a recent approach exploring the inner correlation between features~\cite{qu2019discovering}. Deep neural networks are changing the representation learning of knowledge graphs and texts. This section reviews recent advances of neural relation extraction (NRE), with an overview illustrated in Fig.~\ref{fig:nre}. 

\subsubsection{Neural~Relation~Extraction}
Trendy neural networks are widely applied to NRE. 
CNNs with position features of relative distances to entities~\cite{zeng2014relation} are firstly explored for relation classification, and then extended to relation extraction by multi-window CNN~\cite{nguyen2015relation} with multiple sized convolutional filters. 
Multi-instance learning takes a bag of sentences as input to predict the relationship of the entity pair. 
PCNN~\cite{zeng2015distant} applies the piecewise max pooling over the segments of convolutional representation divided by entity position. Compared with vanilla CNN~\cite{zeng2014relation}, PCNN can more efficiently capture the structural information within the entity pair. 
MIMLCNN~\cite{jiang2016relation} further extends it to multi-label learning with cross-sentence max pooling for feature selection. Side information such as class ties~\cite{ye2017jointly} and relation path~\cite{zeng2017incorporating} is also utilized. 
RNNs are also introduced, for example, SDP-LSTM~\cite{xu2015classifying} adopts multi-channel LSTM while utilizing the shortest dependency path between entity pair, and Miwa et al.~\cite{miwa2016end} stacks sequential and tree-structure LSTMs based on dependency tree.
BRCNN~\cite{cai2016bidirectional} combines RNN for capturing sequential dependency with CNN for representing local semantics using two-channel bidirectional LSTM and CNN. 

\subsubsection{Attention~Mechanism}
Many variants of attention mechanisms are combined with CNNs, including word-level attention to capture semantic information of words~\cite{shen2016attention} and selective attention over multiple instances to alleviate the impact of noisy instances~\cite{lin2016neural}. 
Other side information is also introduced for enriching semantic representation. APCNN~\cite{ji2017distant} introduces entity description by PCNN and sentence-level attention, while HATT~\cite{han2018hierarchical} proposes hierarchical selective attention to capture the relation hierarchy by concatenating attentive representation of each hierarchical layer. 
Rather than CNN-based sentence encoders, Att-BLSTM~\cite{zhou2016attention} proposes word-level attention with BiLSTM. 
Recently, Soares et al.~\cite{soares2019matching} utilized pretrained relation representations from the deep Transformers model.

\subsubsection{Graph~Convolutional~Networks (GCNs)}
GCNs are utilized for encoding a dependency tree over sentences or learning KGEs to leverage relational knowledge for sentence encoding.
C-GCN~\cite{zhang2018graph} is a contextualized GCN model over the pruned dependency tree of sentences after path-centric pruning.
AGGCN~\cite{guo2019attention} also applies GCN over the dependency tree, but utilizes multi-head attention for edge selection in a soft weighting manner. 
Unlike previous two GCN-based models, Zhang et al.,~\cite{zhang2019long} applied GCN for relation embedding in knowledge graph for sentence-based relation extraction. The authors further proposed a coarse-to-fine knowledge-aware attention mechanism for the selection of informative instance. 

\subsubsection{Adversarial~Training}
Adversarial Training (AT) is applied to add adversarial noise to word embeddings for CNN- and RNN-based relation extraction under the MIML learning setting~\cite{wu2017adversarial}. 
DSGAN~\cite{qin2018dsgan} denoises distantly supervised relation extraction by learning a generator of sentence-level true positive samples and a discriminator that minimizes the probability of being true positive of the generator. 

\subsubsection{Reinforcement~Learning}
RL has been integrated into neural relation extraction recently by training instance selector with policy networks.
Qin et al.~\cite{qin2018robust} proposed to train policy-based RL agent of sentential relation classifier to redistribute false positive instances into negative samples to mitigate the effect of noisy data. 
The authors took the F1 score as an evaluation metric and used F1 score based performance change as the reward for policy networks. Similarly, Zeng et al.~\cite{zeng2018large} and Feng et al.~\cite{feng2018reinforcement} proposed different reward strategies. 
The advantage of RL-based NRE is that the relation extractor is model-agnostic. Thus, it could be easily adapted to any neural architectures for effective relation extraction. 
Recently, HRL~\cite{takanobu2018hierarchical} proposed a hierarchical policy learning framework of high-level relation detection and low-level entity extraction.
 
\subsubsection{Other~Advances}
Other advances of deep learning are also applied for neural relation extraction.
Noticing that current NRE methods do not use very deep networks, Huang and Wang~\cite{huang2017deep} applied deep residual learning to noisy relation extraction and found that 9-layer CNNs have improved performance. 
Liu et al.~\cite{liu2018neural} proposed to initialize the neural model by transfer learning from entity classification.
The cooperative CORD~\cite{lei2018cooperative} ensembles text corpus and knowledge graph with external logical rules by bidirectional knowledge distillation and adaptive imitation. 
TK-MF~\cite{jiang2019relation} enriches sentence representation learning by matching sentences and topic words. 
Recently, Shahbazi et al.~\cite{shahbazi2020relation} studied trustworthy relation extraction by benchmarking several explanation mechanisms, including saliency, gradient $\times$ input, and leave one out.

The existence of low-frequency relations in knowledge graphs requires few-shot relation classification with unseen classes or only a few instances. 
Gao et al.~\cite{gao2019hybrid} proposed hybrid attention-based prototypical networks to compute prototypical relation embedding and compare its distance between the query embedding.
Qin et al.~\cite{qu2020few} explored the relationships between relations with a global relation graph and formulated few-shot relation extraction as a Bayesian meta-learning problem to learn the posterior distribution of relations' prototype vectors.

\subsubsection{Joint Entity and Relation Extraction}
Traditional relation extraction models utilize pipeline approaches by first extracting entity mentions and then classifying relations. However, pipeline methods may cause error accumulation.
Several studies show better performance by joint learning~\cite{miwa2014modeling, miwa2016end} than by conventional pipeline methods. 
Katiyar and Cardie~\cite{katiyar2017going} proposed a joint extraction framework with an attention-based LSTM network. 
Some convert joint extraction into different problems such as sequence labeling via a novel tagging scheme~\cite{zheng2017joint} and multi-turn question answering~\cite{li2019entity}. Challenges remain in dealing with entity pair and relation overlapping~\cite{dai2019joint}. 
Wei et al.~\cite{wei2020novel} proposed a cascade binary tagging framework that models relations as subject-object mapping functions to solve the overlapping problem. 

There is a distribution discrepancy between training and inference in the joint learning framework, leading to exposure bias.
Recently, Wang et al.~\cite{wang2020tplinker} proposed a one-stage joint extraction framework by transforming joint entity and relation extraction into a token pair linking task to mitigate error propagation and exposure bias. 
In contrast to the common view that joint models can ease error accumulation by capturing mutual interaction of entities and relations, Zhong and Chen~\cite{zhong2020frustratingly} proposed a simple pipeline-based yet effective approach to learning two independent encoders for entities and relations, revealing that strong contextual representation can preserve distinct features of entities and relations.  
Future research needs to rethink the relation between the pipeline and joint learning methods.

\subsection{Summary}
This section reviews knowledge completion for incomplete knowledge graph and acquisition from plain text. 

\textit{Knowledge graph completion} completes missing links between existing entities or infers entities given entity and relation queries. 
Embedding-based KGC methods generally rely on triple representation learning to capture semantics and do candidate ranking for completion. 
Embedding-based reasoning remains in individual relation level, and is poor at complex reasoning because it ignores the symbolical nature of knowledge graph, and lack of interpretability. 
Hybrid methods with symbolics and embedding incorporate rule-based reasoning, overcome the sparsity of knowledge graph to improve the quality of embedding, facilitate efficient rule injection, and induce interpretable rules. 
With the observation of the graphical nature of knowledge graphs, path search and neural path representation learning are studied. However, they suffer from connectivity deficiency when traverses over large-scale graphs.
The emerging direction of meta relational learning aims to learn fast adaptation over unseen relations in low-resource settings. 

\textit{Entity discovery} acquires entity-oriented knowledge from text and fuses knowledge between knowledge graphs.
There are several categories according to specific settings. 
Entity recognition is explored in a sequence-to-sequence manner, entity typing discusses noisy type labels and zero-shot typing, and entity disambiguation and alignment learn unified embeddings with iterative alignment model proposed to tackle the issue of a limited number of alignment seeds.
However, it may face error accumulation problems if newly-aligned entities suffer from poor performance. 
Language-specific knowledge has increased in recent years and consequentially motivates the research on cross-lingual knowledge alignment.

\textit{Relation extraction} suffers from noisy patterns under the assumption of distant supervision, especially in text corpus of different domains. Thus, weakly supervised relation extraction must mitigate the impact of noisy labeling. For example, multi-instance learning takes bags of sentences as inputs and attention mechanism~\cite{lin2016neural} reduce noisy patterns by soft selection over instances, and RL-based methods formulate instance selection as a hard decision.
Another principle is to learn richer representation as possible. As deep neural networks can solve error propagation in traditional feature extraction methods, this field is dominated by DNN-based models, as summarized in Table~\ref{tab:NRE}.

\begin{table*}[htp]
\scriptsize
\caption{A summary of neural relation extraction and recent advances.}
\begin{center}
\begin{tabular}{ l l p{6cm} l }
\toprule
Category & Method & Mechanism & Auxiliary Information\\
\midrule
\multirow{6}{30pt}{CNNs}	& O-CNN~\cite{zeng2014relation} & CNN + max pooling & position embedding\\
		 	   & Multi CNN~\cite{nguyen2015relation} & Multi-window convolution + max pooling & position embedding\\
		    & PCNN~\cite{zeng2015distant} & CNN + piecewise max pooling & position embedding\\
				  & MIMLCNN~\cite{jiang2016relation} & CNN + piecewise and cross-sentence max pooling & position embedding\\
				  & Ye et al.~\cite{ye2017jointly} & CNN/PCNN + pairwise ranking & position embedding, class ties \\
				  & Zeng et al.~\cite{zeng2017incorporating} & CNN + max pooling & position embedding, relation path\\
				
\hdashline
\multirow{3}{30pt}{RNNs} & SDP-LSTM~\cite{xu2015classifying} & Multichannel LSTM + dropout & dependency tree, POS, GR, hypernyms \\
   		  & LSTM-RNN~\cite{miwa2016end}& Bi-LSTM + Bi-TreeLSTM & POS, dependency tree \\
				  & BRCNN~\cite{cai2016bidirectional} & Two-channel LSTM + CNN + max pooling & dependency tree, POS, NER\\
\hdashline
\multirow{5}{30pt}{Attention}		& Attention-CNN~\cite{shen2016attention} & CNN + word-level attention + max pooling & POS, position embedding \\
		 	& Lin et al.~\cite{lin2016neural} & CNN/PCNN + selective attention + max pooling & position embedding \\
			& Att-BLSTM~\cite{zhou2016attention} & Bi-LSTM + word-level attention & position indicator\\
 	  & APCNN~\cite{ji2017distant}& PCNN + sentence-level attention & entity descriptions\\
			& HATT~\cite{han2018hierarchical} & CNN/PCNN + hierarchical attention & position embedding, relation hierarchy\\
\hdashline
\multirow{3}{30pt}{GCNs} & C-GCN~\cite{zhang2018graph} & LSTM + GCN + path-centric pruning & dependency tree \\
 		& KATT~\cite{zhang2019long} & Pre-training + GCN + CNN + attention & position embedding, relation hierarchy\\
			& AGGCN~\cite{guo2019attention} & GCN + multi-head attention + dense layers & dependency tree\\
\hdashline
\multirow{2}{38pt}{Adversarial}			& Wu et al.~\cite{wu2017adversarial} & AT + PCNN/RNN + selective attention & indicator encoding \\
	  & DSGAN~\cite{qin2018dsgan} & GAN + PCNN/CNN + attention & position embedding\\
\hdashline
\multirow{4}{30pt}{RL}			& Qin et al.~\cite{qin2018robust}& Policy gradient + CNN + performance change reward & position embedding \\
    & Zeng et al.~\cite{zeng2018large}& Policy gradient + CNN + +1/-1 bag-result reward & position embedding \\
			& Feng et al.~\cite{feng2018reinforcement}& Policy gradient + CNN + predictive probability reward & position embedding \\
			& HRL~\cite{takanobu2018hierarchical} & Hierarchical policy learning + Bi-LSTM + MLP & relation indicator \\
\bottomrule
\end{tabular}
\end{center}
\label{tab:NRE}
\end{table*}%

\section{Temporal~Knowledge~Graph}
\label{sec:temporal}
Current knowledge graph research mostly focuses on static knowledge graphs where facts are not changed with time, while the temporal dynamics of a knowledge graph is less explored. However, the temporal information is of great importance because the structured knowledge only holds within a specific period, and the evolution of facts follows a time sequence. Recent research begins to take temporal information into KRL and KGC, which is termed as \textit{temporal knowledge graph} in contrast to the previous static knowledge graph. 
Research efforts have been made for learning temporal and relational embedding simultaneously.
Relevant models for dynamic network embedding also inspire temporal knowledge graph embedding. For example, the temporal graph attention (TGAT) network \cite{xu2020inductive} that captures temporal-topological structure and learn time-feature interactions simultaneously may be useful to preserve temporal-aware relation for knowledge graphs.

\subsection{Temporal Information Embedding}
Temporal information is considered in temporal-aware embedding by extending triples into temporal quadruple as $(h, r, t, \tau)$, where $\tau$ provides additional temporal information about when the fact held. 
Leblay and Chekol~\cite{leblay2018deriving} investigated temporal scope prediction over time-annotated triple, and simply extended existing embedding methods, for example, TransE with the vector-based TTransE defined as 
\begin{equation}\small
f_\tau(h,r,t)=-\|\mathbf{h}+\mathbf{r}+\mathbf{\tau}-\mathbf{t}\|_{L_{1 / 2}}.
\end{equation}
Ma et al.~\cite{ma2019embedding} also generalized existing static embedding methods and proposed ConT by replacing the shared weight vector of Tucker with a timestamp embedding.
Temporally scoped quadruple extends triples by adding a time scope $[\tau_s, \tau_e]$, where $\tau_s$ and $\tau_e$ stand for the beginning and ending of the valid period of a triple, and then a static subgraph $G_\tau$ can be derived from the dynamic knowledge graph when given a specific timestamp $\tau$. HyTE~\cite{dasgupta2018hyte} takes a time stamp as a hyperplane $\mathbf{w_{\tau}}$ and projects entity and relation representation as 
$P_{\tau}\left(\mathbf{h}\right) =\mathbf{h}-\left(\mathbf{w}_{\tau}^{\top} \mathbf{h}\right) \mathbf{w_{\tau}}$, $P_{\tau}\left(\mathbf{t}\right) =\mathbf{t}-\left(\mathbf{w}_{\tau}^{\top} \mathbf{t}\right) \mathbf{w}_{\tau}$, and $P_{\tau}\left(\mathbf{r}\right) =\mathbf{r}-\left(\mathbf{w}_{\tau}^{\top} \mathbf{r}\right) \mathbf{w}_{\tau}$.
The temporally projected scoring function is calculated as
\begin{equation}\small
\label{eq:HyTE}
f_{\tau}(h, r, t)=\left\|P_{\tau}\left(\mathbf{h}\right)+P_{\tau}\left(\mathbf{r}\right)-P_{\tau}\left(\mathbf{t}\right)\right\|_{L_{1} / L_{2}}
\end{equation}
within the projected translation of $P_{\tau}\left(\mathbf{h}\right)+P_{\tau}\left(\mathbf{r}\right)\approx P_{\tau}\left(\mathbf{t}\right)$.
Garc\'{i}a-Dur\'{a}n et al.~\cite{garcia2018learning} concatenated predicate token sequence and temporal token sequence, and used LSTM to encode the concatenated time-aware predicate sequences. The last hidden state of LSTM is taken as temporal-aware relational embedding $r_{temp}$. The scoring function of extended TransE and DistMult are calculated as $\left\|\mathbf{h}+\mathbf{r}_{temp}-\mathbf{t}\right\|_{2}$ and $\left(\mathbf{h} \circ \mathbf{t}\right) \mathbf{r}_{temp}^{T}$, respectively. 
By defining the context of an entity $e$ as an aggregate set of facts containing $e$, Liu et al.~\cite{liu2019context} proposed context selection to capture useful contexts, and measured temporal consistency with selected context.
By formulating temporal KGC as 4-order tensor completion, Lacroix et al.~\cite{lacroix2020tensor} proposed TComplEx, which extends ComplEx decomposition, and introduced weighted regularizers. 

\subsection{Entity~Dynamics}
Real-world events change entities' state, and consequently, affect the corresponding relations. 
To improve temporal scope inference, the contextual temporal profile model~\cite{wijaya2014ctps} formulates the temporal scoping problem as state change detection and utilizes the context to learn state and state change vectors. 
Inspired by the diachronic word embedding, Goel et al.~\cite{goel2020diachronic} took an entity and timestamp as the input of entity embedding function to preserve the temporal-aware characteristics of entities at any time point.
Know-evolve~\cite{trivedi2017know}, a deep evolutionary knowledge network, investigates the knowledge evolution phenomenon of entities and their evolved relations. A multivariate temporal point process is used to model the occurrence of facts, and a novel recurrent network is developed to learn the representation of non-linear temporal evolution.
To capture the interaction between nodes, RE-NET~\cite{jin2019recurrent} models event sequences via RNN-based event encoder, and neighborhood aggregator. Specifically, RNN is used to capture the temporal entity interaction, and the neighborhood aggregator aggregates the concurrent interactions. 

\subsection{Temporal Relational Dependency}
There exists temporal dependencies in relational chains following the timeline, for example, $\texttt{wasBornIn} \rightarrow \texttt{graduateFrom} \rightarrow \texttt{workAt} \rightarrow \texttt{diedIn}$.
Jiang et al.~\cite{jiang2016towards, jiang2016encoding} proposed time-aware embedding, a joint learning framework with temporal regularization, to incorporate temporal order and consistency information. The authors defined a temporal scoring function as
\begin{equation}\small
\label{eq:temporal}
f\left(\left\langle r_{k}, r_{l}\right\rangle\right)=\left\|\mathbf{r}_{k} \mathbf{T}-\mathbf{r}_{l}\right\|_{L_{1/2}},
\end{equation}
where $\mathbf{T}\in \mathbb{R}^{d\times d}$ is an asymmetric matrix that encodes the temporal order of relation, for a temporal ordering relation pair $\left\langle r_{k}, r_{l}\right\rangle$.
Three temporal consistency constraints of disjointness, ordering, and spans are further applied by integer linear programming formulation. 

\subsection{Temporal Logical Reasoning}
Logical rules are also studied for temporal reasoning. Chekol et al.~\cite{chekol2017marrying} explored Markov logic network and probabilistic soft logic for reasoning over uncertain temporal knowledge graphs.
RLvLR-Stream~\cite{omran2019embedding} considers temporal close-path rules and learns the structure of rules from the knowledge graph stream for reasoning. 

\section{Knowledge-Aware~Applications}
\label{sec:application}
Rich structured knowledge can be useful for AI applications. 
However, how to integrate such symbolic knowledge into the computational framework of real-world applications remains a challenge. 
The application of knowledge graphs includes two folds: 1) in-KG applications such as link prediction and named entity recognition; and 2) out-of-KG applications, including relation extraction and more downstream knowledge-aware applications such as question answering and recommendation systems.
This section introduces several recent DNN-based knowledge-driven approaches with the applications on natural language processing and recommendation. More miscellaneous applications such as digital health and search engine are introduced in Appendix~\ref{sup:more-app}.

\subsection{Language~Representation~Learning}
Language representation learning via self-supervised language model pretraining has become an integral component of many NLP systems. 
Traditional language modeling does not exploit factual knowledge with entities frequently observed in the text corpus. 
How to integrate knowledge into language representation has drawn increasing attention. 
Knowledge graph language model (KGLM)~\cite{logan2019barack} learns to render knowledge by selecting and copying entities. 
ERNIE-Tsinghua~\cite{zhang2019ernie} fuses informative entities via aggregated pre-training and random masking. 
K-BERT~\cite{weijie2019kbert} infuses domain knowledge into BERT contextual encoder.
ERNIE-Baidu~\cite{sun2019ernie} introduces named entity masking and phrase masking to integrate knowledge into the language model and is further improved by ERNIE 2.0~\cite{sun2020ernie} via continual multi-task learning.
To capture factual knowledge from text, KEPLER~\cite{wang2020kepler} combines knowledge embedding and masked language modeling losses via joint optimization. 
GLM~\cite{shen2020exploiting} proposes a graph-guided entity masking scheme to utilize knowledge graph implicitly.  
CoLAKE~\cite{sun2020colake} further exploits the knowledge context of an entity through a unified word-knowledge graph and a modified Transformer encoder. 
Similar to the K-BERT model and focusing on the medical corpus, BERT-MK~\cite{he2020integrating} integrates medical knowledge into the pretraining language model via knowledge subgraph extraction.
Rethinking about large-scale training on language model and querying over knowledge graphs, Petroni et al.~\cite{petroni2019language} analyzed the language model and knowledge base. They found that certain factual knowledge can be acquired via pre-training language model.

\subsection{Question~Answering}
Knowledge-graph-based question answering (KG-QA) answers natural language questions with facts from knowledge graphs. Neural network-based approaches represent questions and answers in distributed semantic space, and some also conduct symbolic knowledge injection for commonsense reasoning. 

\subsubsection{Single-fact QA}
Taking a knowledge graph as an external intellectual source, simple factoid QA or single-fact QA is to answer a simple question involving a single knowledge graph fact.
Dai et al.~\cite{dai2016cfo} proposed a conditional focused neural network equipped with focused pruning to reduce the search space.
BAMnet~\cite{chen2019bidirectional} models the two-way interaction between questions and knowledge graph with a bidirectional attention mechanism. 
Although deep learning techniques are intensively applied in KG-QA, they inevitably increase the model complexity. 
Through the evaluation of simple KG-QA with and without neural networks, Mohammed et al.~\cite{mohammed2018strong} found that sophisticated deep models such as LSTM and GRU with heuristics achieve state of the art, and non-neural models also gain reasonably well performance. 

\subsubsection{Multi-hop Reasoning}
To deal with complex multi-hop relation requires a more dedicated design to be capable of multi-hop commonsense reasoning. 
Structured knowledge provides informative commonsense observations and acts as relational inductive biases, which boosts recent studies on commonsense knowledge fusion between symbolic and semantic space for multi-hop reasoning.
Bauer et al.~\cite{bauer2018commonsense} proposed multi-hop bidirectional attention and pointer-generator decoder for effective multi-hop reasoning and coherent answer generation, utilizing external commonsense knowledge by relational path selection from ConceptNet and injection with selectively-gated attention. 
Variational Reasoning Network (VRN)~\cite{zhang2018variational} conducts multi-hop logic reasoning with reasoning-graph embedding, while handles the uncertainty in topic entity recognition.
KagNet~\cite{lin2019kagnet} performs concept recognition to build a schema graph from ConceptNet and learns path-based relational representation via GCN, LSTM, and hierarchical path-based attention. 
CogQA~\cite{ding2019cognitive} combines implicit extraction and explicit reasoning and proposes a cognitive graph model based on BERT and GNN for multi-hop QA.

\subsection{Recommender~Systems}
\label{sec:recommender}
Integrating knowledge graphs as external information enables recommendation systems to have the ability of commonsense reasoning, with the potential to solve the sparsity issue and the cold start problem. 
By injecting knowledge-graph-based side information such as entities, relations, and attributes, many efforts work on embedding-based regularization to improve recommendation. 
The collaborative CKE~\cite{zhang2016collaborative} jointly trains KGEs, item's textual information, and visual content via translational KGE model and stacked auto-encoders. 
Noticing that time-sensitive and topic-sensitive news articles consist of condensed entities and common knowledge, DKN~\cite{wang2018dkn} incorporates knowledge graph by a knowledge-aware CNN model with multi-channel word-entity-aligned textual inputs. 
However, DKN cannot be trained in an end-to-end manner as it needs to learn entity embedding in advance. To enable end-to-end training, MKR~\cite{wang2019multi} associates multi-task knowledge graph representation and recommendation by sharing latent features and modeling high-order item-entity interaction. 
While other works consider the relational path and structure of knowledge graphs,
KPRN~\cite{wang2019explainable} regards the interaction between users and items as an entity-relation path in the knowledge graph and conducts preference inference over the path with LSTM to capture the sequential dependency. PGPR~\cite{xian2019reinforcement} performs reinforcement policy-guided path reasoning over knowledge-graph-based user-item interaction. 
KGAT~\cite{wang2019kgat} applies graph attention network over the collaborative knowledge graph of entity-relation and user-item graphs to encode high-order connectivities via embedding propagation and attention-based aggregation.
Knowledge graph-based recommendation inherently processes interpretability from embedding propagation with multi-hop neighbors in the knowledge graph.

\section{Future~Directions}
Many efforts have been conducted to tackle the challenges of knowledge representation and its related applications. 
However, there remains several formidable open problems and promising future directions.

\subsection{Complex~Reasoning} 
Numerical computing for knowledge representation and reasoning requires a continuous vector space to capture the semantic of entities and relations. 
While embedding-based methods have a limitation on complex logical reasoning, two directions on the relational path and symbolic logic are worthy of being further explored. 
Some promising methods such as recurrent relational path encoding, GNN-based message passing over knowledge graph, and reinforcement learning-based pathfinding and reasoning are up-and-coming for handling complex reasoning. 
For the combination of logic rules and embeddings, recent works~\cite{qu2019probabilistic, zhang2020efficient} combine Markov logic networks with KGE, aiming to leverage logic rules and handling their uncertainty. Enabling probabilistic inference for capturing the uncertainty and domain knowledge with efficiently embedding will be a noteworthy research direction.

\subsection{Unified~Framework}
Several representation learning models on knowledge graphs have been verified as equivalence, for example,
Hayshi and Shimbo~\cite{hayashi2017equivalence} proved that HolE and ComplEx are mathematically equivalent for link prediction with a particular constraint. 
ANALOGY~\cite{liu2017analogical} provides a unified view of several representative models, including DistMult, ComplEx, and HolE.
Wang et al.~\cite{wang2018multi} explored connections among several bilinear models.
Chandrahas et al.~\cite{sharma2018towards} explored the geometric understanding of additive and multiplicative KRL models.
Most works formulated knowledge acquisition KGC and relation extraction separately with different models. 
Han et al.~\cite{han2018neural} put them under the same roof and proposed a joint learning framework with mutual attention for information sharing between knowledge graph and text.
A unified understanding of knowledge representation and reasoning is less explored. An investigation towards unification in a way similar to the unified framework of graph networks~\cite{battaglia2018relational}, however, will be worthy of bridging the research gap.

\subsection{Interpretability}
Interpretability of knowledge representation and injection is a vital issue for knowledge acquisition and real-world applications. Preliminary efforts have been made for interpretability. ITransF~\cite{xie2017interpretable} uses sparse vectors for knowledge transferring and interprets with attention visualization. CrossE~\cite{zhang2019interaction} explores the explanation scheme of knowledge graphs by using embedding-based path searching to generate explanations for link prediction. However, recent neural models have limitations on transparency and interpretability, although they have gained impressive performance. Some methods combine black-box neural models and symbolic reasoning by incorporating logical rules to increase the interoperability. Interpretability can convince people to trust predictions. Thus, further work should go into interpretability and improve the reliability of predicted knowledge.

\subsection{Scalability}
Scalability is crucial in large-scale knowledge graphs. 
There is a trade-off between computational efficiency and model expressiveness, with a limited number of works applied to more than 1 million entities. 
Several embedding methods use simplification to reduce the computation cost, such as simplifying tensor products with circular correlation operation~\cite{nickel2016holographic}. However, these methods still struggle with scaling to millions of entities and relations. 

Probabilistic logic inference using Markov logic networks is computationally intensive, making it hard to scalable to large-scale knowledge graphs. 
Rules in a recent neural logical model~\cite{qu2019probabilistic} are generated by simple brute-force search, making it insufficient on large-scale knowledge graphs. ExpressGNN~\cite{zhang2020efficient} attempts to use NeuralLP~\cite{yang2017differentiable} for efficient rule induction. 
Nevertheless, there still has a long way to go to deal with cumbersome deep architectures and the increasingly growing knowledge graphs.

\subsection{Knowledge~Aggregation}
The aggregation of global knowledge is the core of knowledge-aware applications. For example, recommendation systems use a knowledge graph to model user-item interaction and text classification jointly to encode text and knowledge graph into a semantic space.
Most current knowledge aggregation methods design neural architectures such as attention mechanisms and GNNs. 
The natural language processing community has been boosted from large-scale pre-training via transformers and variants like BERT models. At the same time, a recent finding~\cite{petroni2019language} reveals that the pre-training language model on the unstructured text can acquire certain factual knowledge. Large-scale pre-training can be a straightforward way to injecting knowledge. However, rethinking the way of knowledge aggregation in an efficient and interpretable manner is also of significance.

\subsection{Automatic Construction and Dynamics}
Current knowledge graphs rely highly on manual construction, which is labor-intensive and expensive. The widespread applications of knowledge graphs on different cognitive intelligence fields require automatic knowledge graph construction from large-scale unstructured content. Recent research mainly works on semi-automatic construction under the supervision of existing knowledge graphs. Facing the multimodality, heterogeneity, and large-scale application, automatic construction is still of great challenge. 

The mainstream research focuses on static knowledge graphs, with several works on predicting temporal scope validity and learning temporal information and entity dynamics. Many facts only hold within a specific period. A dynamic knowledge graph, together with learning algorithms capturing dynamics, can address the limitation of traditional knowledge representation and reasoning by considering the temporal nature.

\section{Conclusion}
Knowledge graphs as the ensemble of human knowledge have attracted increasing research attention, with the recent emergence of knowledge representation learning, knowledge acquisition methods, and a wide variety of knowledge-aware applications. 
The paper conducts a comprehensive survey on the following four scopes: 1) knowledge graph embedding, with a full-scale systematic review from embedding space, scoring metrics, encoding models, embedding with external information, and training strategies; 2) knowledge acquisition of entity discovery, relation extraction, and graph completion from three perspectives of embedding learning, relational path inference and logical rule reasoning; 3) temporal knowledge graph representation learning and completion; 4) real-world knowledge-aware applications on natural language understanding, recommendation systems, question answering and other miscellaneous applications.
Besides, some useful resources of datasets and open-source libraries, and future research directions are introduced and discussed. Knowledge graph hosts a large research community and has a wide range of methodologies and applications. We conduct this survey to have a summary of current representative research efforts and trends and expect it can facilitate future research.

\appendices
\section{A Brief History of Knowledge Bases}
\label{sup:history}
\begin{figure*}[htbp]
\begin{center}
\caption{A brief history of knowledge bases}
\includegraphics[width=0.8\textwidth]{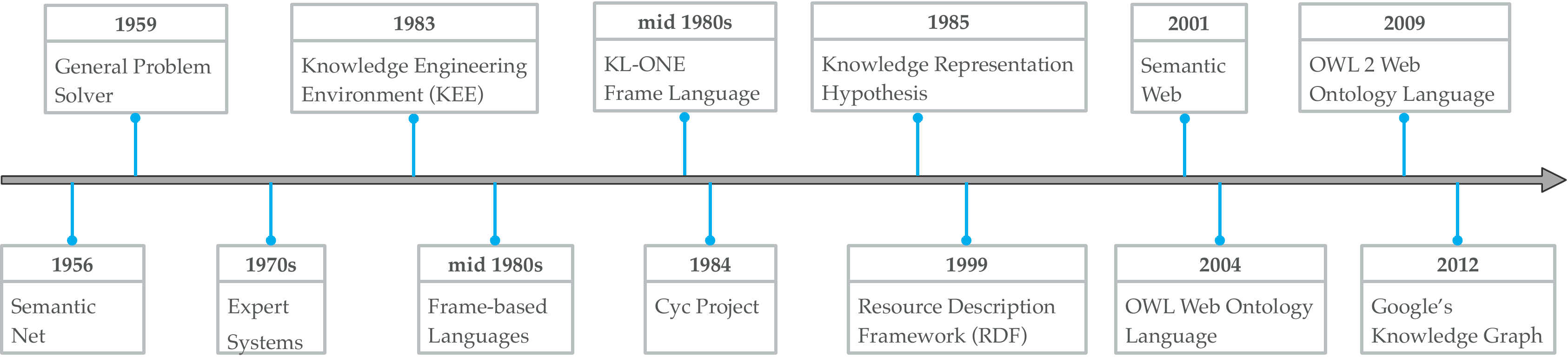}
\label{fig:history}
\end{center}
\end{figure*}

Figure \ref{fig:history} illustrates a brief history of knolwedge bases. 

\section{Mathematical~Operations}
\label{sup:math}
Hermitian dot product (Eq.~\ref{eq:Hermitian}) and Hamilton product (Eq.~\ref{eq:Hamilton}) are used in complex vector space (Sec.~\ref{sec:space-complex}).
Given $\mathbf{h}$ and $\mathbf{t}$ represented in complex space $\mathbb{C}^d$, the Hermitian dot product $\langle,\rangle : \mathbb{C}^d \times \mathbb{C}^d \longrightarrow \mathbb{C}$ is calculated as the sesquilinear form of
\begin{equation}\small\label{eq:Hermitian}
  \langle \mathbf{h}, \mathbf{t}\rangle =\overline{\mathbf{h}}^{T} \mathbf{t},
\end{equation}
where $\overline{\mathbf{h}}=\operatorname{Re}(\mathbf{h})-i \operatorname{Im}(\mathbf{h})$ is the conjugate operation over $\mathbf{h}\in \mathbb{C}^d$.
The quaternion extends complex numbers into four-dimensional hypercomplex space. With two $d$-dimensional quaternions defined as $\mathbf{Q_{1}}=\mathbf{a_{1}}+\mathbf{b_{1}} \mathbf{i}+\mathbf{c_{1}} \mathbf{j}+\mathbf{d_{1}} \mathbf{k}$ and $\mathbf{Q_2}=\mathbf{a_{2}}+\mathbf{b_{2}} \mathbf{i}+\mathbf{c_{2}} \mathbf{j}+\mathbf{d_{2}} \mathbf{k}$, the Hamilton product $\otimes: \mathbb{H}^{d} \times \mathbb{H}^{d} \rightarrow \mathbb{H}^{d}$ is defined as
\begin{equation}\small\label{eq:Hamilton}
\begin{aligned} 
  \mathbf{Q_{1}} \otimes \mathbf{Q_{2}} &=\left(\mathbf{a_{1}}\circ \mathbf{a_{2}}-\mathbf{b_{1}}\circ \mathbf{b_{2}}-\mathbf{c_{1}} \circ \mathbf{c_{2}}-\mathbf{d_{1}}\circ \mathbf{d_{2}}\right)\\ &+\left(\mathbf{a_{1}}\circ \mathbf{b_{2}}+\mathbf{b_{1}}\circ \mathbf{a_{2}}+\mathbf{c_{1}}\circ \mathbf{d_{2}}-\mathbf{d_{1}}\circ \mathbf{c_{2}}\right) \mathbf{i} \\ &+\left(\mathbf{a_{1}}\circ \mathbf{c_{2}}-\mathbf{b_{1}}\circ \mathbf{d_{2}}+\mathbf{c_{1}}\circ \mathbf{a_{2}}+\mathbf{d_{1}} \mathbf{b_{2}}\right) \mathbf{j} \\ &+\left(\mathbf{a_{1}}\circ \mathbf{d_{2}}+\mathbf{b_{1}}\circ \mathbf{c_{2}}-\mathbf{c_{1}}\circ \mathbf{b_{2}}+\mathbf{d_{1}}\circ \mathbf{a_{2}}\right) \mathbf{k}. 
\end{aligned}
\end{equation}{}
The Hadmard product (Eq.~\ref{eq:Hadmard}) and circular correlation (Eq.~\ref{eq:circular-correlation}) are utilized in semantic matching based methods (Sec.~\ref{sec:semantic}).
Hadmard product, denoted as $\circ$ or $\odot: \mathbb{R}^{d} \times \mathbb{R}^{d} \rightarrow \mathbb{R}^{d}$, is also known as element-wise product or Schur product. 
\begin{equation}\small\label{eq:Hadmard}
(\mathbf{h} \circ \mathbf{t})_{i}=(\mathbf{h} \odot \mathbf{t})_{i}=(\mathbf{h})_{i}(\mathbf{t})_{i}
\end{equation}
Circular correlation $\star: \mathbb{R}^{d} \times \mathbb{R}^{d} \rightarrow \mathbb{R}^{d}$ is an efficient computation calculated as:
\begin{equation}\small\label{eq:circular-correlation}
  [\boldsymbol{a} \star \boldsymbol{b}]_{k}=\sum_{i=0}^{d-1} a_{i} b_{(k+i) \bmod d}.
\end{equation}

\section{A Summary of KRL Models}
\label{sup:KRL-summary}

\begin{table*}[!ht]
\scriptsize
\caption{A comprehensive summary of knowledge representation learning models}
\begin{center}
\renewcommand{\arraystretch}{1.25}
\begin{tabular}{ l l l l l }
\toprule
Category & Model & Ent. embed. & Rel. embed. & Scoring Function $f_r(h,t)$ \\
\midrule
\multirow{2}{30pt}{Polar coordinate} &	\multirow{2}{30pt}{HAKE~\cite{zhang2020learning}}	&	$\mathbf{h}_{m}, \mathbf{t}_{m} \in \mathbb{R}^{k}$	&	$\mathbf{r}_{m} \in \mathbb{R}_{+}^{k}$	&	\multirow{2}{150pt}{$-\left\|\mathbf{h}_{m} \circ \mathbf{r}_{m}-\mathbf{t}_{m}\right\|_{2}-\lambda\left\|\sin \left(\left(\mathbf{h}_{p}+\mathbf{r}_{p}-\mathbf{t}_{p}\right) / 2\right)\right\|_{1}$	}\\
& & $\mathbf{h}_{p}, \mathbf{t}_{p} \in[0,2 \pi)^{k}$ & $\mathbf{r}_{p}, \in[0,2 \pi)^{k}$ & \\
\hline	
\multirow{3}{30pt}{Complex vector}	&	ComplEx~\cite{trouillon2016complex}	&	$\mathbf{h}, \mathbf{t} \in \mathbb{C}^{d}$	&	$\mathbf{r} \in \mathbb{C}^{d}$	&	$\operatorname{Re}\left(<\mathbf{r}, \mathbf{h}, \overline{\mathbf{t}}>\right)=\operatorname{Re}\left(\sum_{k=1}^{K} \mathbf{r}_k \mathbf{h}_{k} \overline{\mathbf{t}}_{k}\right) $	\\
	&	RotatE~\cite{sun2018rotate}	&	$\mathbf{h}, \mathbf{t} \in \mathbb{C}^{d}$	&	$\mathbf{r} \in \mathbb{C}^{d}$	&	$\|\mathbf{h} \circ \mathbf{r}-\mathbf{t}\|$	\\
	&	QuatE~\cite{zhang2019quaternion}	&	$\mathbf{h}, \mathbf{t} \in \mathbb{H}^d$	&	$\mathbf{r} \in \mathbb{H}^d$ 	&	$\mathbf{h} \otimes \frac{\mathbf{r}}{|\mathbf{r}|} \cdot \mathbf{t}$	\\
\hline									
\multirow{3}{30pt}{Manifold \& Group}	&	ManifoldE~\cite{xiao2016one}	&	$\mathbf{h}, \mathbf{t} \in \mathbb{R}^{d}$	&	$\mathbf{r} \in \mathbb{R}^{d}$	&	$\left\|\mathcal{M}(h, r, t)-D_{r}^{2}\right\|^{2}$	\\
	&	TorusE~\cite{ebisu2018toruse}	&	$[\mathbf{h}], [\mathbf{t}] \in \mathbb{T}^{n}$	&	$[\mathbf{r}] \in \mathbb{T}^{n}$	&	 $\min _{(x, y) \in([h]+[r]) \times[t]}\|x-y\|_{i}$	\\
	&	DihEdral~\cite{xu2019relation}	&	$\mathbf{h}^{(l)}, \mathbf{t}^{(l)} \in \mathbb{R}^{2}$	&	$\mathbf{R}^{(l)} \in \mathbb{D}_{K}$	&	$\sum_{l=1}^{L} \mathbf{h}^{(l) \top} \mathbf{R}^{(l)} \mathbf{t}^{(l)}$	\\
	&	MuRP~\cite{balazevic2019multi}	&	$\mathbf{h}, \mathbf{t} \in \mathbb{B}_{c}^{d}, b_{h}, b_{t} \in \mathbb{R}$	&	$\mathbf{r} \in \mathbb{B}_{c}^{d}$	&	$-d_{\mathbb{B}}\left(\exp _{\mathbf{0}}^{c}\left(\mathbf{R} \log _{\mathbf{0}}^{c}\left(\mathbf{h}\right)\right), \mathbf{t} \oplus_{c} \mathbf{r}\right)^{2}+b_{h}+b_{t}$	\\
	&	AttH~\cite{chami2020low}	&	$\mathbf{h}, \mathbf{t} \in \mathbb{B}_{c}^{d}, b_{h}, b_{t} \in \mathbb{R}$	&	$\mathbf{r} \in \mathbb{B}_{c}^{d}$	&	$-d_{\mathbb{B}}^{c_{r}}\left(\operatorname{Att}\left(\mathbf{q}_{\mathrm{Rot}}^{H}, \mathbf{q}_{\mathrm{Ref}}^{H} ; \mathbf{a}_{r}\right) \oplus^{c_{r}} \mathbf{r}_{r}^{H}, \mathbf{e}_{t}^{H}\right)^{2}+b_{h}+b_{t}$	\\
\hline								
\multirow{7}{30pt}{Gaussian}	&	\multirow{4}{30pt}{KG2E~\cite{he2015learning}}	&	$\mathbf{h} \sim \mathcal{N}\left(\boldsymbol{\mu}_{h}, \mathbf{\Sigma}_{h}\right)$	&	\multirow{2}{60pt}{$\mathbf{r} \sim \mathcal{N}\left(\boldsymbol{\mu}_{r}, \Sigma_{r}\right)$}	&	\multirow{2}{60pt}{$\int_{x \in \mathcal{R}^{k_{e}}} \mathcal{N}\left(x ; \boldsymbol{\mu}_{r}, \boldsymbol{\Sigma}_{r}\right) \log \frac{\mathcal{N}\left(x ; \boldsymbol{\mu}_{e}, \boldsymbol{\Sigma}_{e}\right)}{\mathcal{N}\left(x ; \boldsymbol{\mu}_{r}, \boldsymbol{\Sigma}_{r}\right)} d x$}	\\
	&		&	$\mathbf{t} \sim \mathcal{N}\left(\boldsymbol{\mu}_{t}, \Sigma_{t}\right)$	&		&		\\
	&		&	$\boldsymbol{\mu}_{h}, \boldsymbol{\mu}_{t} \in \mathbb{R}^{d}$	&	\multirow{2}{80pt}{$\boldsymbol{\mu}_{r} \in \mathbb{R}^{d}, \Sigma_{r} \in \mathbb{R}^{d \times d}$}	&	\multirow{2}{60pt}{$\log \int_{x \in \mathcal{R}^{k_{e}}} \mathcal{N}\left(x ; \boldsymbol{\mu}_{e}, \boldsymbol{\Sigma}_{e}\right) \mathcal{N}\left(x ; \boldsymbol{\mu}_{r}, \boldsymbol{\Sigma}_{r}\right) d x$}	\\
	&		&	$\Sigma_{h}, \Sigma_{t} \in \mathbb{R}^{d \times d}$	&		&		\\
	\cdashline{2-5}
	&	\multirow{3}{30pt}{TransG~\cite{xiao2016transg}}	&	$\mathbf{h} \sim \mathcal{N}\left(\boldsymbol{\mu}_{h}, \boldsymbol{\sigma}_{h}^{2} \mathbf{I}\right)$	&	$\boldsymbol{\mu}_{r}^{i} \sim \mathcal{N}\left(\boldsymbol{\mu}_{t}-\boldsymbol{\mu}_{h},\left(\sigma_{h}^{2}+\sigma_{t}^{2}\right) \mathbf{I}\right)$	&	\multirow{3}{60pt}{$\sum_{i} \pi_{r}^{i} \exp \left(-\frac{\left\|\boldsymbol{\mu}_{h}+\boldsymbol{\mu}_{r}^{i}-\boldsymbol{\mu}_{t}\right\|_{2}^{2}}{\sigma_{h}^{2}+\sigma_{t}^{2}}\right)$}	\\
	&		&	$\mathbf{t} \sim \mathcal{N}\left(\boldsymbol{\mu}_{t}, \Sigma_{t}\right)$	&	\multirow{2}{80pt}{$\mathbf{r}=\sum_{i} \pi_{r}^{i} \boldsymbol{\mu}_{r}^{i} \in \mathbb{R}^{d}$}	&		\\
	&		&	$\boldsymbol{\mu}_{h}, \boldsymbol{\mu}_{t} \in \mathbb{R}^{d}$	&		&		\\
\hline									
\multirow{11}{40pt}{Translational Distance}	&	TransE~\cite{bordes2013translating}	&	$\mathbf{h}, \mathbf{t} \in \mathbb{R}^{d}$	&	$\mathbf{r} \in \mathbb{R}^{d}$	&	$-\|\mathbf{h}+\mathbf{r}-\mathbf{t}\|_{1 / 2}$	\\
	&	TransR~\cite{lin2015learning}	&	$\mathbf{h}, \mathbf{t} \in \mathbb{R}^{d}$	&	$\mathbf{r} \in \mathbb{R}^{k}, \mathbf{M}_{r} \in \mathbb{R}^{k \times d}$	&	$-\left\|\mathbf{M}_{r} \mathbf{h}+\mathbf{r}-\mathbf{M}_{r} \mathbf{t}\right\|_{2}^{2}$	\\
	&	TransH~\cite{wang2014knowledge}	&	$\mathbf{h}, \mathbf{t} \in \mathbb{R}^{d}$	&	$\mathbf{r}, \mathbf{w}_{r} \in \mathbb{R}^{d}$	&	$-\left\|\left(\mathbf{h}-\mathbf{w}_{r}^{\top} \mathbf{h} \mathbf{w}_{r}\right)+\mathbf{r}-\left(\mathbf{t}-\mathbf{w}_{r}^{\top} \mathbf{t} \mathbf{w}_{r}\right)\right\|_{2}^{2}$	\\
	&	TransA~\cite{xiao2015transa}	&	$\mathbf{h}, \mathbf{t} \in \mathbb{R}^{d}$	&	$\mathbf{r} \in \mathbb{R}^{d}, \mathbf{M}_{r} \in \mathbb{R}^{d \times d}$	&	$(|\mathbf{h}+\mathbf{r}-\mathbf{t}|)^{\top} \mathbf{W}_{\mathbf{r}}(|\mathbf{h}+\mathbf{r}-\mathbf{t}|)$	\\
	&	TransF~\cite{feng2016knowledge} 	&	$\mathbf{h}, \mathbf{t} \in \mathbb{R}^{d}$	&	$\mathbf{r} \in \mathbb{R}^{d}$	&	$(\mathbf{h}+\mathbf{r})^{\top} \mathbf{t}+(\mathbf{t}-\mathbf{r})^{\top} \mathbf{h}$	\\
	&	ITransF~\cite{xie2017interpretable}	&	 $\mathbf{h}, \mathbf{t} \in \mathbb{R}^d$	&	$\mathbf{r}\in\mathbb{R}^d$	&	$\left\|\boldsymbol{\alpha}_{r}^{H} \cdot \mathbf{D} \cdot \mathbf{h}+\mathbf{r}-\boldsymbol{\alpha}_{r}^{T} \cdot \mathbf{D} \cdot \mathbf{t}\right\|_{\ell}$ 	\\
	&	TransAt~\cite{qian2018translating}	&	$\mathbf{h}, \mathbf{t} \in \mathbb{R}^{d}$	&	$\mathbf{r} \in \mathbb{R}^{d}$	&	$P_{r}\left(\sigma\left(\mathbf{r}_{h}\right) \mathbf{h}\right)+\mathbf{r}-P_{r}\left(\sigma\left(\mathbf{r}_{t}\right) \mathbf{t}\right)$	\\
	&	TransD~\cite{ji2015knowledge}	&	$\mathbf{h}, \mathbf{t}, \mathbf{w}_{h} \mathbf{w}_{t} \in \mathbb{R}^{d}$	&	$\mathbf{r}, \mathbf{w}_{r} \in \mathbb{R}^{k}$	&	$-\left\|\left(\mathbf{w}_{r} \mathbf{w}_{h}^{\top}+\mathbf{I}\right) \mathbf{h}+\mathbf{r}-\left(\mathbf{w}_{r} \mathbf{w}_{t}^{\top}+\mathbf{I}\right) \mathbf{t}\right\|_{2}^{2}$	\\
	&	TransM~\cite{fan2014transition}	&	$\mathbf{h}, \mathbf{t} \in \mathbb{R}^{d}$	&	$\mathbf{r} \in \mathbb{R}^{d}$	&	$-\theta_{r}\|\mathbf{h}+\mathbf{r}-\mathbf{t}\|_{1 / 2}$	\\
	&	\multirow{2}{30pt}{TranSparse~\cite{ji2016knowledge}}	&	\multirow{2}{50pt}{$\mathbf{h}, \mathbf{t} \in \mathbb{R}^{d}$}	&	$\mathbf{r} \in \mathbb{R}^{k}, \mathbf{M}_{r}\left(\theta_{r}\right) \in \mathbb{R}^{k \times d}$	&	$-\left\|\mathbf{M}_{r}\left(\theta_{r}\right) \mathbf{h}+\mathbf{r}-\mathbf{M}_{r}\left(\theta_{r}\right) \mathbf{t}\right\|_{1 / 2}^{2}$	\\
	&		&		&	$\mathbf{M}_{r}^{1}\left(\theta_{r}^{1}\right), \mathbf{M}_{r}^{2}\left(\theta_{r}^{2}\right) \in \mathbb{R}^{k \times d}$	&	$-\left\|\mathbf{M}_{r}^{1}\left(\theta_{r}^{1}\right) \mathbf{h}+\mathbf{r}-\mathbf{M}_{r}^{2}\left(\theta_{r}^{2}\right) \mathbf{t}\right\|_{1 / 2}^{2}$	\\
\hline									
\multirow{13}{30pt}{Semantic Matching}	&	TATEC~\cite{garcia2014effective}	&	$\mathbf{h}, \mathbf{t} \in \mathbb{R}^{d}$	&	$\mathbf{r} \in \mathbb{R}^{d}, \mathbf{M}_{r} \in \mathbb{R}^{d \times d}$	&	$\mathbf{h}^{\top} \mathbf{M}_{r} \mathbf{t}+\mathbf{h}^{\top} \mathbf{r}+\mathbf{t}^{\top} \mathbf{r}+\mathbf{h}^{\top} \mathbf{D} \mathbf{t}$	\\
	&	ANALOGY~\cite{liu2017analogical}	&	$\mathbf{h}, \mathbf{t} \in \mathbb{R}^{d}$	&	$\mathbf{M}_{r} \in \mathbb{R}^{d \times d}$	&	$ \mathbf{h}^{\top} \mathbf{M}_{r} \mathbf{t}$	\\
	&	CrossE~\cite{zhang2019interaction}	&	 $\mathbf{h}, \mathbf{t} \in \mathbb{R}^d$	&	$\mathbf{r}\in\mathbb{R}^d$	&	$\sigma\left(\tanh \left(\mathbf{c}_{r} \circ \mathbf{h}+\mathbf{c}_{r} \circ \mathbf{h} \circ \mathbf{r}+\mathbf{b}\right) \mathbf{t}^{\top}\right)$	\\
  &	SME~\cite{bordes2014semantic}	&	$\mathbf{h}, \mathbf{t} \in \mathbb{R}^{d}$	&	$\mathbf{r} \in \mathbb{R}^{d}$	&	$g_\text{left}(\mathbf{h}, \mathbf{r})^\top g_\text{right}(\mathbf{r}, \mathbf{t})$	\\
	&	DistMult~\cite{yang2014embedding} 	&	$\mathbf{h}, \mathbf{t} \in \mathbb{R}^{d}$	&	$\mathbf{r} \in \mathbb{R}^{d}$	&	$\mathbf{h}^{\top} \operatorname{diag}(\mathbf{M}_{r}) \mathbf{t}$	\\
	&	HolE~\cite{nickel2016holographic}	&	$\mathbf{h}, \mathbf{t} \in \mathbb{R}^{d}$	&	$\mathbf{r} \in \mathbb{R}^{d}$	&	$\mathbf{r}^\top (h\star t)$	\\
	&	HolEx~\cite{xue2018expanding}	&	$\mathbf{h}, \mathbf{t} \in \mathbb{R}^d$ 	&	$\mathbf{r}\in\mathbb{R}^d$	&	$\sum_{j=0}^{l} p\left(\mathbf{h}, \boldsymbol{r} ; \boldsymbol{c}_{j}\right) \cdot \boldsymbol{t}$	\\
	&	SE~\cite{bordes2011learning}	&	$\mathbf{h}, \mathbf{t} \in \mathbb{R}^{d}$	&	$\mathbf{M}_{r}^{1}, \mathbf{M}_{r}^{2} \in \mathbb{R}^{d \times d}$	&	$-\left\|\mathbf{M}_{r}^{1} \mathbf{h}-\mathbf{M}_{r}^{2} \mathbf{t}\right\|_{1}$	\\
	&	SimplE~\cite{kazemi2018simple}	&	$\mathbf{h}, \mathbf{t} \in \mathbb{R}^d$	&	$\mathbf{r}, \mathbf{r}^\prime\in\mathbb{R}^d$	&	$\frac{1}{2}\left( \mathbf{h} \circ \mathbf{r} \mathbf{t}+ \mathbf{t}\circ \mathbf{r}^{\prime} \mathbf{t}\right)$	\\
  &	RESCAL~\cite{nickel2011three}	&	$\mathbf{h}, \mathbf{t} \in \mathbb{R}^{d}$	&	$\mathbf{M}_{r} \in \mathbb{R}^{d \times d}$	&	$\mathbf{h}^{\top} \mathbf{M}_{r} \mathbf{t}$	\\
	&	LFM~\cite{jenatton2012latent}	&	$\mathbf{h}, \mathbf{t} \in \mathbb{R}^{d}$	&	$\mathbf{u}_{r}, \mathbf{v}_{r} \in \mathbb{R}^{p}$	&	$\mathbf{h}^{\top} \sum_{i=1}^d \boldsymbol{\alpha}_i^r \mathbf{u}_i \mathbf{v}_i^\top \mathbf{t}$	\\
	&	TuckER~\cite{balavzevic2019tucker}	&	$\mathbf{h}, \mathbf{t} \in \mathbb{R}^d_e$	&	$\mathbf{r}\in\mathbb{R}^d_r$	&	$\mathcal{W} \times_{1} \mathbf{h} \times_{2} \mathbf{r} \times_{3} \mathbf{t}$	\\
	&	LowFER~\cite{amin2020lowfer}	&	$\mathbf{h}, \mathbf{t} \in \mathbb{R}^{d}$	&	$\mathbf{r} \in \mathbb{R}^{d}$	&	$\left(\mathbf{S}^{k} \operatorname{diag}\left(\mathbf{U}^{T} \mathbf{h}\right) \mathbf{V}^{T} \mathbf{r}\right)^{T} \mathbf{t}$	\\
\hline									
\multirow{7}{30pt}{Neural Networks} 	&	MLP~\cite{dong2014knowledge}	&	$\mathbf{h}, \mathbf{t} \in \mathbb{R}^{d}$	&	$\mathbf{r} \in \mathbb{R}^{d}$	&	$\operatorname{\sigma}(\mathbf{w}^\top \operatorname{\sigma}(\mathbf{W}[\mathbf{h}, \mathbf{r}, \mathbf{t}]))$	\\
	&	NAM~\cite{liu2016probabilistic}	&	$\mathbf{h}, \mathbf{t} \in \mathbb{R}^{d}$	&	$\mathbf{r} \in \mathbb{R}^{d}$	&	$\sigma\left(\mathbf{z}^{(L)} \cdot \mathbf{t}+\mathbf{B}^{(L+1)} \mathbf{r}\right)$	\\
  &	ConvE~\cite{dettmers2018convolutional}	&	$\mathbf{M}_h \in \mathbb{R}^{d_w\times d_h}, \mathbf{t} \in \mathbb{R}^d$	&	$\mathbf{M}_r \in\mathbb{R}^{d_w\times d_h}$	&	$\sigma\left(\operatorname{vec}\left(\sigma\left(\left[\mathbf{M}_h ; \mathbf{M}_r\right] * \boldsymbol{\omega}\right)\right) \mathbf{W}\right) \mathbf{t}$	\\
	&	ConvKB~\cite{nguyen2017novel}	&	$\mathbf{h}, \mathbf{t} \in \mathbb{R}^d$	&	$\mathbf{r}\in\mathbb{R}^d$	&	$\operatorname{concat}\left(\sigma\left(\left[\boldsymbol{h}, \boldsymbol{r}, \boldsymbol{t}\right] * \boldsymbol{\omega}\right)\right) \cdot \mathbf{w}$	\\
	&	HypER~\cite{balavzevic2019hypernetwork}	&	$\mathbf{h}, \mathbf{t} \in \mathbb{R}^d$	&	$\mathbf{w}_{r} \in \mathbb{R}^{d_r}$	&	$\sigma\left(\operatorname{vec}\left(\mathbf{h} * \operatorname{vec}^{-1}\left(\mathbf{w}_{r} \mathbf{H}\right)\right) \mathbf{W}\right) \mathbf{t}$	\\
	&	SACN~\cite{shang2018end}	&	$\mathbf{h}, \mathbf{t} \in \mathbb{R}^d$	&	$\mathbf{r}\in\mathbb{R}^d$	&	$g\left(\operatorname{vec}\left(\mathbf{M}\left(\mathbf{h}, \mathbf{r}\right)\right) W\right) \mathbf{t}$	\\
	&	\multirow{2}{30pt}{NTN~\cite{socher2013reasoning}}	&	\multirow{2}{50pt}{$\mathbf{h}, \mathbf{t} \in \mathbb{R}^{d}$}	&	$\mathbf{r},\mathbf{b}_r\in \mathbb{R}^k,\mathbf{\widehat{M}}\in \mathbb{R}^{d\times d \times k}$	&	\multirow{2}{100pt}{$\mathbf{r}^\top \sigma \left(\mathbf{h}^T\mathbf{\widehat{M}}\mathbf{t} + \mathbf{M}_{r,1}\mathbf{h} + \mathbf{M}_{r,2}\mathbf{t} + \mathbf{b}_r\right)$}	\\
	&		&		&	$\mathbf{M}_{r,1},\mathbf{M}_{r,2}\in \mathbb{R}^{k\times d}$	&		\\
\bottomrule 
\end{tabular}
\renewcommand{\arraystretch}{1}
\end{center}
\label{tab:more-KRL}
\end{table*}

We conduct a comprehensive summary of KRL models in Table~\ref{tab:more-KRL}.
The representation space has an impact on the expressiveness of KRL methods to some extent. By expanding point-wise Euclidean space~\cite{bordes2013translating, socher2013reasoning, nickel2016holographic}, manifold space~\cite{xiao2016one}, complex space~\cite{trouillon2016complex, sun2018rotate, zhang2019quaternion} and Gaussian distribution~\cite{he2015learning, xiao2016transg} are introduced.

Distance-based and semantic matching scoring functions consist of the foundation stones of plausibility measure in KRL. Translational distance-based methods, especially the groundbreaking TransE~\cite{bordes2013translating}, borrowed the idea of distributed word representation learning and inspired many following approaches, such as TransH~\cite{wang2014knowledge} and TransR~\cite{lin2015learning} which specify complex relations (1-to-N, N-to-1, and N-to-N) and the recent TransMS~\cite{yang2019transms} which models multi-directional semantics.
As for the semantic matching side, many methods utilizes mathematical operations or compositional operators including linear matching in SME~\cite{bordes2014semantic}, bilinear mapping in DistMult~\cite{yang2014embedding}, tensor product in NTN~\cite{socher2013reasoning}, circular correlation in HolE~\cite{nickel2016holographic} and ANALOGY~\cite{liu2017analogical}, Hadamard product in CrossE~\cite{zhang2019interaction}, and quaternion inner product in QuatE~\cite{zhang2019quaternion}.

Recent encoding models for knowledge representation have developed rapidly and generally fall into two families of bilinear and neural networks.
Linear and bilinear models use product-based functions over entities and relations, while factorization models regard knowledge graphs as three-way tensors.
With the multiplicative operations, RESCAL~\cite{nickel2011three}, ComplEx~\cite{trouillon2016complex}, and SimplE~\cite{kazemi2018simple} also belong to the bilinear models. DistMult~\cite{yang2014embedding} can only model symmetric relations, while its extension of ComplEx~\cite{trouillon2016complex} managed to preserve antisymmetric relations, but involves redundant computations~\cite{kazemi2018simple}. ComplEx~\cite{trouillon2016complex}, SimplE~\cite{kazemi2018simple}, and TuckER~\cite{balavzevic2019tucker} can guarantee full expressiveness under specific embedding dimensionality bounds. 
Neural network-based encoding models start from distributed representation of entities and relations, and some utilizes complex neural structures such as tensor networks~\cite{socher2013reasoning}, graph convolution networks~\cite{schlichtkrull2018modeling,shang2018end,nathani2019learning}, recurrent networks~\cite{guo2019learning} and transformers~\cite{wang2019coke, yao2019kgbert} to learn richer representation.  
These deep models have achieved very competitive results, but they are not transparent, and lack of interpretability. 
As deep learning techniques are growing prosperity and gaining extensive superiority in many tasks, the recent trend is still likely to focus on more powerful neural architectures or large-scale pre-training, while deep interpretable models remains a challenge. 

\section{KRL Model Training}\label{sup:KRL-training}
Open world assumption (OWA) and closed world assumption (CWA)~\cite{reiter1978deductive} are considered when training knowledge representation learning models.
During training, a negative sample set $\mathcal{F}^\prime$ is randomly generated by corrupting a golden triple set $\mathcal{F}$ under the OWA. 
Mini-batch optimization and Stochastic Gradient Descent (SGD) are carried out to minimize a certain loss function.
Under the OWA, negative samples are generated with specific sampling strategies designed to reduce the number of false negatives.

\subsection{Open and Closed World Assumption} 
The CWA assumes that unobserved facts are false. By contrast, the OWA has a relaxed assumption that unobserved ones can be either missing or false. 
Generally, OWA has an advantage over CWA because of the incompleteness nature of knowledge graphs. 
RESCAL \cite{nickel2011three} is a typical model trained under the CWA, while more models are formulated under the OWA.

\subsection{Loss Function}
Several families of loss function are introduced for KRL model optimization. 
First, a margin-based loss is optimized to learn representations that positive samples have higher scores than negative ones. 
Some literature also called it as pairwise ranking loss. As shown in Eq.~\ref{eq:margin} , the rank-based hinge loss maximizes the discriminative margin between a golden triple $(h, r, t)$ and an invalid triple $(h^{\prime}, r, t^{\prime})$.
\begin{equation}\small
\label{eq:margin}
\min _{\Theta} \sum_{(h, r, t) \in \mathcal{F}} \sum_{\left(h^{\prime}, r, t^{\prime}\right) \in \mathcal{F}^{\prime}} \max \left(0, f_{r}(h, t)+\gamma-f_{r}\left(h^{\prime}, t^{\prime}\right)\right)
\end{equation}
here $\gamma$ is a margin. The invalid triple $(h^{\prime}, r, t^{\prime})$ is constructed by randomly changing a head or tail entity or both entities in the knowledge graph. Most translation-based embedding methods use margin-based loss~\cite{cai2018kbgan}. 
The second kind of loss function is logistic-based loss in Eq.~\ref{eq:logistic}, which is to minimize negative log-likelihood of logistic models.
\begin{equation}\small
\label{eq:logistic}
\min _{\Theta} \sum_{(h, r, t) \in \mathcal{F} \cup \mathcal{F}^{\prime}} \log \left(1+\exp \left(-y_{h r t} \cdot f_{r}(h, t)\right)\right)
\end{equation}
here $y_{h r t}$ is the label of triple instance.
Some methods also use other kinds of loss functions. For example, ConvE and TuckER use binary cross-entropy or the so-called Bernoulli negative log-likelihood loss function defined as:
\begin{equation}\small
\label{log}
-\frac{1}{N_e} \sum_{i}^{N_e}\left(y_{i} \cdot \log \left(p_{i}\right)+\left(1-y_{i}\right) \cdot \log \left(1-p_{i}\right)\right),
\end{equation}
where $p$ is the prediction and $y$ is the ground label. 
And RotatE uses the form of loss function in Eq.~\ref{eq:loss-rotate}.
\begin{equation}\small
\label{eq:loss-rotate}
-\log \sigma\left(\gamma-f_{r}({h}, {t})\right)-\sum_{i=1}^{n} \frac{1}{k} \log \sigma\left(f_{r}\left({h}_{i}^{\prime}, {t}_{i}^{\prime}\right)-\gamma\right)
\end{equation}
For all those kinds of loss functions, specific regularization like L2 on parameters or constraints can also be applied, as well as combined with the joint learning paradigm. 

\subsection{Negative~Sampling}
Several heuristics of sampling distribution are proposed to corrupt the head or tail entities. The widest applied one is uniform sampling~\cite{bordes2013translating, bordes2014semantic, lin2015learning} that uniformly replaces entities.
But it leads to the sampling of false-negative labels.
More effective negative sampling strategies are required to learn semantic representation and improve predictive performance. 

Considering the mapping property of relations, Bernoulli sampling~\cite{wang2014knowledge} introduces a heuristic of a sampling distribution as
$\frac{t p h}{t p h+h p t}$, where $tph$ and $hpt$ denote the average number of tail entities per head entity and the average number of head entities per tail entity respectively. 
Domain sampling~\cite{xie2017interpretable} chooses corrupted samples from entities in the same domain or from the whole entity set with a relation-dependent probability $p_r$ or $1-p_r$ respectively, with the head and tail domain of relation $r$ denoted as $\mathrm{M}_{r}^{H}=\{h~|~\exists~t(h, r, t) \in P\}$ and $\mathrm{M}_{r}^{T}=\{t~|~\exists~h(h, r, t) \in P\}$, and induced relational set denoted as $\mathrm{N}_{r}=\{(h, r, t) \in P\}$.

Recently, two adversarial sampling methods are further proposed. 
KBGAN~\cite{cai2018kbgan} introduces adversarial learning for negative sampling, where the generator uses probability-based log-loss embedding models. The probability of generating negative samples $p\left(h_{j}^{\prime}, r, t_{j}^{\prime} |\left\{\left(h_{i}, r_{i}, t_{i}\right)\right\}\right)$ is defined as 
\begin{equation}\small
\frac{\exp f_{G}\left(h_{i}^{\prime}, r, t_{i}^{\prime}\right)}{\sum_{j=1} \exp f_{G}\left(h_{j}^{\prime}, r, t_{j}^{\prime}\right)},
\end{equation}
where $f_G(h, r, t)$ is the scoring function of generator.
Similarly, Sun et al.~\cite{sun2018rotate} proposed self-adversarial negative sampling based on self scoring function by sampling negative triples from the distribution in Eq.~\ref{eq:self-adversarial}, where $\alpha$ is the temperature of sampling. 
\begin{equation}\small
\label{eq:self-adversarial}
p\left(h_{j}^{\prime}, r, t_{j}^{\prime} |\left\{\left(h_{i}, r_{i}, t_{i}\right)\right\}\right)=\frac{\exp \alpha f\left(h_{j}^{\prime}, r, t_{j}^{\prime}\right)}{\sum_{i} \exp \alpha f\left(h_{i}^{\prime},r, t_{i}^{\prime}\right)}
\end{equation}
Negative sampling strategies are summarized in Table~\ref{tab:sampling}.
Trouillon et al.~\cite{trouillon2016complex} studied the number of negative samples generated per positive training sample and found a trade-off between accuracy and training time.

\begin{table}[htp]
\scriptsize
\caption{A summary of negative sampling}
\begin{center}
\begin{tabular}{ l l l }
\toprule
Sampling & Mechanism & Sampling probability \\
\midrule
Uniform~\cite{bordes2014semantic} & uniform distribution & $\frac{1}{n}$\\
Bernoulli~\cite{wang2014knowledge} & mapping property &$\frac{t p h}{t p h+h p t}$ \\
Domain~\cite{xie2017interpretable} & relation-depend domain & $\min \left(\frac{\lambda\left|\mathrm{M}_{r}^{T}\right|\left|\mathrm{M}_{r}^{H}\right|}{\left|N_{r}\right|}, 0.5\right)$\\
Adversarial~\cite{cai2018kbgan} & generator embedding & $\frac{\exp f_{G}\left(h_{i}^{\prime}, r, t_{i}^{\prime}\right)}{\sum_{j=1} \exp f_{G}\left(h_{j}^{\prime}, r, t_{j}^{\prime}\right)}$ \\
Self-adversarial~\cite{sun2018rotate} & current embedding & $\frac{\exp \alpha f\left(h_{j}^{\prime}, r, t_{j}^{\prime}\right)}{\sum_{i} \exp \alpha f\left(h_{i}^{\prime},r, t_{i}^{\prime}\right)}$ \\
\bottomrule
\end{tabular}
\end{center}
\label{tab:sampling}
\end{table}

\section{More Knowledge-aware Applications}
\label{sup:more-app}

\subsection{Text Classification and Task-Specific Applications}
Knowledge-aware NLU enhances language representation with structured knowledge injected into a unified semantic space.
Recent knowledge-driven advances utilize explicit factual knowledge and implicit language representation.
Wang et al.~\cite{wang2017combining} augmented short text representation learning with knowledge-based conceptualization by a weighted word-concept embedding. 
Peng et al.~\cite{peng2019fine} integrated an external knowledge base to build a heterogeneous information graph for event categorization in short social text.
In the mental healthcare domain, models with knowledge graph facilitate a good understanding of mental conditions and risk factors of mental disorders and are applied to effective prevention of mental health leaded suicide. 
Gaurs et al.~\cite{gaur2019knowledge} developed a rule-based classifier for knowledge-aware suicide risk assessment with a suicide risk severity lexicon incorporating medical knowledge bases and suicide ontology. 

Sentiment analysis integrated with sentiment-related concepts can better understand people's opinions and sentiments. SenticNet~\cite{cambria2018senticnet} learns conceptual primitives for sentiment analysis, which can also be used as a commonsense knowledge source. To enable sentiment-related information filtering, Sentic LSTM~\cite{ma2018targeted} injects knowledge concepts to the vanilla LSTM and designs a knowledge output gate for concept-level output as a complement to the token level.

\subsection{Dialogue~Systems}
QA can also be viewed as a single-turn dialogue system by generating the correct answer as a response, while dialogue systems consider conversational sequences and aim to generate fluent responses to enable multi-round conversations via semantic augmentation and knowledge graph walk. 
Liu et al.~\cite{liu2019knowledge} encoded knowledge to augment semantic representation and generated knowledge aware response by knowledge graph retrieval and graph attention mechanism under an encoder-decoder framework.
DialKG Walker~\cite{moon2019opendialkg} traverses a symbolic knowledge graph to learn contextual transition in dialogue and predicts entity responses with attentive graph path decoder.

Semantic parsing via formal logical representation is another direction for dialog systems. 
By predefining a set of base actions, Dialog-to-Action~\cite{guo2018dialog} is an encoder-decoder approach that maps executable logical forms from the utterance in conversation, to generate action sequence under the control of a grammar-guided decoder.

\subsection{Medicine and Biology}
Knowledge-aware models and their applications pave the way to incorporate domain knowledge for precise prediction in medicine and biology domains. 
Medical~applications involve a domain-specific knowledge graph of medical concepts.
Sousa et al. \cite{sousa2020evolving} adopted knowledge graph similarity for protein-protein interaction prediction using the Gene Ontology.
Mohamed et al. \cite{mohamed2020discovering} formulated drug-target interaction prediction as a link prediction in biomedical knowledge graphs with drugs and their potential targets. 
Lin et al. \cite{lin2020kgnn} developed a knowledge graph network to learn structural information and semantic relation for drug-drug interaction prediction.
In the clinical domain, biomedical knowledge from the Unified Medical Language Systems (UMLS) ontology is integrated into language model pretraining for downstream clinical applications such as clinical entity recognition and medical language inference~\cite{hao2020enhancing}. 
Li et al.~\cite{li2019knowledge} formulated the task of medical image report generation with three steps of encoding, retrieval, and paraphrasing, where the medical image is encoded by the abnormality graph.

\subsection{Other Applications}
There are also many other applications that utilize knowledge-driven methods. 
\textit{1) Academic~search~engine} helps research to find relevant academic papers.  
Xiong et al.~\cite{xiong2017explicit} proposed explicit semantic ranking with knowledge graph embedding to help academic search better understand the meaning of query concepts. 
\textit{2) Zero-shot image classification} gets benefits from knowledge graph propagation with semantic descriptions of classes. Wang et al.~\cite{wang2018zero} proposed a multi-layer GCN to learn zero-shot classifiers using semantic embeddings of categories and categorical relationships. APNet~\cite{liu2020attribute} propagates attribute representations with a category graph. 
\textit{3) Text generation} synthesizes and composes coherent multi-sentence texts. Koncel-Kedziorski et al.~\cite{koncel2019text} studied text generation for information extraction systems and proposed a graph transforming encoder for graph-to-text generation from the knowledge graph. 
Question generation focuses on generating natural language questions. 
Seyler et al.~\cite{seyler2017knowledge} studied quiz-style knowledge question generation by generating a structured triple-pattern query over the knowledge graph while estimating how difficult the questions are. 
However, for verbalizing the question, the authors used a template-based method, which may have a limitation on generating more natural expressions.

\section{Datasets~and~Libraries}
\label{sec:practice}
In this section, we introduce and list useful resources of knowledge graph datasets and open-source libraries. 

\subsection{Datasets}
\label{sup:datasets}
Many public datasets have been released. We conduct an introduction and a summary of general, domain-specific, task-specific, and temporal datasets.

\subsubsection{General~Datasets}
\label{app:general-datasets}
Datasets with general ontological knowledge include WordNet~\cite{miller1995wordnet}, Cyc~\cite{matuszek2006introduction}, DBpedia~\cite{auer2007dbpedia}, YAGO~\cite{suchanek2007yago}, Freebase~\cite{bollacker2008freebase}, NELL~\cite{carlson2010toward} and Wikidata~\cite{vrandevcic2014wikidata}.
It is hard to compare them within a table as their ontologies are different. Thus, only an informal comparison is illustrated in Table~\ref{tab:general}, where their volumes kept going after their release. 

WordNet, firstly released in 1995, is a lexical database that contains about 117,000 synsets. 
DBpedia is a community-driven dataset extracted from Wikipedia. It contains 103 million triples and can be enlarged when interlinked with other open datasets.
To solve the problems of low coverage and low quality of single-source ontological knowledge, YAGO utilized the concept information in the category page of Wikipedia and the hierarchy information of concepts in WordNet to build a multi-source dataset with high coverage and quality. Moreover, it is extendable by other knowledge sources. It is available online with more than 10 million entities and 120 million facts currently. 
Freebase, a scalable knowledge base, came up for the storage of the world's knowledge in 2008.
Its current number of triples is 1.9 billion. 
NELL is built from the Web via an intelligent agent called Never-Ending Language Learner. It has 2,810,379 beliefs with high confidence by far. 
Wikidata is a free structured knowledge base, which is created and maintained by human editors to facilitate the management of Wikipedia data. It is multi-lingual with 358 different languages.

The aforementioned datasets are openly published and maintained by communities or research institutions. There are also some commercial datasets. The Cyc knowledge base from Cycorp contains about 1.5 million general concepts and more than 20 million general rules, with an accessible version called OpenCyc deprecated sine 2017.
Google knowledge graph hosts more than 500 million entities and 3.5 billion facts and relations.
Microsoft builds a probabilistic taxonomy called Probase~\cite{wu2012probase} with 2.7 million concepts.

\begin{table*}[htp]
\scriptsize
\caption{Statistics of datasets with general knowledge when originally released}
\begin{center}
\begin{tabular}{llll}
\toprule 
Dataset & \# entities & \# facts & Website\\
\midrule
WordNet~\cite{miller1995wordnet} & 117,597 & 207,016 & \url{https://wordnet.princeton.edu}\\
OpenCyc~\cite{matuszek2006introduction} & 47,000 & 306,000 & \url{https://www.cyc.com/opencyc/} \\
Cyc~\cite{matuszek2006introduction} & $\sim$250,000 & $\sim$2,200,000 & \url{https://www.cyc.com} \\
YAGO~\cite{suchanek2007yago} & 1,056,638 & $\sim$5,000,000 & \url{http://www.mpii.mpg.de/~suchanek/yago} \\
DBpedia~\cite{auer2007dbpedia} & $\sim$1,950,000 & $\sim$103,000,000& \url{https://wiki.dbpedia.org/develop/datasets}\\
Freebase~\cite{bollacker2008freebase} & - & $\sim$125,000,000 & \url{https://developers.google.com/freebase/} \\
NELL~\cite{carlson2010toward} & - & 242,453 & \url{http://rtw.ml.cmu.edu/rtw/}\\
Wikidata~\cite{vrandevcic2014wikidata}& 14,449,300 & 30,263,656 & \url{https://www.wikidata.org/wiki}\\
Probase IsA & 12,501,527 & 85,101,174 &\url{https://concept.research.microsoft.com/Home/Download} \\
Google KG & $>$ 500 million & $>$ 3.5 billion & \url{https://developers.google.com/knowledge-graph} \\
\bottomrule
\end{tabular}
\end{center}
\label{tab:general}
\end{table*}%

\subsubsection{Domain-Specific~Datasets}
\label{app:domain-specific}
Some knowledge bases on specific domains are designed and collected to evaluate domain-specific tasks. Some notable domains include life science, health care, and scientific research, covering complex domains and relations such as compounds, diseases, and tissues. 
Examples of domain-specific knowledge graphs are
ResearchSpace\footnote{\url{https://www.researchspace.org/index.html}}, a cultural heritage knowledge graph;
UMLS~\cite{mccray2003upper}, a unified medical language system; 
SNOMED CT\footnote{\url{http://www.snomed.org/snomed-ct/five-step-briefing}}, a commercial clinical terminology;
and a medical knowledge graph from Yidu Research\footnote{\url{https://www.yiducloud.com.cn/en/academy.html}}.
More biological databases with domain-specific knowledge include STRING, protein-protein interaction networks \footnote{\url{https://string-db.org}}; SKEMPI, a Structural Kinetic and Energetic database of Mutant Protein Interactions~\cite{moal2012skempi}; Protein Data Bank (PDB) database\footnote{E.g., RCSB PDB (https://www.rcsb.org), a member of the worldwide PDB}, containing biological molecular data~\cite{berman2000protein}; GeneOntology\footnote{\url{http://geneontology.org}}, a gene ontology resource that describes protein function; and DrugBank\footnote{\url{https://go.drugbank.com}}, a pharmaceutical knowledge base~\cite{wishart2006drugbank, wishart2008drugbank}.

\subsubsection{Task-Specific~Datasets}
A popular way of generating task-specific datasets is to sample subsets from large general datasets. Statistics of several datasets for tasks on the knowledge graph itself are listed in Table~\ref{tab:inKG}. Notice that WN18 and FB15k suffer from test set leakage~\cite{dettmers2018convolutional}. For KRL with auxiliary information and other downstream knowledge-aware applications, texts and images are also collected,
for example, WN18-IMG~\cite{xie2017image} with sampled images and textual relation extraction dataset including SemEval 2010 dataset, NYT~\cite{riedel2010modeling} and Google-RE\footnote{\url{https://code.google.com/archive/p/relation-extraction-corpus/}}. IsaCore~\cite{cambria2012semantic}, an analogical closure of Probase for opinion mining and sentiment analysis, is built by common knowledge base blending and multi-dimensional scaling. Recently, the FewRel dataset~\cite{han2018fewrel} was built to evaluate the emerging few-shot relation classification task. There are also more datasets for specific tasks such as cross-lingual DBP15K~\cite{sun2017cross} and DWY100K~\cite{sun2018bootstrapping} for entity alignment, multi-view knowledge graphs of YAGO26K-906 and DB111K-174~\cite{hao2019universal} with instances and ontologies. 

\begin{table}[htp]
\scriptsize
\caption{A summary of datasets for tasks on knowledge graph itself}
\begin{center}
\begin{tabular}{llllll}
\toprule
Dataset	&	\# Rel.	&	\#Ent.	&	\# Train	&	\# Valid.	&	\# Test	\\
\midrule
WN18~\cite{bordes2013translating}	&	18	&	40,943	&	141,442	&	5,000	&	5,000	\\
FB15K~\cite{bordes2013translating}	&	1,345	&	14,951	&	483,142	&	50,000	&	59,071	\\
WN11~\cite{socher2013reasoning}	&	11	&	38,696	&	112,581	&	2,609	&	10,544	\\
FB13~\cite{socher2013reasoning}	&	13	&	75,043	&	316,232	&	5,908	&	23,733	\\
WN18RR~\cite{dettmers2018convolutional}	&	11	&	40,943	&	86,835	&	3,034	&	3,134	\\
FB15k-237~\cite{toutanova2015observed}	&	237	&	14,541	&	272,115	&	17,535	&	20,466	\\
FB5M~\cite{wang2014knowledge}	&	1,192	&	5,385,322	&	19,193,556	&	50,000	&	59,071	\\
FB40K~\cite{lin2015learning}	&	1,336	&	39,528	&	370,648	&	67,946	&	96,678	\\
\bottomrule
\end{tabular}
\end{center}
\label{tab:inKG}
\end{table}%

Numerous downstream knowledge-aware applications also come up with many datasets, for example, 
WikiFacts~\cite{ahn2016neural} for language modeling; SimpleQuestions~\cite{bordes2015large} and LC-QuAD~\cite{trivedi2017lc} for question answering;
and Freebase Semantic Scholar~\cite{xiong2017explicit} for academic search.

\subsection{Open-Source~Libraries}
Recent research has boosted the open-source campaign, with several libraries listed in Table~\ref{tab:open}. They are AmpliGraph~\cite{ampligraph} for knowledge representation learning, Grakn for integration knowledge graph with machine learning techniques, and Akutan for knowledge graph store and query. The research community has also released codes to facilitate further research. Notably, there are three useful toolkits, namely scikit-kge and OpenKE~\cite{han2018openke} for knowledge graph embedding, and OpenNRE~\cite{han2019opennre} for relation extraction. We provide an online collection of knowledge graph publications, together with links to some open-source implementations of them, hosted at \url{https://shaoxiongji.github.io/knowledge-graphs/}.

\begin{table}[htp]
\setlength{\tabcolsep}{1pt}
\scriptsize
\caption{A summary of open-source libraries}
\begin{center}
\begin{tabular}{llll}
\toprule
Task & Library & Language & URL \\
\midrule
General & Grakn & Python & \href{https://github.com/graknlabs/kglib}{github.com/graknlabs/kglib} \\
General & AmpliGraph & TensorFlow & \href{https://github.com/Accenture/AmpliGraph}{github.com/Accenture/AmpliGraph} \\
General & GraphVite & Python & \href{https://graphvite.io/}{graphvite.io} \\
Database	&	Akutan	&	Go	&	\href{https://github.com/eBay/akutan}{github.com/eBay/akutan}	\\
KRL & OpenKE & PyTorch & \href{https://github.com/thunlp/OpenKE}{github.com/thunlp/OpenKE}\\
KRL & Fast-TransX & C++ & \href{https://github.com/thunlp/Fast-TransX}{github.com/thunlp/Fast-TransX} \\
KRL & scikit-kge & Python & \href{https://github.com/mnick/scikit-kge}{github.com/mnick/scikit-kge} \\
KRL & LibKGE & PyTorch & \href{https://github.com/uma-pi1/kge}{github.com/uma-pi1/kge} \\
KRL & PyKEEN & Python & \href{https://github.com/SmartDataAnalytics/PyKEEN}{github.com/SmartDataAnalytics/PyKEEN} \\
RE & OpenNRE & PyTorch & \href{https://github.com/thunlp/OpenNRE}{github.com/thunlp/OpenNRE}\\ 
\bottomrule
\end{tabular}
\end{center}
\label{tab:open}
\end{table}%

\ifCLASSOPTIONcaptionsoff
 \newpage
\fi

\balance
\bibliographystyle{IEEEtran}
\bibliography{ref-survey-kg}

\begin{thebibliography}{100}
\providecommand{\url}[1]{#1}
\csname url@samestyle\endcsname
\providecommand{\newblock}{\relax}
\providecommand{\bibinfo}[2]{#2}
\providecommand{\BIBentrySTDinterwordspacing}{\spaceskip=0pt\relax}
\providecommand{\BIBentryALTinterwordstretchfactor}{4}
\providecommand{\BIBentryALTinterwordspacing}{\spaceskip=\fontdimen2\font plus
\BIBentryALTinterwordstretchfactor\fontdimen3\font minus
  \fontdimen4\font\relax}
\providecommand{\BIBforeignlanguage}[2]{{%
\expandafter\ifx\csname l@#1\endcsname\relax
\typeout{** WARNING: IEEEtran.bst: No hyphenation pattern has been}%
\typeout{** loaded for the language `#1'. Using the pattern for}%
\typeout{** the default language instead.}%
\else
\language=\csname l@#1\endcsname
\fi
#2}}
\providecommand{\BIBdecl}{\relax}
\BIBdecl

\bibitem{newell1959report}
A.~Newell, J.~C. Shaw, and H.~A. Simon, ``Report on a general problem solving
  program,'' in \emph{IFIP congress}, vol. 256, 1959, p.~64.

\bibitem{shortliffe2012computer}
E.~Shortliffe, \emph{Computer-based medical consultations: {MYCIN}}.\hskip 1em
  plus 0.5em minus 0.4em\relax Elsevier, 2012, vol.~2.

\bibitem{dong2014knowledge}
X.~Dong, E.~Gabrilovich, G.~Heitz, W.~Horn, N.~Lao, K.~Murphy, T.~Strohmann,
  S.~Sun, and W.~Zhang, ``Knowledge vault: A web-scale approach to
  probabilistic knowledge fusion,'' in \emph{SIGKDD}.\hskip 1em plus 0.5em
  minus 0.4em\relax ACM, 2014, pp. 601--610.

\bibitem{nickel2015review}
M.~Nickel, K.~Murphy, V.~Tresp, and E.~Gabrilovich, ``A review of relational
  machine learning for knowledge graphs,'' \emph{Proceedings of the IEEE}, vol.
  104, no.~1, pp. 11--33, 2016.

\bibitem{wang2017knowledge}
Q.~Wang, Z.~Mao, B.~Wang, and L.~Guo, ``Knowledge graph embedding: A survey of
  approaches and applications,'' \emph{IEEE TKDE}, vol.~29, no.~12, pp.
  2724--2743, 2017.

\bibitem{hogan2020knowledge}
A.~Hogan, E.~Blomqvist, M.~Cochez, C.~d'Amato, G.~de~Melo, C.~Gutierrez,
  J.~E.~L. Gayo, S.~Kirrane, S.~Neumaier, A.~Polleres \emph{et~al.},
  ``Knowledge graphs,'' \emph{arXiv preprint arXiv:2003.02320}, 2020.

\bibitem{stokman1988structuring}
F.~N. Stokman and P.~H. de~Vries, ``Structuring knowledge in a graph,'' in
  \emph{Human-Computer Interaction}, 1988, pp. 186--206.

\bibitem{bordes2011learning}
A.~Bordes, J.~Weston, R.~Collobert, and Y.~Bengio, ``Learning structured
  embeddings of knowledge bases,'' in \emph{AAAI}, 2011, pp. 301--306.

\bibitem{lin2018knowledge}
Y.~Lin, X.~Han, R.~Xie, Z.~Liu, and M.~Sun, ``Knowledge representation
  learning: A quantitative review,'' \emph{arXiv preprint arXiv:1812.10901},
  2018.

\bibitem{richens1956preprogramming}
R.~H. Richens, ``Preprogramming for mechanical translation.'' \emph{Mechanical
  Translation}, vol.~3, no.~1, pp. 20--25, 1956.

\bibitem{paulheim2017knowledge}
H.~Paulheim, ``Knowledge graph refinement: A survey of approaches and
  evaluation methods,'' \emph{Semantic web}, vol.~8, no.~3, pp. 489--508, 2017.

\bibitem{ehrlinger2016towards}
L.~Ehrlinger and W.~W{\"o}{\ss}, ``Towards a definition of knowledge graphs,''
  \emph{SEMANTiCS (Posters, Demos, SuCCESS)}, vol.~48, pp. 1--4, 2016.

\bibitem{wu2018survey}
T.~Wu, G.~Qi, C.~Li, and M.~Wang, ``A survey of techniques for constructing
  chinese knowledge graphs and their applications,'' \emph{Sustainability},
  vol.~10, no.~9, p. 3245, 2018.

\bibitem{chen2020review}
X.~Chen, S.~Jia, and Y.~Xiang, ``A review: Knowledge reasoning over knowledge
  graph,'' \emph{Expert Systems with Applications}, vol. 141, p. 112948, 2020.

\bibitem{ebisu2018toruse}
T.~Ebisu and R.~Ichise, ``{TorusE}: Knowledge graph embedding on a lie group,''
  in \emph{AAAI}, 2018, pp. 1819--1826.

\bibitem{bordes2013translating}
A.~Bordes, N.~Usunier, A.~Garcia-Duran, J.~Weston, and O.~Yakhnenko,
  ``Translating embeddings for modeling multi-relational data,'' in
  \emph{NIPS}, 2013, pp. 2787--2795.

\bibitem{lin2015learning}
Y.~Lin, Z.~Liu, M.~Sun, Y.~Liu, and X.~Zhu, ``Learning entity and relation
  embeddings for knowledge graph completion,'' in \emph{AAAI}, 2015, pp.
  2181--2187.

\bibitem{socher2013reasoning}
R.~Socher, D.~Chen, C.~D. Manning, and A.~Ng, ``Reasoning with neural tensor
  networks for knowledge base completion,'' in \emph{NIPS}, 2013, pp. 926--934.

\bibitem{zhang2020learning}
Z.~Zhang, J.~Cai, Y.~Zhang, and J.~Wang, ``Learning hierarchy-aware knowledge
  graph embeddings for link prediction.'' in \emph{AAAI}, 2020, pp. 3065--3072.

\bibitem{wang2014knowledge}
Z.~Wang, J.~Zhang, J.~Feng, and Z.~Chen, ``Knowledge graph embedding by
  translating on hyperplanes,'' in \emph{AAAI}, 2014, pp. 1112--1119.

\bibitem{nickel2016holographic}
M.~Nickel, L.~Rosasco, and T.~Poggio, ``Holographic embeddings of knowledge
  graphs,'' in \emph{AAAI}, 2016, pp. 1955--1961.

\bibitem{liu2017analogical}
H.~Liu, Y.~Wu, and Y.~Yang, ``Analogical inference for multi-relational
  embeddings,'' in \emph{ICML}, 2017, pp. 2168--2178.

\bibitem{trouillon2016complex}
T.~Trouillon, J.~Welbl, S.~Riedel, {\'E}.~Gaussier, and G.~Bouchard, ``Complex
  embeddings for simple link prediction,'' in \emph{ICML}, 2016, pp.
  2071--2080.

\bibitem{sun2018rotate}
Z.~Sun, Z.-H. Deng, J.-Y. Nie, and J.~Tang, ``{RotatE}: Knowledge graph
  embedding by relational rotation in complex space,'' in \emph{ICLR}, 2019,
  pp. 1--18.

\bibitem{zhang2019quaternion}
S.~Zhang, Y.~Tay, L.~Yao, and Q.~Liu, ``Quaternion knowledge graph embedding,''
  in \emph{NeurIPS}, 2019, pp. 2731--2741.

\bibitem{he2015learning}
S.~He, K.~Liu, G.~Ji, and J.~Zhao, ``Learning to represent knowledge graphs
  with gaussian embedding,'' in \emph{CIKM}, 2015, pp. 623--632.

\bibitem{xiao2016transg}
H.~Xiao, M.~Huang, and X.~Zhu, ``{TransG}: A generative model for knowledge
  graph embedding,'' in \emph{ACL}, vol.~1, 2016, pp. 2316--2325.

\bibitem{xiao2016one}
------, ``From one point to a manifold: Orbit models for knowledge graph
  embedding,'' in \emph{IJCAI}, 2016, pp. 1315--1321.

\bibitem{balazevic2019multi}
I.~Balazevic, C.~Allen, and T.~Hospedales, ``Multi-relational poincar{\'e}
  graph embeddings,'' in \emph{NeurIPS}, 2019, pp. 4463--4473.

\bibitem{chami2020low}
I.~Chami, A.~Wolf, D.-C. Juan, F.~Sala, S.~Ravi, and C.~R{\'e},
  ``Low-dimensional hyperbolic knowledge graph embeddings,'' in \emph{ACL},
  2020.

\bibitem{xu2019relation}
C.~Xu and R.~Li, ``Relation embedding with dihedral group in knowledge graph,''
  in \emph{ACL}, 2019, pp. 263--272.

\bibitem{yang2014embedding}
B.~Yang, W.-t. Yih, X.~He, J.~Gao, and L.~Deng, ``Embedding entities and
  relations for learning and inference in knowledge bases,'' in \emph{ICLR},
  2015, pp. 1--13.

\bibitem{ji2015knowledge}
G.~Ji, S.~He, L.~Xu, K.~Liu, and J.~Zhao, ``Knowledge graph embedding via
  dynamic mapping matrix,'' in \emph{ACL-IJCNLP}, vol.~1, 2015, pp. 687--696.

\bibitem{xiao2015transa}
H.~Xiao, M.~Huang, Y.~Hao, and X.~Zhu, ``{TransA}: An adaptive approach for
  knowledge graph embedding,'' in \emph{AAAI}, 2015, pp. 1--7.

\bibitem{feng2016knowledge}
J.~Feng, M.~Huang, M.~Wang, M.~Zhou, Y.~Hao, and X.~Zhu, ``Knowledge graph
  embedding by flexible translation,'' in \emph{KR}, 2016, pp. 557--560.

\bibitem{xie2017interpretable}
Q.~Xie, X.~Ma, Z.~Dai, and E.~Hovy, ``An interpretable knowledge transfer model
  for knowledge base completion,'' in \emph{ACL}, 2017, pp. 950--962.

\bibitem{qian2018translating}
W.~Qian, C.~Fu, Y.~Zhu, D.~Cai, and X.~He, ``Translating embeddings for
  knowledge graph completion with relation attention mechanism.'' in
  \emph{IJCAI}, 2018, pp. 4286--4292.

\bibitem{yang2019transms}
S.~Yang, J.~Tian, H.~Zhang, J.~Yan, H.~He, and Y.~Jin, ``{TransMS}: knowledge
  graph embedding for complex relations by multidirectional semantics,'' in
  \emph{IJCAI}, 2019, pp. 1935--1942.

\bibitem{bordes2014semantic}
A.~Bordes, X.~Glorot, J.~Weston, and Y.~Bengio, ``A semantic matching energy
  function for learning with multi-relational data,'' \emph{Machine Learning},
  vol.~94, no.~2, pp. 233--259, 2014.

\bibitem{xue2018expanding}
Y.~Xue, Y.~Yuan, Z.~Xu, and A.~Sabharwal, ``Expanding holographic embeddings
  for knowledge completion,'' in \emph{NeurIPS}, 2018, pp. 4491--4501.

\bibitem{hayashi2017equivalence}
K.~Hayashi and M.~Shimbo, ``On the equivalence of holographic and complex
  embeddings for link prediction,'' in \emph{ACL}, 2017, pp. 554--559.

\bibitem{zhang2019interaction}
W.~Zhang, B.~Paudel, W.~Zhang, A.~Bernstein, and H.~Chen, ``Interaction
  embeddings for prediction and explanation in knowledge graphs,'' in
  \emph{WSDM}, 2019, pp. 96--104.

\bibitem{nguyen2017novel}
D.~Q. Nguyen, T.~D. Nguyen, D.~Q. Nguyen, and D.~Phung, ``A novel embedding
  model for knowledge base completion based on convolutional neural network,''
  in \emph{NAACL}, 2018, pp. 327--333.

\bibitem{shang2018end}
C.~Shang, Y.~Tang, J.~Huang, J.~Bi, X.~He, and B.~Zhou, ``End-to-end
  structure-aware convolutional networks for knowledge base completion,'' in
  \emph{AAAI}, vol.~33, 2019, pp. 3060--3067.

\bibitem{guo2019learning}
L.~Guo, Z.~Sun, and W.~Hu, ``Learning to exploit long-term relational
  dependencies in knowledge graphs,'' in \emph{ICML}, 2019, pp. 2505--2514.

\bibitem{wang2019coke}
Q.~Wang, P.~Huang, H.~Wang, S.~Dai, W.~Jiang, J.~Liu, Y.~Lyu, Y.~Zhu, and
  H.~Wu, ``{CoKE}: Contextualized knowledge graph embedding,'' \emph{arXiv
  preprint arXiv:1911.02168}, 2019.

\bibitem{wang2018multi}
Y.~Wang, R.~Gemulla, and H.~Li, ``On multi-relational link prediction with
  bilinear models,'' in \emph{AAAI}, 2018, pp. 4227--4234.

\bibitem{kazemi2018simple}
S.~M. Kazemi and D.~Poole, ``{SimplE} embedding for link prediction in
  knowledge graphs,'' in \emph{NeurIPS}, 2018, pp. 4284--4295.

\bibitem{nickel2011three}
M.~Nickel, V.~Tresp, and H.-P. Kriegel, ``A three-way model for collective
  learning on multi-relational data,'' in \emph{ICML}, vol.~11, 2011, pp.
  809--816.

\bibitem{nickel2012factorizing}
------, ``Factorizing {YAGO}: scalable machine learning for linked data,'' in
  \emph{WWW}, 2012, pp. 271--280.

\bibitem{jenatton2012latent}
R.~Jenatton, N.~L. Roux, A.~Bordes, and G.~R. Obozinski, ``A latent factor
  model for highly multi-relational data,'' in \emph{NIPS}, 2012, pp.
  3167--3175.

\bibitem{balavzevic2019tucker}
I.~Bala{\v{z}}evi{\'c}, C.~Allen, and T.~M. Hospedales, ``{TuckER}: Tensor
  factorization for knowledge graph completion,'' in \emph{EMNLP-IJCNLP}, 2019,
  pp. 5185--5194.

\bibitem{amin2020lowfer}
S.~Amin, S.~Varanasi, K.~A. Dunfield, and G.~Neumann, ``{LowFER}: Low-rank
  bilinear pooling for link prediction,'' in \emph{ICML}, 2020, pp. 1--11.

\bibitem{liu2016probabilistic}
Q.~Liu, H.~Jiang, A.~Evdokimov, Z.-H. Ling, X.~Zhu, S.~Wei, and Y.~Hu,
  ``Probabilistic reasoning via deep learning: Neural association models,''
  \emph{arXiv preprint arXiv:1603.07704}, 2016.

\bibitem{dettmers2018convolutional}
T.~Dettmers, P.~Minervini, P.~Stenetorp, and S.~Riedel, ``Convolutional 2d
  knowledge graph embeddings,'' in \emph{AAAI}, vol.~32, 2018, pp. 1811--1818.

\bibitem{balavzevic2019hypernetwork}
I.~Bala{\v{z}}evi{\'c}, C.~Allen, and T.~M. Hospedales, ``Hypernetwork
  knowledge graph embeddings,'' in \emph{ICANN}, 2019, pp. 553--565.

\bibitem{gardner2014incorporating}
M.~Gardner, P.~Talukdar, J.~Krishnamurthy, and T.~Mitchell, ``Incorporating
  vector space similarity in random walk inference over knowledge bases,'' in
  \emph{EMNLP}, 2014, pp. 397--406.

\bibitem{neelakantan2015compositional}
A.~Neelakantan, B.~Roth, and A.~McCallum, ``Compositional vector space models
  for knowledge base completion,'' in \emph{ACL-IJCNLP}, vol.~1, 2015, pp.
  156--166.

\bibitem{yao2019kgbert}
L.~Yao, C.~Mao, and Y.~Luo, ``{KG-BERT}: {BERT} for knowledge graph
  completion,'' \emph{arXiv preprent arXiv:1909.03193}, 2019.

\bibitem{schlichtkrull2018modeling}
M.~Schlichtkrull, T.~N. Kipf, P.~Bloem, R.~Van Den~Berg, I.~Titov, and
  M.~Welling, ``Modeling relational data with graph convolutional networks,''
  in \emph{ESWC}, 2018, pp. 593--607.

\bibitem{kipf2016semi}
T.~N. Kipf and M.~Welling, ``Semi-supervised classification with graph
  convolutional networks,'' in \emph{ICLR}, 2017, pp. 1--14.

\bibitem{nathani2019learning}
D.~Nathani, J.~Chauhan, C.~Sharma, and M.~Kaul, ``Learning attention-based
  embeddings for relation prediction in knowledge graphs,'' in \emph{ACL},
  2019, pp. 4710--4723.

\bibitem{vashishth2020composition}
S.~Vashishth, S.~Sanyal, V.~Nitin, and P.~Talukdar, ``Composition-based
  multi-relational graph convolutional networks,'' in \emph{ICLR}, 2020, pp.
  1--15.

\bibitem{wang2014knowledge_text}
Z.~Wang, J.~Zhang, J.~Feng, and Z.~Chen, ``Knowledge graph and text jointly
  embedding,'' in \emph{EMNLP}, 2014, pp. 1591--1601.

\bibitem{xie2016entity_desc}
R.~Xie, Z.~Liu, J.~Jia, H.~Luan, and M.~Sun, ``Representation learning of
  knowledge graphs with entity descriptions,'' in \emph{AAAI}, 2016, pp.
  2659--2665.

\bibitem{xiao2017ssp}
H.~Xiao, M.~Huang, L.~Meng, and X.~Zhu, ``{SSP}: semantic space projection for
  knowledge graph embedding with text descriptions,'' in \emph{AAAI}, 2017, pp.
  3104--3110.

\bibitem{guo2015semantically}
S.~Guo, Q.~Wang, B.~Wang, L.~Wang, and L.~Guo, ``Semantically smooth knowledge
  graph embedding,'' in \emph{ACL-IJCNLP}, vol.~1, 2015, pp. 84--94.

\bibitem{xie2016representation_type}
R.~Xie, Z.~Liu, and M.~Sun, ``Representation learning of knowledge graphs with
  hierarchical types,'' in \emph{IJCAI}, 2016, pp. 2965--2971.

\bibitem{lin2016knowledge}
Y.~Lin, Z.~Liu, and M.~Sun, ``Knowledge representation learning with entities,
  attributes and relations,'' in \emph{IJCAI}, 2016, pp. 2866--2872.

\bibitem{zhang2018knowledge}
Z.~Zhang, F.~Zhuang, M.~Qu, F.~Lin, and Q.~He, ``Knowledge graph embedding with
  hierarchical relation structure,'' in \emph{EMNLP}, 2018, pp. 3198--3207.

\bibitem{xie2017image}
R.~Xie, Z.~Liu, H.~Luan, and M.~Sun, ``Image-embodied knowledge representation
  learning,'' in \emph{IJCAI}, 2017, pp. 3140--3146.

\bibitem{wu2012probase}
W.~Wu, H.~Li, H.~Wang, and K.~Q. Zhu, ``Probase: A probabilistic taxonomy for
  text understanding,'' in \emph{SIGMOD}, 2012, pp. 481--492.

\bibitem{carlson2010toward}
A.~Carlson, J.~Betteridge, B.~Kisiel, B.~Settles, E.~R. Hruschka, and T.~M.
  Mitchell, ``Toward an architecture for never-ending language learning,'' in
  \emph{AAAI}, 2010, pp. 1306--1313.

\bibitem{speer2017conceptnet}
R.~Speer, J.~Chin, and C.~Havasi, ``{ConceptNet 5.5: an open multilingual graph
  of general knowledge},'' in \emph{Proceedings of AAAI}, 2017, pp. 4444--4451.

\bibitem{chen2019embedding}
X.~Chen, M.~Chen, W.~Shi, Y.~Sun, and C.~Zaniolo, ``Embedding uncertain
  knowledge graphs,'' in \emph{AAAI}, vol.~33, 2019, pp. 3363--3370.

\bibitem{tabacof2019probability}
P.~Tabacof and L.~Costabello, ``Probability calibration for knowledge graph
  embedding models,'' in \emph{ICLR}, 2019.

\bibitem{safavi2020evaluating}
T.~Safavi, D.~Koutra, and E.~Meij, ``{Evaluating the Calibration of Knowledge
  Graph Embeddings for Trustworthy Link Prediction},'' in \emph{Proceedings of
  EMNLP}, 2020, pp. 8308--8321.

\bibitem{han2018neural}
X.~Han, Z.~Liu, and M.~Sun, ``Neural knowledge acquisition via mutual attention
  between knowledge graph and text,'' in \emph{AAAI}, 2018, pp. 4832--4839.

\bibitem{dong2019triple}
T.~Dong, Z.~Wang, J.~Li, C.~Bauckhage, and A.~B. Cremers, ``Triple
  classification using regions and fine-grained entity typing,'' in
  \emph{AAAI}, vol.~33, 2019, pp. 77--85.

\bibitem{zhou2016attention}
P.~Zhou, W.~Shi, J.~Tian, Z.~Qi, B.~Li, H.~Hao, and B.~Xu, ``Attention-based
  bidirectional long short-term memory networks for relation classification,''
  in \emph{ACL}, vol.~2, 2016, pp. 207--212.

\bibitem{cao2020open}
E.~Cao, D.~Wang, J.~Huang, and W.~Hu, ``Open knowledge enrichment for long-tail
  entities,'' in \emph{The Web Conference}, 2020, pp. 384--394.

\bibitem{shi2017proje}
B.~Shi and T.~Weninger, ``{ProjE}: Embedding projection for knowledge graph
  completion,'' in \emph{AAAI}, 2017, pp. 1236--1242.

\bibitem{guan2018shared}
S.~Guan, X.~Jin, Y.~Wang, and X.~Cheng, ``Shared embedding based neural
  networks for knowledge graph completion,'' in \emph{CIKM}, 2018, pp.
  247--256.

\bibitem{shi2018open}
B.~Shi and T.~Weninger, ``Open-world knowledge graph completion,'' in
  \emph{AAAI}, 2018, pp. 1957--1964.

\bibitem{zhang2018generative}
C.~Zhang, Y.~Li, N.~Du, W.~Fan, and P.~S. Yu, ``On the generative discovery of
  structured medical knowledge,'' in \emph{SIGKDD}, 2018, pp. 2720--2728.

\bibitem{lao2010relational}
N.~Lao and W.~W. Cohen, ``Relational retrieval using a combination of
  path-constrained random walks,'' \emph{Machine learning}, vol.~81, no.~1, pp.
  53--67, 2010.

\bibitem{das2017chains}
R.~Das, A.~Neelakantan, D.~Belanger, and A.~McCallum, ``Chains of reasoning
  over entities, relations, and text using recurrent neural networks,'' in
  \emph{EACL}, vol.~1, 2017, pp. 132--141.

\bibitem{chen2018variational}
W.~Chen, W.~Xiong, X.~Yan, and W.~Y. Wang, ``Variational knowledge graph
  reasoning,'' in \emph{NAACL}, 2018, pp. 1823--1832.

\bibitem{xiong2017deeppath}
W.~Xiong, T.~Hoang, and W.~Y. Wang, ``{DeepPath}: A reinforcement learning
  method for knowledge graph reasoning,'' in \emph{EMNLP}, 2017, pp. 564--573.

\bibitem{das2017go}
R.~Das, S.~Dhuliawala, M.~Zaheer, L.~Vilnis, I.~Durugkar, A.~Krishnamurthy,
  A.~Smola, and A.~McCallum, ``Go for a walk and arrive at the answer:
  Reasoning over paths in knowledge bases using reinforcement learning,'' in
  \emph{ICLR}, 2018, pp. 1--18.

\bibitem{lin2018multi}
X.~V. Lin, R.~Socher, and C.~Xiong, ``Multi-hop knowledge graph reasoning with
  reward shaping,'' in \emph{EMNLP}, 2018, pp. 3243--3253.

\bibitem{shen2018m}
Y.~Shen, J.~Chen, P.-S. Huang, Y.~Guo, and J.~Gao, ``{M-Walk}: Learning to walk
  over graphs using monte carlo tree search,'' in \emph{NeurIPS}, 2018, pp.
  6786--6797.

\bibitem{fu2019collaborative}
C.~Fu, T.~Chen, M.~Qu, W.~Jin, and X.~Ren, ``Collaborative policy learning for
  open knowledge graph reasoning,'' in \emph{EMNLP}, 2019, pp. 2672--2681.

\bibitem{galarraga2013amie}
L.~A. Gal{\'a}rraga, C.~Teflioudi, K.~Hose, and F.~Suchanek, ``{AMIE}:
  association rule mining under incomplete evidence in ontological knowledge
  bases,'' in \emph{WWW}, 2013, pp. 413--422.

\bibitem{omran2019embedding}
P.~G. Omran, K.~Wang, and Z.~Wang, ``An embedding-based approach to rule
  learning in knowledge graphs,'' \emph{IEEE TKDE}, pp. 1--12, 2019.

\bibitem{guo2016jointly}
S.~Guo, Q.~Wang, L.~Wang, B.~Wang, and L.~Guo, ``Jointly embedding knowledge
  graphs and logical rules,'' in \emph{EMNLP}, 2016, pp. 192--202.

\bibitem{guo2018knowledge}
------, ``Knowledge graph embedding with iterative guidance from soft rules,''
  in \emph{AAAI}, 2018, pp. 4816--4823.

\bibitem{zhang2019iteratively}
W.~Zhang, B.~Paudel, L.~Wang, J.~Chen, H.~Zhu, W.~Zhang, A.~Bernstein, and
  H.~Chen, ``Iteratively learning embeddings and rules for knowledge graph
  reasoning,'' in \emph{WWW}, 2019, pp. 2366--2377.

\bibitem{rocktaschel2017end}
T.~Rockt{\"a}schel and S.~Riedel, ``End-to-end differentiable proving,'' in
  \emph{NIPS}, 2017, pp. 3788--3800.

\bibitem{yang2017differentiable}
F.~Yang, Z.~Yang, and W.~W. Cohen, ``Differentiable learning of logical rules
  for knowledge base reasoning,'' in \emph{NIPS}, 2017, pp. 2319--2328.

\bibitem{wang2020differentiable}
P.-W. Wang, D.~Stepanova, C.~Domokos, and J.~Z. Kolter, ``Differentiable
  learning of numerical rules in knowledge graphs,'' in \emph{ICLR}, 2020, pp.
  1--12.

\bibitem{qu2019probabilistic}
M.~Qu and J.~Tang, ``Probabilistic logic neural networks for reasoning,'' in
  \emph{NeurIPS}, 2019, pp. 7710--7720.

\bibitem{zhang2020efficient}
Y.~Zhang, X.~Chen, Y.~Yang, A.~Ramamurthy, B.~Li, Y.~Qi, and L.~Song,
  ``Efficient probabilistic logic reasoning with graph neural networks,'' in
  \emph{ICLR}, 2020, pp. 1--20.

\bibitem{xiong2018one}
W.~Xiong, M.~Yu, S.~Chang, X.~Guo, and W.~Y. Wang, ``One-shot relational
  learning for knowledge graphs,'' in \emph{EMNLP}, 2018, pp. 1980--1990.

\bibitem{lv2019adapting}
X.~Lv, Y.~Gu, X.~Han, L.~Hou, J.~Li, and Z.~Liu, ``Adapting meta knowledge
  graph information for multi-hop reasoning over few-shot relations,'' in
  \emph{EMNLP-IJCNLP}, 2019, pp. 3374--3379.

\bibitem{chen2019meta}
M.~Chen, W.~Zhang, W.~Zhang, Q.~Chen, and H.~Chen, ``Meta relational learning
  for few-shot link prediction in knowledge graphs,'' in \emph{EMNLP-IJCNLP},
  2019, pp. 4217--4226.

\bibitem{zhang2020few}
C.~Zhang, H.~Yao, C.~Huang, M.~Jiang, Z.~Li, and N.~V. Chawla, ``Few-shot
  knowledge graph completion,'' in \emph{AAAI}, 2020, pp. 1--8.

\bibitem{qin2020generative}
P.~Qin, X.~Wang, W.~Chen, C.~Zhang, W.~Xu, and W.~Y. Wang, ``Generative
  adversarial zero-shot relational learning for knowledge graphs,'' in
  \emph{AAAI}, 2020, pp. 1--8.

\bibitem{baek2020learning}
J.~Baek, D.~B. Lee, and S.~J. Hwang, ``{Learning to Extrapolate Knowledge:
  Transductive Few-shot Out-of-Graph Link Prediction},'' in \emph{NeurIPS},
  2020.

\bibitem{chiu2016named}
J.~P. Chiu and E.~Nichols, ``Named entity recognition with bidirectional
  {LSTM-CNNs},'' \emph{Transactions of ACL}, vol.~4, pp. 357--370, 2016.

\bibitem{lample2016neural}
G.~Lample, M.~Ballesteros, S.~Subramanian, K.~Kawakami, and C.~Dyer, ``Neural
  architectures for named entity recognition,'' in \emph{NAACL}, 2016, pp.
  260--270.

\bibitem{xia2019multi}
C.~Xia, C.~Zhang, T.~Yang, Y.~Li, N.~Du, X.~Wu, W.~Fan, F.~Ma, and P.~Yu,
  ``Multi-grained named entity recognition,'' in \emph{ACL}, 2019, pp.
  1430--1440.

\bibitem{hu2020leveraging}
A.~Hu, Z.~Dou, J.-Y. Nie, and J.-R. Wen, ``Leveraging multi-token entities in
  document-level named entity recognition.'' in \emph{AAAI}, 2020, pp.
  7961--7968.

\bibitem{li2020unified}
X.~Li, J.~Feng, Y.~Meng, Q.~Han, F.~Wu, and J.~Li, ``A unified {MRC} framework
  for named entity recognition,'' in \emph{ACL}, 2020, pp. 5849--5859.

\bibitem{sun2020ernie}
Y.~Sun, S.~Wang, Y.~Li, S.~Feng, H.~Tian, H.~Wu, and H.~Wang, ``{ERNIE} 2.0: A
  continual pre-training framework for language understanding,'' in
  \emph{AAAI}, 2020, pp. 8968--8975.

\bibitem{weijie2019kbert}
W.~Liu, P.~Zhou, Z.~Zhao, Z.~Wang, Q.~Ju, H.~Deng, and P.~Wang, ``{K-BERT}:
  Enabling language representation with knowledge graph,'' in \emph{AAAI},
  2020, pp. 1--8.

\bibitem{ren2016label}
X.~Ren, W.~He, M.~Qu, C.~R. Voss, H.~Ji, and J.~Han, ``Label noise reduction in
  entity typing by heterogeneous partial-label embedding,'' in \emph{SIGKDD},
  2016, pp. 1825--1834.

\bibitem{ma2016label}
Y.~Ma, E.~Cambria, and S.~Gao, ``Label embedding for zero-shot fine-grained
  named entity typing,'' in \emph{COLING}, 2016, pp. 171--180.

\bibitem{hao2019universal}
J.~Hao, M.~Chen, W.~Yu, Y.~Sun, and W.~Wang, ``Universal representation
  learning of knowledge bases by jointly embedding instances and ontological
  concepts,'' in \emph{KDD}, 2019, pp. 1709--1719.

\bibitem{zhao2020connecting}
Y.~Zhao, R.~Xie, K.~Liu, W.~Xiaojie \emph{et~al.}, ``Connecting embeddings for
  knowledge graph entity typing,'' in \emph{ACL}, 2020, pp. 6419--6428.

\bibitem{huang2015leveraging}
H.~Huang, L.~Heck, and H.~Ji, ``Leveraging deep neural networks and knowledge
  graphs for entity disambiguation,'' \emph{arXiv preprint arXiv:1504.07678},
  2015.

\bibitem{fang2016entity}
W.~Fang, J.~Zhang, D.~Wang, Z.~Chen, and M.~Li, ``Entity disambiguation by
  knowledge and text jointly embedding,'' in \emph{SIGNLL}, 2016, pp. 260--269.

\bibitem{ganea2017deep}
O.-E. Ganea and T.~Hofmann, ``Deep joint entity disambiguation with local
  neural attention,'' in \emph{EMNLP}, 2017, pp. 2619--2629.

\bibitem{le2018improving}
P.~Le and I.~Titov, ``Improving entity linking by modeling latent relations
  between mentions,'' in \emph{ACL}, vol.~1, 2018, pp. 1595--1604.

\bibitem{chen2017multilingual}
M.~Chen, Y.~Tian, M.~Yang, and C.~Zaniolo, ``Multilingual knowledge graph
  embeddings for cross-lingual knowledge alignment,'' in \emph{IJCAI}, 2017,
  pp. 1511--1517.

\bibitem{zhu2017iterative}
H.~Zhu, R.~Xie, Z.~Liu, and M.~Sun, ``Iterative entity alignment via joint
  knowledge embeddings,'' in \emph{IJCAI}, 2017, pp. 4258--4264.

\bibitem{sun2018bootstrapping}
Z.~Sun, W.~Hu, Q.~Zhang, and Y.~Qu, ``Bootstrapping entity alignment with
  knowledge graph embedding.'' in \emph{IJCAI}, 2018, pp. 4396--4402.

\bibitem{sun2017cross}
Z.~Sun, W.~Hu, and C.~Li, ``Cross-lingual entity alignment via joint
  attribute-preserving embedding,'' in \emph{ISWC}, 2017, pp. 628--644.

\bibitem{chen2018co}
M.~Chen, Y.~Tian, K.-W. Chang, S.~Skiena, and C.~Zaniolo, ``Co-training
  embeddings of knowledge graphs and entity descriptions for cross-lingual
  entity alignment,'' in \emph{IJCAI}, 2018, pp. 3998--4004.

\bibitem{zhang2019multi}
Q.~Zhang, Z.~Sun, W.~Hu, M.~Chen, L.~Guo, and Y.~Qu, ``Multi-view knowledge
  graph embedding for entity alignment,'' in \emph{IJCAI}, 2019, pp.
  5429--5435.

\bibitem{trsedya2019entity}
B.~D. Trsedya, J.~Qi, and R.~Zhang, ``Entity alignment between knowledge graphs
  using attribute embeddings,'' in \emph{AAAI}, vol.~33, 2019, pp. 297--304.

\bibitem{sun2020benchmarking}
Z.~Sun, Q.~Zhang, W.~Hu, C.~Wang, M.~Chen, F.~Akrami, and C.~Li, ``A
  benchmarking study of embedding-based entity alignment for knowledge
  graphs,'' in \emph{VLDB}, 2020.

\bibitem{craven1999constructing}
M.~Craven, J.~Kumlien \emph{et~al.}, ``Constructing biological knowledge bases
  by extracting information from text sources,'' in \emph{ISMB}, vol. 1999,
  1999, pp. 77--86.

\bibitem{mintz2009distant}
M.~Mintz, S.~Bills, R.~Snow, and D.~Jurafsky, ``Distant supervision for
  relation extraction without labeled data,'' in \emph{ACL and IJCNLP of the
  AFNLP}, 2009, pp. 1003--1011.

\bibitem{qu2019discovering}
J.~Qu, D.~Ouyang, W.~Hua, Y.~Ye, and X.~Zhou, ``Discovering correlations
  between sparse features in distant supervision for relation extraction,'' in
  \emph{WSDM}, 2019, pp. 726--734.

\bibitem{zeng2014relation}
D.~Zeng, K.~Liu, S.~Lai, G.~Zhou, and J.~Zhao, ``Relation classification via
  convolutional deep neural network,'' in \emph{COLING}, 2014, pp. 2335--2344.

\bibitem{nguyen2015relation}
T.~H. Nguyen and R.~Grishman, ``Relation extraction: Perspective from
  convolutional neural networks,'' in \emph{ACL Workshop on Vector Space
  Modeling for Natural Language Processing}, 2015, pp. 39--48.

\bibitem{zeng2015distant}
D.~Zeng, K.~Liu, Y.~Chen, and J.~Zhao, ``Distant supervision for relation
  extraction via piecewise convolutional neural networks,'' in \emph{EMNLP},
  2015, pp. 1753--1762.

\bibitem{jiang2016relation}
X.~Jiang, Q.~Wang, P.~Li, and B.~Wang, ``Relation extraction with
  multi-instance multi-label convolutional neural networks,'' in \emph{COLING},
  2016, pp. 1471--1480.

\bibitem{ye2017jointly}
H.~Ye, W.~Chao, Z.~Luo, and Z.~Li, ``Jointly extracting relations with class
  ties via effective deep ranking,'' in \emph{ACL}, vol.~1, 2017, pp.
  1810--1820.

\bibitem{zeng2017incorporating}
W.~Zeng, Y.~Lin, Z.~Liu, and M.~Sun, ``Incorporating relation paths in neural
  relation extraction,'' in \emph{EMNLP}, 2017, pp. 1768--1777.

\bibitem{xu2015classifying}
Y.~Xu, L.~Mou, G.~Li, Y.~Chen, H.~Peng, and Z.~Jin, ``Classifying relations via
  long short term memory networks along shortest dependency paths,'' in
  \emph{EMNLP}, 2015, pp. 1785--1794.

\bibitem{miwa2016end}
M.~Miwa and M.~Bansal, ``End-to-end relation extraction using lstms on
  sequences and tree structures,'' in \emph{ACL}, vol.~1, 2016, pp. 1105--1116.

\bibitem{cai2016bidirectional}
R.~Cai, X.~Zhang, and H.~Wang, ``Bidirectional recurrent convolutional neural
  network for relation classification,'' in \emph{ACL}, vol.~1, 2016, pp.
  756--765.

\bibitem{shen2016attention}
Y.~Shen and X.~Huang, ``Attention-based convolutional neural network for
  semantic relation extraction,'' in \emph{COLING}, 2016, pp. 2526--2536.

\bibitem{lin2016neural}
Y.~Lin, S.~Shen, Z.~Liu, H.~Luan, and M.~Sun, ``Neural relation extraction with
  selective attention over instances,'' in \emph{ACL}, vol.~1, 2016, pp.
  2124--2133.

\bibitem{ji2017distant}
G.~Ji, K.~Liu, S.~He, and J.~Zhao, ``Distant supervision for relation
  extraction with sentence-level attention and entity descriptions,'' in
  \emph{AAAI}, 2017, pp. 3060--3066.

\bibitem{han2018hierarchical}
X.~Han, P.~Yu, Z.~Liu, M.~Sun, and P.~Li, ``Hierarchical relation extraction
  with coarse-to-fine grained attention,'' in \emph{EMNLP}, 2018, pp.
  2236--2245.

\bibitem{soares2019matching}
L.~B. Soares, N.~FitzGerald, J.~Ling, and T.~Kwiatkowski, ``Matching the
  blanks: Distributional similarity for relation learning,'' in \emph{ACL},
  2019, pp. 2895--2905.

\bibitem{zhang2018graph}
Y.~Zhang, P.~Qi, and C.~D. Manning, ``Graph convolution over pruned dependency
  trees improves relation extraction,'' in \emph{EMNLP}, 2018, pp. 2205--2215.

\bibitem{guo2019attention}
Z.~Guo, Y.~Zhang, and W.~Lu, ``Attention guided graph convolutional networks
  for relation extraction,'' in \emph{ACL}, 2019, pp. 241--251.

\bibitem{zhang2019long}
N.~Zhang, S.~Deng, Z.~Sun, G.~Wang, X.~Chen, W.~Zhang, and H.~Chen, ``Long-tail
  relation extraction via knowledge graph embeddings and graph convolution
  networks,'' in \emph{NAACL}, 2019, pp. 3016--3025.

\bibitem{wu2017adversarial}
Y.~Wu, D.~Bamman, and S.~Russell, ``Adversarial training for relation
  extraction,'' in \emph{EMNLP}, 2017, pp. 1778--1783.

\bibitem{qin2018dsgan}
P.~Qin, X.~Weiran, and W.~Y. Wang, ``{DSGAN}: Generative adversarial training
  for distant supervision relation extraction,'' in \emph{ACL}, vol.~1, 2018,
  pp. 496--505.

\bibitem{qin2018robust}
P.~Qin, W.~Xu, and W.~Y. Wang, ``Robust distant supervision relation extraction
  via deep reinforcement learning,'' in \emph{ACL}, vol.~1, 2018, pp.
  2137--2147.

\bibitem{zeng2018large}
X.~Zeng, S.~He, K.~Liu, and J.~Zhao, ``Large scaled relation extraction with
  reinforcement learning,'' in \emph{AAAI}, 2018, pp. 5658--5665.

\bibitem{feng2018reinforcement}
J.~Feng, M.~Huang, L.~Zhao, Y.~Yang, and X.~Zhu, ``Reinforcement learning for
  relation classification from noisy data,'' in \emph{AAAI}, 2018, pp.
  5779--5786.

\bibitem{takanobu2018hierarchical}
R.~Takanobu, T.~Zhang, J.~Liu, and M.~Huang, ``A hierarchical framework for
  relation extraction with reinforcement learning,'' in \emph{AAAI}, vol.~33,
  2019, pp. 7072--7079.

\bibitem{huang2017deep}
Y.~Huang and W.~Y. Wang, ``Deep residual learning for weakly-supervised
  relation extraction,'' in \emph{EMNLP}, 2017, pp. 1803--1807.

\bibitem{liu2018neural}
T.~Liu, X.~Zhang, W.~Zhou, and W.~Jia, ``Neural relation extraction via
  inner-sentence noise reduction and transfer learning,'' in \emph{EMNLP},
  2018, pp. 2195--2204.

\bibitem{lei2018cooperative}
K.~Lei, D.~Chen, Y.~Li, N.~Du, M.~Yang, W.~Fan, and Y.~Shen, ``Cooperative
  denoising for distantly supervised relation extraction,'' in \emph{COLING},
  2018, pp. 426--436.

\bibitem{jiang2019relation}
H.~Jiang, L.~Cui, Z.~Xu, D.~Yang, J.~Chen, C.~Li, J.~Liu, J.~Liang, C.~Wang,
  Y.~Xiao, and W.~Wang, ``Relation extraction using supervision from topic
  knowledge of relation labels,'' in \emph{IJCAI}, 2019, pp. 5024--5030.

\bibitem{shahbazi2020relation}
H.~Shahbazi, X.~Z. Fern, R.~Ghaeini, and P.~Tadepalli, ``Relation extraction
  with explanation,'' in \emph{ACL}, 2020, pp. 6488--6494.

\bibitem{gao2019hybrid}
T.~Gao, X.~Han, Z.~Liu, and M.~Sun, ``Hybrid attention-based prototypical
  networks for noisy few-shot relation classification,'' in \emph{AAAI},
  vol.~33, 2019, pp. 6407--6414.

\bibitem{qu2020few}
M.~Qu, T.~Gao, L.-P.~A. Xhonneux, and J.~Tang, ``Few-shot relation extraction
  via bayesian meta-learning on relation graphs,'' in \emph{ICML}, 2020, pp.
  1--10.

\bibitem{miwa2014modeling}
M.~Miwa and Y.~Sasaki, ``Modeling joint entity and relation extraction with
  table representation,'' in \emph{EMNLP}, 2014, pp. 1858--1869.

\bibitem{katiyar2017going}
A.~Katiyar and C.~Cardie, ``Going out on a limb: Joint extraction of entity
  mentions and relations without dependency trees,'' in \emph{ACL}, 2017, pp.
  917--928.

\bibitem{zheng2017joint}
S.~Zheng, F.~Wang, H.~Bao, Y.~Hao, P.~Zhou, and B.~Xu, ``Joint extraction of
  entities and relations based on a novel tagging scheme,'' in \emph{ACL},
  2017, pp. 1227--1236.

\bibitem{li2019entity}
X.~Li, F.~Yin, Z.~Sun, X.~Li, A.~Yuan, D.~Chai, M.~Zhou, and J.~Li,
  ``Entity-relation extraction as multi-turn question answering,'' in
  \emph{ACL}, 2019, pp. 1340--1350.

\bibitem{dai2019joint}
D.~Dai, X.~Xiao, Y.~Lyu, S.~Dou, Q.~She, and H.~Wang, ``Joint extraction of
  entities and overlapping relations using position-attentive sequence
  labeling,'' in \emph{AAAI}, vol.~33, 2019, pp. 6300--6308.

\bibitem{wei2020novel}
Z.~Wei, J.~Su, Y.~Wang, Y.~Tian, and Y.~Chang, ``A novel cascade binary tagging
  framework for relational triple extraction,'' in \emph{ACL}, 2020, pp.
  1476--1488.

\bibitem{wang2020tplinker}
Y.~Wang, B.~Yu, Y.~Zhang, T.~Liu, H.~Zhu, and L.~Sun, ``Tplinker: Single-stage
  joint extraction of entities and relations through token pair linking,'' in
  \emph{COLING}, 2020, pp. 1572--1582.

\bibitem{zhong2020frustratingly}
Z.~Zhong and D.~Chen, ``A frustratingly easy approach for joint entity and
  relation extraction,'' \emph{arXiv preprint arXiv:2010.12812}, 2020.

\bibitem{xu2020inductive}
D.~Xu, C.~Ruan, E.~Korpeoglu, S.~Kumar, and K.~Achan, ``Inductive
  representation learning on temporal graphs,'' in \emph{ICLR}, 2020, pp.
  1--19.

\bibitem{leblay2018deriving}
J.~Leblay and M.~W. Chekol, ``Deriving validity time in knowledge graph,'' in
  \emph{WWW}, 2018, pp. 1771--1776.

\bibitem{ma2019embedding}
Y.~Ma, V.~Tresp, and E.~A. Daxberger, ``Embedding models for episodic knowledge
  graphs,'' \emph{Journal of Web Semantics}, vol.~59, p. 100490, 2019.

\bibitem{dasgupta2018hyte}
S.~S. Dasgupta, S.~N. Ray, and P.~Talukdar, ``Hyte: Hyperplane-based temporally
  aware knowledge graph embedding,'' in \emph{EMNLP}, 2018, pp. 2001--2011.

\bibitem{garcia2018learning}
A.~Garc{\'\i}a-Dur{\'a}n, S.~Duman{\v{c}}i{\'c}, and M.~Niepert, ``Learning
  sequence encoders for temporal knowledge graph completion,'' in \emph{EMNLP},
  2018, pp. 4816--4821.

\bibitem{liu2019context}
Y.~Liu, W.~Hua, K.~Xin, and X.~Zhou, ``Context-aware temporal knowledge graph
  embedding,'' in \emph{WISE}, 2019, pp. 583--598.

\bibitem{lacroix2020tensor}
T.~Lacroix, G.~Obozinski, and N.~Usunier, ``Tensor decompositions for temporal
  knowledge base completion,'' in \emph{ICLR}, 2020, pp. 1--12.

\bibitem{wijaya2014ctps}
D.~T. Wijaya, N.~Nakashole, and T.~M. Mitchell, ``{CTPs}: Contextual temporal
  profiles for time scoping facts using state change detection,'' in
  \emph{EMNLP}, 2014, pp. 1930--1936.

\bibitem{goel2020diachronic}
R.~Goel, S.~M. Kazemi, M.~Brubaker, and P.~Poupart, ``Diachronic embedding for
  temporal knowledge graph completion,'' in \emph{AAAI}, 2020, pp. 3988--3995.

\bibitem{trivedi2017know}
R.~Trivedi, H.~Dai, Y.~Wang, and L.~Song, ``Know-evolve: Deep temporal
  reasoning for dynamic knowledge graphs,'' in \emph{ICML}, 2017, pp.
  3462--3471.

\bibitem{jin2019recurrent}
W.~Jin, C.~Zhang, P.~Szekely, and X.~Ren, ``Recurrent event network for
  reasoning over temporal knowledge graphs,'' in \emph{ICLR RLGM Workshop},
  2019.

\bibitem{jiang2016towards}
T.~Jiang, T.~Liu, T.~Ge, L.~Sha, B.~Chang, S.~Li, and Z.~Sui, ``Towards
  time-aware knowledge graph completion,'' in \emph{COLING}, 2016, pp.
  1715--1724.

\bibitem{jiang2016encoding}
T.~Jiang, T.~Liu, T.~Ge, L.~Sha, S.~Li, B.~Chang, and Z.~Sui, ``Encoding
  temporal information for time-aware link prediction,'' in \emph{EMNLP}, 2016,
  pp. 2350--2354.

\bibitem{chekol2017marrying}
M.~W. Chekol, G.~Pirr{\`o}, J.~Schoenfisch, and H.~Stuckenschmidt, ``Marrying
  uncertainty and time in knowledge graphs,'' in \emph{AAAI}, 2017, pp. 88--94.

\bibitem{logan2019barack}
R.~Logan, N.~F. Liu, M.~E. Peters, M.~Gardner, and S.~Singh, ``Barack's wife
  hillary: Using knowledge graphs for fact-aware language modeling,'' in
  \emph{ACL}, 2019, pp. 5962--5971.

\bibitem{zhang2019ernie}
Z.~Zhang, X.~Han, Z.~Liu, X.~Jiang, M.~Sun, and Q.~Liu, ``{ERNIE}: Enhanced
  language representation with informative entities,'' in \emph{ACL}, 2019, pp.
  1441--1451.

\bibitem{sun2019ernie}
Y.~Sun, S.~Wang, Y.~Li, S.~Feng, X.~Chen, H.~Zhang, X.~Tian, D.~Zhu, H.~Tian,
  and H.~Wu, ``{ERNIE}: Enhanced representation through knowledge
  integration,'' \emph{arXiv preprint arXiv:1904.09223}, 2019.

\bibitem{wang2020kepler}
X.~Wang, T.~Gao, Z.~Zhu, Z.~Liu, J.~Li, and J.~Tang, ``{KEPLER: A unified model
  for knowledge embedding and pre-trained language representation},''
  \emph{TACL}, 2020.

\bibitem{shen2020exploiting}
T.~Shen, Y.~Mao, P.~He, G.~Long, A.~Trischler, and W.~Chen, ``Exploiting
  structured knowledge in text via graph-guided representation learning,'' in
  \emph{EMNLP}, 2020.

\bibitem{sun2020colake}
T.~Sun, Y.~Shao, X.~Qiu, Q.~Guo, Y.~Hu, X.-J. Huang, and Z.~Zhang, ``{CoLAKE:
  Contextualized Language and Knowledge Embedding},'' in \emph{COLING}, 2020,
  pp. 3660--3670.

\bibitem{he2020integrating}
B.~He, D.~Zhou, J.~Xiao, Q.~Liu, N.~J. Yuan, T.~Xu \emph{et~al.}, ``{BERT-MK:
  Integrating Graph Contextualized Knowledge into Pre-trained Language
  Models},'' in \emph{Findings of EMNLP}, 2020, p. 2281–2290.

\bibitem{petroni2019language}
F.~Petroni, T.~Rockt{\"a}schel, S.~Riedel, P.~Lewis, A.~Bakhtin, Y.~Wu, and
  A.~Miller, ``Language models as knowledge bases?'' in \emph{EMNLP-IJCNLP},
  2019, pp. 2463--2473.

\bibitem{dai2016cfo}
Z.~Dai, L.~Li, and W.~Xu, ``{CFO}: Conditional focused neural question
  answering with large-scale knowledge bases,'' in \emph{ACL}, vol.~1, 2016,
  pp. 800--810.

\bibitem{chen2019bidirectional}
Y.~Chen, L.~Wu, and M.~J. Zaki, ``Bidirectional attentive memory networks for
  question answering over knowledge bases,'' in \emph{NAACL}, 2019, pp.
  2913--2923.

\bibitem{mohammed2018strong}
S.~Mohammed, P.~Shi, and J.~Lin, ``Strong baselines for simple question
  answering over knowledge graphs with and without neural networks,'' in
  \emph{NAACL}, 2018, pp. 291--296.

\bibitem{bauer2018commonsense}
L.~Bauer, Y.~Wang, and M.~Bansal, ``Commonsense for generative multi-hop
  question answering tasks,'' in \emph{EMNLP}, 2018, pp. 4220--4230.

\bibitem{zhang2018variational}
Y.~Zhang, H.~Dai, Z.~Kozareva, A.~J. Smola, and L.~Song, ``Variational
  reasoning for question answering with knowledge graph,'' in \emph{AAAI},
  2018, pp. 6069--6076.

\bibitem{lin2019kagnet}
B.~Y. Lin, X.~Chen, J.~Chen, and X.~Ren, ``{KagNet}: Knowledge-aware graph
  networks for commonsense reasoning,'' in \emph{EMNLP-IJCNLP}, 2019, pp.
  2829--2839.

\bibitem{ding2019cognitive}
M.~Ding, C.~Zhou, Q.~Chen, H.~Yang, and J.~Tang, ``Cognitive graph for
  multi-hop reading comprehension at scale,'' in \emph{ACL}, 2019, pp.
  2694--2703.

\bibitem{zhang2016collaborative}
F.~Zhang, N.~J. Yuan, D.~Lian, X.~Xie, and W.-Y. Ma, ``Collaborative knowledge
  base embedding for recommender systems,'' in \emph{SIGKDD}, 2016, pp.
  353--362.

\bibitem{wang2018dkn}
H.~Wang, F.~Zhang, X.~Xie, and M.~Guo, ``{DKN}: Deep knowledge-aware network
  for news recommendation,'' in \emph{WWW}, 2018, pp. 1835--1844.

\bibitem{wang2019multi}
H.~Wang, F.~Zhang, M.~Zhao, W.~Li, X.~Xie, and M.~Guo, ``Multi-task feature
  learning for knowledge graph enhanced recommendation,'' in \emph{WWW}, 2019,
  pp. 2000--2010.

\bibitem{wang2019explainable}
X.~Wang, D.~Wang, C.~Xu, X.~He, Y.~Cao, and T.-S. Chua, ``Explainable reasoning
  over knowledge graphs for recommendation,'' in \emph{AAAI}, vol.~33, 2019,
  pp. 5329--5336.

\bibitem{xian2019reinforcement}
Y.~Xian, Z.~Fu, S.~Muthukrishnan, G.~de~Melo, and Y.~Zhang, ``Reinforcement
  knowledge graph reasoning for explainable recommendation,'' in \emph{SIGIR},
  2019.

\bibitem{wang2019kgat}
X.~Wang, X.~He, Y.~Cao, M.~Liu, and T.-S. Chua, ``{KGAT}: Knowledge graph
  attention network for recommendation,'' in \emph{SIGKDD}, 2019, pp. 950--958.

\bibitem{sharma2018towards}
A.~Sharma, P.~Talukdar \emph{et~al.}, ``Towards understanding the geometry of
  knowledge graph embeddings,'' in \emph{ACL}, 2018, pp. 122--131.

\bibitem{battaglia2018relational}
P.~W. Battaglia, J.~B. Hamrick, V.~Bapst, A.~Sanchez-Gonzalez, V.~Zambaldi,
  M.~Malinowski, A.~Tacchetti, D.~Raposo, A.~Santoro, R.~Faulkner
  \emph{et~al.}, ``Relational inductive biases, deep learning, and graph
  networks,'' \emph{arXiv preprint arXiv:1806.01261}, 2018.

\bibitem{fan2014transition}
M.~Fan, Q.~Zhou, E.~Chang, and T.~F. Zheng, ``Transition-based knowledge graph
  embedding with relational mapping properties,'' in \emph{PACLIC}, 2014, pp.
  328--337.

\bibitem{ji2016knowledge}
G.~Ji, K.~Liu, S.~He, and J.~Zhao, ``Knowledge graph completion with adaptive
  sparse transfer matrix,'' in \emph{AAAI}, 2016, pp. 985--991.

\bibitem{garcia2014effective}
A.~Garc{\'\i}a-Dur{\'a}n, A.~Bordes, and N.~Usunier, ``Effective blending of
  two and three-way interactions for modeling multi-relational data,'' in
  \emph{ECML}.\hskip 1em plus 0.5em minus 0.4em\relax Springer, 2014, pp.
  434--449.

\bibitem{reiter1978deductive}
R.~Reiter, ``Deductive question-answering on relational data bases,'' in
  \emph{Logic and data bases}.\hskip 1em plus 0.5em minus 0.4em\relax Springer,
  1978, pp. 149--177.

\bibitem{cai2018kbgan}
L.~Cai and W.~Y. Wang, ``{KBGAN}: Adversarial learning for knowledge graph
  embeddings,'' in \emph{NAACL}, 2018, pp. 1470--1480.

\bibitem{wang2017combining}
J.~Wang, Z.~Wang, D.~Zhang, and J.~Yan, ``Combining knowledge with deep
  convolutional neural networks for short text classification.'' in
  \emph{IJCAI}, 2017, pp. 2915--2921.

\bibitem{peng2019fine}
H.~Peng, J.~Li, Q.~Gong, Y.~Song, Y.~Ning, K.~Lai, and P.~S. Yu, ``Fine-grained
  event categorization with heterogeneous graph convolutional networks,'' in
  \emph{IJCAI}, 2019, pp. 3238--3245.

\bibitem{gaur2019knowledge}
M.~Gaur, A.~Alambo, J.~P. Sain, U.~Kursuncu, K.~Thirunarayan, R.~Kavuluru,
  A.~Sheth, R.~S. Welton, and J.~Pathak, ``Knowledge-aware assessment of
  severity of suicide risk for early intervention,'' in \emph{WWW}, 2019, pp.
  514--525.

\bibitem{cambria2018senticnet}
E.~Cambria, S.~Poria, D.~Hazarika, and K.~Kwok, ``{SenticNet} 5: Discovering
  conceptual primitives for sentiment analysis by means of context
  embeddings,'' in \emph{AAAI}, 2018, pp. 1795--1802.

\bibitem{ma2018targeted}
Y.~Ma, H.~Peng, and E.~Cambria, ``Targeted aspect-based sentiment analysis via
  embedding commonsense knowledge into an attentive lstm,'' in \emph{AAAI},
  2018, pp. 5876--5883.

\bibitem{liu2019knowledge}
Z.~Liu, Z.-Y. Niu, H.~Wu, and H.~Wang, ``Knowledge aware conversation
  generation with explainable reasoning over augmented graphs,'' in
  \emph{EMNLP}, 2019, pp. 1782--1792.

\bibitem{moon2019opendialkg}
S.~Moon, P.~Shah, A.~Kumar, and R.~Subba, ``{OpenDialKG}: Explainable
  conversational reasoning with attention-based walks over knowledge graphs,''
  in \emph{ACL}, 2019, pp. 845--854.

\bibitem{guo2018dialog}
D.~Guo, D.~Tang, N.~Duan, M.~Zhou, and J.~Yin, ``{Dialog-to-Action}:
  Conversational question answering over a large-scale knowledge base,'' in
  \emph{NeurIPS}, 2018, pp. 2942--2951.

\bibitem{sousa2020evolving}
R.~T. Sousa, S.~Silva, and C.~Pesquita, ``Evolving knowledge graph similarity
  for supervised learning in complex biomedical domains,'' \emph{BMC
  Bioinformatics}, vol.~21, no.~1, p.~6, 2020.

\bibitem{mohamed2020discovering}
S.~K. Mohamed, V.~Nov{\'a}{\v{c}}ek, and A.~Nounu, ``{Discovering protein drug
  targets using knowledge graph embeddings},'' \emph{Bioinformatics}, vol.~36,
  no.~2, pp. 603--610, 2020.

\bibitem{lin2020kgnn}
X.~Lin, Z.~Quan, Z.-J. Wang, T.~Ma, and X.~Zeng, ``{KGNN: Knowledge graph
  neural network for drug-drug interaction prediction}.''\hskip 1em plus 0.5em
  minus 0.4em\relax IJCAI, 2020.

\bibitem{hao2020enhancing}
B.~Hao, H.~Zhu, and I.~Paschalidis, ``{Enhancing Clinical BERT Embedding using
  a Biomedical Knowledge Base},'' in \emph{Proceedings of COLING}, 2020, pp.
  657--661.

\bibitem{li2019knowledge}
C.~Y. Li, X.~Liang, Z.~Hu, and E.~P. Xing, ``Knowledge-driven encode, retrieve,
  paraphrase for medical image report generation,'' \emph{arXiv preprint
  arXiv:1903.10122}, 2019.

\bibitem{xiong2017explicit}
C.~Xiong, R.~Power, and J.~Callan, ``Explicit semantic ranking for academic
  search via knowledge graph embedding,'' in \emph{WWW}, 2017, pp. 1271--1279.

\bibitem{wang2018zero}
X.~Wang, Y.~Ye, and A.~Gupta, ``Zero-shot recognition via semantic embeddings
  and knowledge graphs,'' in \emph{CVPR}, 2018, pp. 6857--6866.

\bibitem{liu2020attribute}
L.~Liu, T.~Zhou, G.~Long, J.~Jiang, and C.~Zhang, ``Attribute propagation
  network for graph zero-shot learning,'' in \emph{AAAI}, 2020, pp. 4868--4875.

\bibitem{koncel2019text}
R.~Koncel-Kedziorski, D.~Bekal, Y.~Luan, M.~Lapata, and H.~Hajishirzi, ``Text
  generation from knowledge graphs with graph transformers,'' in \emph{NAACL},
  2019, pp. 2284--2293.

\bibitem{seyler2017knowledge}
D.~Seyler, M.~Yahya, and K.~Berberich, ``Knowledge questions from knowledge
  graphs,'' in \emph{SIGIR}, 2017, pp. 11--18.

\bibitem{miller1995wordnet}
G.~A. Miller, ``{WordNet}: a lexical database for english,''
  \emph{Communications of the ACM}, vol.~38, no.~11, pp. 39--41, 1995.

\bibitem{matuszek2006introduction}
C.~Matuszek, M.~Witbrock, J.~Cabral, and J.~DeOliveira, ``An introduction to
  the syntax and content of cyc,'' in \emph{AAAI Spring Symposium on
  Formalizing and Compiling Background Knowledge and Its Applications to
  Knowledge Representation and Question Answering}, 2006, pp. 1--6.

\bibitem{auer2007dbpedia}
S.~Auer, C.~Bizer, G.~Kobilarov, J.~Lehmann, R.~Cyganiak, and Z.~Ives,
  ``Dbpedia: A nucleus for a web of open data,'' in \emph{The semantic web},
  2007, pp. 722--735.

\bibitem{suchanek2007yago}
F.~M. Suchanek, G.~Kasneci, and G.~Weikum, ``Yago: a core of semantic
  knowledge,'' in \emph{WWW}, 2007, pp. 697--706.

\bibitem{bollacker2008freebase}
K.~Bollacker, C.~Evans, P.~Paritosh, T.~Sturge, and J.~Taylor, ``Freebase: a
  collaboratively created graph database for structuring human knowledge,'' in
  \emph{SIGMOD}, 2008, pp. 1247--1250.

\bibitem{vrandevcic2014wikidata}
D.~Vrande{\v{c}}i{\'c} and M.~Kr{\"o}tzsch, ``Wikidata: a free collaborative
  knowledge base,'' \emph{Communications of the ACM}, vol.~57, no.~10, pp.
  78--85, 2014.

\bibitem{mccray2003upper}
A.~T. McCray, ``An upper-level ontology for the biomedical domain,''
  \emph{International Journal of Genomics}, vol.~4, no.~1, pp. 80--84, 2003.

\bibitem{moal2012skempi}
I.~H. Moal and J.~Fern{\'a}ndez-Recio, ``{SKEMPI: a Structural Kinetic and
  Energetic database of Mutant Protein Interactions and its use in empirical
  models},'' \emph{Bioinformatics}, vol.~28, no.~20, pp. 2600--2607, 2012.

\bibitem{berman2000protein}
H.~M. Berman, J.~Westbrook, Z.~Feng, G.~Gilliland, T.~N. Bhat, H.~Weissig,
  I.~N. Shindyalov, and P.~E. Bourne, ``The protein data bank,'' \emph{Nucleic
  acids research}, vol.~28, no.~1, pp. 235--242, 2000.

\bibitem{wishart2006drugbank}
D.~S. Wishart, C.~Knox, A.~C. Guo, S.~Shrivastava, M.~Hassanali, P.~Stothard,
  Z.~Chang, and J.~Woolsey, ``{DrugBank: a comprehensive resource for in silico
  drug discovery and exploration},'' \emph{Nucleic acids research}, vol.~34,
  no. suppl\_1, pp. D668--D672, 2006.

\bibitem{wishart2008drugbank}
D.~S. Wishart, C.~Knox, A.~C. Guo, D.~Cheng, S.~Shrivastava, D.~Tzur,
  B.~Gautam, and M.~Hassanali, ``{DrugBank: a knowledgebase for drugs, drug
  actions and drug targets},'' \emph{Nucleic acids research}, vol.~36, no.
  suppl\_1, pp. D901--D906, 2008.

\bibitem{riedel2010modeling}
S.~Riedel, L.~Yao, and A.~McCallum, ``Modeling relations and their mentions
  without labeled text,'' in \emph{ECML}, 2010, pp. 148--163.

\bibitem{cambria2012semantic}
E.~Cambria, Y.~Song, H.~Wang, and N.~Howard, ``Semantic multidimensional
  scaling for open-domain sentiment analysis,'' \emph{IEEE Intelligent
  Systems}, vol.~29, no.~2, pp. 44--51, 2012.

\bibitem{han2018fewrel}
X.~Han, H.~Zhu, P.~Yu, Z.~Wang, Y.~Yao, Z.~Liu, and M.~Sun, ``Fewrel: A
  large-scale supervised few-shot relation classification dataset with
  state-of-the-art evaluation,'' in \emph{EMNLP}, 2018, pp. 4803--4809.

\bibitem{toutanova2015observed}
K.~Toutanova and D.~Chen, ``Observed versus latent features for knowledge base
  and text inference,'' in \emph{ACL Workshop on CVSC}, 2015, pp. 57--66.

\bibitem{ahn2016neural}
S.~Ahn, H.~Choi, T.~P{\"a}rnamaa, and Y.~Bengio, ``A neural knowledge language
  model,'' \emph{arXiv preprint arXiv:1608.00318}, 2016.

\bibitem{bordes2015large}
A.~Bordes, N.~Usunier, S.~Chopra, and J.~Weston, ``Large-scale simple question
  answering with memory networks,'' \emph{arXiv preprint arXiv:1506.02075},
  2015.

\bibitem{trivedi2017lc}
P.~Trivedi, G.~Maheshwari, M.~Dubey, and J.~Lehmann, ``{LC-QuAD}: A corpus for
  complex question answering over knowledge graphs,'' in \emph{ISWC}, 2017, pp.
  210--218.

\bibitem{ampligraph}
L.~Costabello, S.~Pai, C.~L. Van, R.~McGrath, and N.~McCarthy, ``{AmpliGraph: a
  Library for Representation Learning on Knowledge Graphs},'' 2019.

\bibitem{han2018openke}
X.~Han, S.~Cao, L.~Xin, Y.~Lin, Z.~Liu, M.~Sun, and J.~Li, ``{OpenKE}: An open
  toolkit for knowledge embedding,'' in \emph{EMNLP}, 2018, pp. 139--144.

\bibitem{han2019opennre}
X.~Han, T.~Gao, Y.~Yao, D.~Ye, Z.~Liu, and M.~Sun, ``{O}pen{NRE}: An open and
  extensible toolkit for neural relation extraction,'' in \emph{EMNLP-IJCNLP},
  2019, pp. 169--174.

\end{thebibliography}


\begin{thebibliography}{10}
\providecommand{\url}[1]{#1}
\csname url@samestyle\endcsname
\providecommand{\newblock}{\relax}
\providecommand{\bibinfo}[2]{#2}
\providecommand{\BIBentrySTDinterwordspacing}{\spaceskip=0pt\relax}
\providecommand{\BIBentryALTinterwordstretchfactor}{4}
\providecommand{\BIBentryALTinterwordspacing}{\spaceskip=\fontdimen2\font plus
\BIBentryALTinterwordstretchfactor\fontdimen3\font minus
  \fontdimen4\font\relax}
\providecommand{\BIBforeignlanguage}[2]{{%
\expandafter\ifx\csname l@#1\endcsname\relax
\typeout{** WARNING: IEEEtran.bst: No hyphenation pattern has been}%
\typeout{** loaded for the language `#1'. Using the pattern for}%
\typeout{** the default language instead.}%
\else
\language=\csname l@#1\endcsname
\fi
#2}}
\providecommand{\BIBdecl}{\relax}
\BIBdecl

\bibitem{zhang2020learning}
Z.~Zhang, J.~Cai, Y.~Zhang, and J.~Wang, ``Learning hierarchy-aware knowledge
  graph embeddings for link prediction.'' in \emph{AAAI}, 2020, pp. 3065--3072.

\bibitem{trouillon2016complex}
T.~Trouillon, J.~Welbl, S.~Riedel, {\'E}.~Gaussier, and G.~Bouchard, ``Complex
  embeddings for simple link prediction,'' in \emph{ICML}, 2016, pp.
  2071--2080.

\bibitem{sun2018rotate}
Z.~Sun, Z.-H. Deng, J.-Y. Nie, and J.~Tang, ``{RotatE}: Knowledge graph
  embedding by relational rotation in complex space,'' in \emph{ICLR}, 2019,
  pp. 1--18.

\bibitem{zhang2019quaternion}
S.~Zhang, Y.~Tay, L.~Yao, and Q.~Liu, ``Quaternion knowledge graph embedding,''
  in \emph{NeurIPS}, 2019, pp. 2731--2741.

\bibitem{xiao2016one}
H.~Xiao, M.~Huang, and X.~Zhu, ``From one point to a manifold: Orbit models for
  knowledge graph embedding,'' in \emph{IJCAI}, 2016, pp. 1315--1321.

\bibitem{ebisu2018toruse}
T.~Ebisu and R.~Ichise, ``{TorusE}: Knowledge graph embedding on a lie group,''
  in \emph{AAAI}, 2018, pp. 1819--1826.

\bibitem{xu2019relation}
C.~Xu and R.~Li, ``Relation embedding with dihedral group in knowledge graph,''
  in \emph{ACL}, 2019, pp. 263--272.

\bibitem{balazevic2019multi}
I.~Balazevic, C.~Allen, and T.~Hospedales, ``Multi-relational poincar{\'e}
  graph embeddings,'' in \emph{NeurIPS}, 2019, pp. 4463--4473.

\bibitem{chami2020low}
I.~Chami, A.~Wolf, D.-C. Juan, F.~Sala, S.~Ravi, and C.~R{\'e},
  ``Low-dimensional hyperbolic knowledge graph embeddings,'' in \emph{ACL},
  2020.

\bibitem{he2015learning}
S.~He, K.~Liu, G.~Ji, and J.~Zhao, ``Learning to represent knowledge graphs
  with gaussian embedding,'' in \emph{CIKM}, 2015, pp. 623--632.

\bibitem{xiao2016transg}
H.~Xiao, M.~Huang, and X.~Zhu, ``{TransG}: A generative model for knowledge
  graph embedding,'' in \emph{ACL}, vol.~1, 2016, pp. 2316--2325.

\bibitem{bordes2013translating}
A.~Bordes, N.~Usunier, A.~Garcia-Duran, J.~Weston, and O.~Yakhnenko,
  ``Translating embeddings for modeling multi-relational data,'' in
  \emph{NIPS}, 2013, pp. 2787--2795.

\bibitem{lin2015learning}
Y.~Lin, Z.~Liu, M.~Sun, Y.~Liu, and X.~Zhu, ``Learning entity and relation
  embeddings for knowledge graph completion,'' in \emph{AAAI}, 2015, pp.
  2181--2187.

\bibitem{wang2014knowledge}
Z.~Wang, J.~Zhang, J.~Feng, and Z.~Chen, ``Knowledge graph embedding by
  translating on hyperplanes,'' in \emph{AAAI}, 2014, pp. 1112--1119.

\bibitem{xiao2015transa}
H.~Xiao, M.~Huang, Y.~Hao, and X.~Zhu, ``{TransA}: An adaptive approach for
  knowledge graph embedding,'' in \emph{AAAI}, 2015, pp. 1--7.

\bibitem{feng2016knowledge}
J.~Feng, M.~Huang, M.~Wang, M.~Zhou, Y.~Hao, and X.~Zhu, ``Knowledge graph
  embedding by flexible translation,'' in \emph{KR}, 2016, pp. 557--560.

\bibitem{xie2017interpretable}
Q.~Xie, X.~Ma, Z.~Dai, and E.~Hovy, ``An interpretable knowledge transfer model
  for knowledge base completion,'' in \emph{ACL}, 2017, pp. 950--962.

\bibitem{qian2018translating}
W.~Qian, C.~Fu, Y.~Zhu, D.~Cai, and X.~He, ``Translating embeddings for
  knowledge graph completion with relation attention mechanism.'' in
  \emph{IJCAI}, 2018, pp. 4286--4292.

\bibitem{ji2015knowledge}
G.~Ji, S.~He, L.~Xu, K.~Liu, and J.~Zhao, ``Knowledge graph embedding via
  dynamic mapping matrix,'' in \emph{ACL-IJCNLP}, vol.~1, 2015, pp. 687--696.

\bibitem{fan2014transition}
M.~Fan, Q.~Zhou, E.~Chang, and T.~F. Zheng, ``Transition-based knowledge graph
  embedding with relational mapping properties,'' in \emph{PACLIC}, 2014, pp.
  328--337.

\bibitem{ji2016knowledge}
G.~Ji, K.~Liu, S.~He, and J.~Zhao, ``Knowledge graph completion with adaptive
  sparse transfer matrix,'' in \emph{AAAI}, 2016, pp. 985--991.

\bibitem{garcia2014effective}
A.~Garc{\'\i}a-Dur{\'a}n, A.~Bordes, and N.~Usunier, ``Effective blending of
  two and three-way interactions for modeling multi-relational data,'' in
  \emph{ECML}.\hskip 1em plus 0.5em minus 0.4em\relax Springer, 2014, pp.
  434--449.

\bibitem{liu2017analogical}
H.~Liu, Y.~Wu, and Y.~Yang, ``Analogical inference for multi-relational
  embeddings,'' in \emph{ICML}, 2017, pp. 2168--2178.

\bibitem{zhang2019interaction}
W.~Zhang, B.~Paudel, W.~Zhang, A.~Bernstein, and H.~Chen, ``Interaction
  embeddings for prediction and explanation in knowledge graphs,'' in
  \emph{WSDM}, 2019, pp. 96--104.

\bibitem{bordes2014semantic}
A.~Bordes, X.~Glorot, J.~Weston, and Y.~Bengio, ``A semantic matching energy
  function for learning with multi-relational data,'' \emph{Machine Learning},
  vol.~94, no.~2, pp. 233--259, 2014.

\bibitem{yang2014embedding}
B.~Yang, W.-t. Yih, X.~He, J.~Gao, and L.~Deng, ``Embedding entities and
  relations for learning and inference in knowledge bases,'' in \emph{ICLR},
  2015, pp. 1--13.

\bibitem{nickel2016holographic}
M.~Nickel, L.~Rosasco, and T.~Poggio, ``Holographic embeddings of knowledge
  graphs,'' in \emph{AAAI}, 2016, pp. 1955--1961.

\bibitem{xue2018expanding}
Y.~Xue, Y.~Yuan, Z.~Xu, and A.~Sabharwal, ``Expanding holographic embeddings
  for knowledge completion,'' in \emph{NeurIPS}, 2018, pp. 4491--4501.

\bibitem{bordes2011learning}
A.~Bordes, J.~Weston, R.~Collobert, and Y.~Bengio, ``Learning structured
  embeddings of knowledge bases,'' in \emph{AAAI}, 2011, pp. 301--306.

\bibitem{kazemi2018simple}
S.~M. Kazemi and D.~Poole, ``{SimplE} embedding for link prediction in
  knowledge graphs,'' in \emph{NeurIPS}, 2018, pp. 4284--4295.

\bibitem{nickel2011three}
M.~Nickel, V.~Tresp, and H.-P. Kriegel, ``A three-way model for collective
  learning on multi-relational data,'' in \emph{ICML}, vol.~11, 2011, pp.
  809--816.

\bibitem{jenatton2012latent}
R.~Jenatton, N.~L. Roux, A.~Bordes, and G.~R. Obozinski, ``A latent factor
  model for highly multi-relational data,'' in \emph{NIPS}, 2012, pp.
  3167--3175.

\bibitem{balavzevic2019tucker}
I.~Bala{\v{z}}evi{\'c}, C.~Allen, and T.~M. Hospedales, ``{TuckER}: Tensor
  factorization for knowledge graph completion,'' in \emph{EMNLP-IJCNLP}, 2019,
  pp. 5185--5194.

\bibitem{amin2020lowfer}
S.~Amin, S.~Varanasi, K.~A. Dunfield, and G.~Neumann, ``{LowFER}: Low-rank
  bilinear pooling for link prediction,'' in \emph{ICML}, 2020, pp. 1--11.

\bibitem{dong2014knowledge}
X.~Dong, E.~Gabrilovich, G.~Heitz, W.~Horn, N.~Lao, K.~Murphy, T.~Strohmann,
  S.~Sun, and W.~Zhang, ``Knowledge vault: A web-scale approach to
  probabilistic knowledge fusion,'' in \emph{SIGKDD}.\hskip 1em plus 0.5em
  minus 0.4em\relax ACM, 2014, pp. 601--610.

\bibitem{liu2016probabilistic}
Q.~Liu, H.~Jiang, A.~Evdokimov, Z.-H. Ling, X.~Zhu, S.~Wei, and Y.~Hu,
  ``Probabilistic reasoning via deep learning: Neural association models,''
  \emph{arXiv preprint arXiv:1603.07704}, 2016.

\bibitem{dettmers2018convolutional}
T.~Dettmers, P.~Minervini, P.~Stenetorp, and S.~Riedel, ``Convolutional 2d
  knowledge graph embeddings,'' in \emph{AAAI}, vol.~32, 2018, pp. 1811--1818.

\bibitem{nguyen2017novel}
D.~Q. Nguyen, T.~D. Nguyen, D.~Q. Nguyen, and D.~Phung, ``A novel embedding
  model for knowledge base completion based on convolutional neural network,''
  in \emph{NAACL}, 2018, pp. 327--333.

\bibitem{balavzevic2019hypernetwork}
I.~Bala{\v{z}}evi{\'c}, C.~Allen, and T.~M. Hospedales, ``Hypernetwork
  knowledge graph embeddings,'' in \emph{ICANN}, 2019, pp. 553--565.

\bibitem{shang2018end}
C.~Shang, Y.~Tang, J.~Huang, J.~Bi, X.~He, and B.~Zhou, ``End-to-end
  structure-aware convolutional networks for knowledge base completion,'' in
  \emph{AAAI}, vol.~33, 2019, pp. 3060--3067.

\bibitem{socher2013reasoning}
R.~Socher, D.~Chen, C.~D. Manning, and A.~Ng, ``Reasoning with neural tensor
  networks for knowledge base completion,'' in \emph{NIPS}, 2013, pp. 926--934.

\bibitem{yang2019transms}
S.~Yang, J.~Tian, H.~Zhang, J.~Yan, H.~He, and Y.~Jin, ``{TransMS}: knowledge
  graph embedding for complex relations by multidirectional semantics,'' in
  \emph{IJCAI}, 2019, pp. 1935--1942.

\bibitem{schlichtkrull2018modeling}
M.~Schlichtkrull, T.~N. Kipf, P.~Bloem, R.~Van Den~Berg, I.~Titov, and
  M.~Welling, ``Modeling relational data with graph convolutional networks,''
  in \emph{ESWC}, 2018, pp. 593--607.

\bibitem{nathani2019learning}
D.~Nathani, J.~Chauhan, C.~Sharma, and M.~Kaul, ``Learning attention-based
  embeddings for relation prediction in knowledge graphs,'' in \emph{ACL},
  2019, pp. 4710--4723.

\bibitem{guo2019learning}
L.~Guo, Z.~Sun, and W.~Hu, ``Learning to exploit long-term relational
  dependencies in knowledge graphs,'' in \emph{ICML}, 2019, pp. 2505--2514.

\bibitem{wang2019coke}
Q.~Wang, P.~Huang, H.~Wang, S.~Dai, W.~Jiang, J.~Liu, Y.~Lyu, Y.~Zhu, and
  H.~Wu, ``{CoKE}: Contextualized knowledge graph embedding,'' \emph{arXiv
  preprint arXiv:1911.02168}, 2019.

\bibitem{yao2019kgbert}
L.~Yao, C.~Mao, and Y.~Luo, ``{KG-BERT}: {BERT} for knowledge graph
  completion,'' \emph{arXiv preprent arXiv:1909.03193}, 2019.

\bibitem{reiter1978deductive}
R.~Reiter, ``Deductive question-answering on relational data bases,'' in
  \emph{Logic and data bases}.\hskip 1em plus 0.5em minus 0.4em\relax Springer,
  1978, pp. 149--177.

\bibitem{cai2018kbgan}
L.~Cai and W.~Y. Wang, ``{KBGAN}: Adversarial learning for knowledge graph
  embeddings,'' in \emph{NAACL}, 2018, pp. 1470--1480.

\bibitem{wang2017combining}
J.~Wang, Z.~Wang, D.~Zhang, and J.~Yan, ``Combining knowledge with deep
  convolutional neural networks for short text classification.'' in
  \emph{IJCAI}, 2017, pp. 2915--2921.

\bibitem{peng2019fine}
H.~Peng, J.~Li, Q.~Gong, Y.~Song, Y.~Ning, K.~Lai, and P.~S. Yu, ``Fine-grained
  event categorization with heterogeneous graph convolutional networks,'' in
  \emph{IJCAI}, 2019, pp. 3238--3245.

\bibitem{gaur2019knowledge}
M.~Gaur, A.~Alambo, J.~P. Sain, U.~Kursuncu, K.~Thirunarayan, R.~Kavuluru,
  A.~Sheth, R.~S. Welton, and J.~Pathak, ``Knowledge-aware assessment of
  severity of suicide risk for early intervention,'' in \emph{WWW}, 2019, pp.
  514--525.

\bibitem{cambria2018senticnet}
E.~Cambria, S.~Poria, D.~Hazarika, and K.~Kwok, ``{SenticNet} 5: Discovering
  conceptual primitives for sentiment analysis by means of context
  embeddings,'' in \emph{AAAI}, 2018, pp. 1795--1802.

\bibitem{ma2018targeted}
Y.~Ma, H.~Peng, and E.~Cambria, ``Targeted aspect-based sentiment analysis via
  embedding commonsense knowledge into an attentive lstm,'' in \emph{AAAI},
  2018, pp. 5876--5883.

\bibitem{liu2019knowledge}
Z.~Liu, Z.-Y. Niu, H.~Wu, and H.~Wang, ``Knowledge aware conversation
  generation with explainable reasoning over augmented graphs,'' in
  \emph{EMNLP}, 2019, pp. 1782--1792.

\bibitem{moon2019opendialkg}
S.~Moon, P.~Shah, A.~Kumar, and R.~Subba, ``{OpenDialKG}: Explainable
  conversational reasoning with attention-based walks over knowledge graphs,''
  in \emph{ACL}, 2019, pp. 845--854.

\bibitem{guo2018dialog}
D.~Guo, D.~Tang, N.~Duan, M.~Zhou, and J.~Yin, ``{Dialog-to-Action}:
  Conversational question answering over a large-scale knowledge base,'' in
  \emph{NeurIPS}, 2018, pp. 2942--2951.

\bibitem{sousa2020evolving}
R.~T. Sousa, S.~Silva, and C.~Pesquita, ``Evolving knowledge graph similarity
  for supervised learning in complex biomedical domains,'' \emph{BMC
  Bioinformatics}, vol.~21, no.~1, p.~6, 2020.

\bibitem{mohamed2020discovering}
S.~K. Mohamed, V.~Nov{\'a}{\v{c}}ek, and A.~Nounu, ``{Discovering protein drug
  targets using knowledge graph embeddings},'' \emph{Bioinformatics}, vol.~36,
  no.~2, pp. 603--610, 2020.

\bibitem{lin2020kgnn}
X.~Lin, Z.~Quan, Z.-J. Wang, T.~Ma, and X.~Zeng, ``{KGNN: Knowledge graph
  neural network for drug-drug interaction prediction}.''\hskip 1em plus 0.5em
  minus 0.4em\relax IJCAI, 2020.

\bibitem{hao2020enhancing}
B.~Hao, H.~Zhu, and I.~Paschalidis, ``{Enhancing Clinical BERT Embedding using
  a Biomedical Knowledge Base},'' in \emph{Proceedings of COLING}, 2020, pp.
  657--661.

\bibitem{li2019knowledge}
C.~Y. Li, X.~Liang, Z.~Hu, and E.~P. Xing, ``Knowledge-driven encode, retrieve,
  paraphrase for medical image report generation,'' \emph{arXiv preprint
  arXiv:1903.10122}, 2019.

\bibitem{xiong2017explicit}
C.~Xiong, R.~Power, and J.~Callan, ``Explicit semantic ranking for academic
  search via knowledge graph embedding,'' in \emph{WWW}, 2017, pp. 1271--1279.

\bibitem{wang2018zero}
X.~Wang, Y.~Ye, and A.~Gupta, ``Zero-shot recognition via semantic embeddings
  and knowledge graphs,'' in \emph{CVPR}, 2018, pp. 6857--6866.

\bibitem{liu2020attribute}
L.~Liu, T.~Zhou, G.~Long, J.~Jiang, and C.~Zhang, ``Attribute propagation
  network for graph zero-shot learning,'' in \emph{AAAI}, 2020, pp. 4868--4875.

\bibitem{koncel2019text}
R.~Koncel-Kedziorski, D.~Bekal, Y.~Luan, M.~Lapata, and H.~Hajishirzi, ``Text
  generation from knowledge graphs with graph transformers,'' in \emph{NAACL},
  2019, pp. 2284--2293.

\bibitem{seyler2017knowledge}
D.~Seyler, M.~Yahya, and K.~Berberich, ``Knowledge questions from knowledge
  graphs,'' in \emph{SIGIR}, 2017, pp. 11--18.

\bibitem{miller1995wordnet}
G.~A. Miller, ``{WordNet}: a lexical database for english,''
  \emph{Communications of the ACM}, vol.~38, no.~11, pp. 39--41, 1995.

\bibitem{matuszek2006introduction}
C.~Matuszek, M.~Witbrock, J.~Cabral, and J.~DeOliveira, ``An introduction to
  the syntax and content of cyc,'' in \emph{AAAI Spring Symposium on
  Formalizing and Compiling Background Knowledge and Its Applications to
  Knowledge Representation and Question Answering}, 2006, pp. 1--6.

\bibitem{auer2007dbpedia}
S.~Auer, C.~Bizer, G.~Kobilarov, J.~Lehmann, R.~Cyganiak, and Z.~Ives,
  ``Dbpedia: A nucleus for a web of open data,'' in \emph{The semantic web},
  2007, pp. 722--735.

\bibitem{suchanek2007yago}
F.~M. Suchanek, G.~Kasneci, and G.~Weikum, ``Yago: a core of semantic
  knowledge,'' in \emph{WWW}, 2007, pp. 697--706.

\bibitem{bollacker2008freebase}
K.~Bollacker, C.~Evans, P.~Paritosh, T.~Sturge, and J.~Taylor, ``Freebase: a
  collaboratively created graph database for structuring human knowledge,'' in
  \emph{SIGMOD}, 2008, pp. 1247--1250.

\bibitem{carlson2010toward}
A.~Carlson, J.~Betteridge, B.~Kisiel, B.~Settles, E.~R. Hruschka, and T.~M.
  Mitchell, ``Toward an architecture for never-ending language learning,'' in
  \emph{AAAI}, 2010, pp. 1306--1313.

\bibitem{vrandevcic2014wikidata}
D.~Vrande{\v{c}}i{\'c} and M.~Kr{\"o}tzsch, ``Wikidata: a free collaborative
  knowledge base,'' \emph{Communications of the ACM}, vol.~57, no.~10, pp.
  78--85, 2014.

\bibitem{wu2012probase}
W.~Wu, H.~Li, H.~Wang, and K.~Q. Zhu, ``Probase: A probabilistic taxonomy for
  text understanding,'' in \emph{SIGMOD}, 2012, pp. 481--492.

\bibitem{mccray2003upper}
A.~T. McCray, ``An upper-level ontology for the biomedical domain,''
  \emph{International Journal of Genomics}, vol.~4, no.~1, pp. 80--84, 2003.

\bibitem{moal2012skempi}
I.~H. Moal and J.~Fern{\'a}ndez-Recio, ``{SKEMPI: a Structural Kinetic and
  Energetic database of Mutant Protein Interactions and its use in empirical
  models},'' \emph{Bioinformatics}, vol.~28, no.~20, pp. 2600--2607, 2012.

\bibitem{berman2000protein}
H.~M. Berman, J.~Westbrook, Z.~Feng, G.~Gilliland, T.~N. Bhat, H.~Weissig,
  I.~N. Shindyalov, and P.~E. Bourne, ``The protein data bank,'' \emph{Nucleic
  acids research}, vol.~28, no.~1, pp. 235--242, 2000.

\bibitem{wishart2006drugbank}
D.~S. Wishart, C.~Knox, A.~C. Guo, S.~Shrivastava, M.~Hassanali, P.~Stothard,
  Z.~Chang, and J.~Woolsey, ``{DrugBank: a comprehensive resource for in silico
  drug discovery and exploration},'' \emph{Nucleic acids research}, vol.~34,
  no. suppl\_1, pp. D668--D672, 2006.

\bibitem{wishart2008drugbank}
D.~S. Wishart, C.~Knox, A.~C. Guo, D.~Cheng, S.~Shrivastava, D.~Tzur,
  B.~Gautam, and M.~Hassanali, ``{DrugBank: a knowledgebase for drugs, drug
  actions and drug targets},'' \emph{Nucleic acids research}, vol.~36, no.
  suppl\_1, pp. D901--D906, 2008.

\bibitem{xie2017image}
R.~Xie, Z.~Liu, H.~Luan, and M.~Sun, ``Image-embodied knowledge representation
  learning,'' in \emph{IJCAI}, 2017, pp. 3140--3146.

\bibitem{riedel2010modeling}
S.~Riedel, L.~Yao, and A.~McCallum, ``Modeling relations and their mentions
  without labeled text,'' in \emph{ECML}, 2010, pp. 148--163.

\bibitem{cambria2012semantic}
E.~Cambria, Y.~Song, H.~Wang, and N.~Howard, ``Semantic multidimensional
  scaling for open-domain sentiment analysis,'' \emph{IEEE Intelligent
  Systems}, vol.~29, no.~2, pp. 44--51, 2012.

\bibitem{han2018fewrel}
X.~Han, H.~Zhu, P.~Yu, Z.~Wang, Y.~Yao, Z.~Liu, and M.~Sun, ``Fewrel: A
  large-scale supervised few-shot relation classification dataset with
  state-of-the-art evaluation,'' in \emph{EMNLP}, 2018, pp. 4803--4809.

\bibitem{sun2017cross}
Z.~Sun, W.~Hu, and C.~Li, ``Cross-lingual entity alignment via joint
  attribute-preserving embedding,'' in \emph{ISWC}, 2017, pp. 628--644.

\bibitem{sun2018bootstrapping}
Z.~Sun, W.~Hu, Q.~Zhang, and Y.~Qu, ``Bootstrapping entity alignment with
  knowledge graph embedding.'' in \emph{IJCAI}, 2018, pp. 4396--4402.

\bibitem{hao2019universal}
J.~Hao, M.~Chen, W.~Yu, Y.~Sun, and W.~Wang, ``Universal representation
  learning of knowledge bases by jointly embedding instances and ontological
  concepts,'' in \emph{KDD}, 2019, pp. 1709--1719.

\bibitem{toutanova2015observed}
K.~Toutanova and D.~Chen, ``Observed versus latent features for knowledge base
  and text inference,'' in \emph{ACL Workshop on CVSC}, 2015, pp. 57--66.

\bibitem{ahn2016neural}
S.~Ahn, H.~Choi, T.~P{\"a}rnamaa, and Y.~Bengio, ``A neural knowledge language
  model,'' \emph{arXiv preprint arXiv:1608.00318}, 2016.

\bibitem{bordes2015large}
A.~Bordes, N.~Usunier, S.~Chopra, and J.~Weston, ``Large-scale simple question
  answering with memory networks,'' \emph{arXiv preprint arXiv:1506.02075},
  2015.

\bibitem{trivedi2017lc}
P.~Trivedi, G.~Maheshwari, M.~Dubey, and J.~Lehmann, ``{LC-QuAD}: A corpus for
  complex question answering over knowledge graphs,'' in \emph{ISWC}, 2017, pp.
  210--218.

\bibitem{ampligraph}
L.~Costabello, S.~Pai, C.~L. Van, R.~McGrath, and N.~McCarthy, ``{AmpliGraph: a
  Library for Representation Learning on Knowledge Graphs},'' 2019.

\bibitem{han2018openke}
X.~Han, S.~Cao, L.~Xin, Y.~Lin, Z.~Liu, M.~Sun, and J.~Li, ``{OpenKE}: An open
  toolkit for knowledge embedding,'' in \emph{EMNLP}, 2018, pp. 139--144.

\bibitem{han2019opennre}
X.~Han, T.~Gao, Y.~Yao, D.~Ye, Z.~Liu, and M.~Sun, ``{O}pen{NRE}: An open and
  extensible toolkit for neural relation extraction,'' in \emph{EMNLP-IJCNLP},
  2019, pp. 169--174.

\end{thebibliography}

\end{document}